\documentclass[10pt]{article}
\usepackage{arxiv}
\usepackage[T1]{fontenc}
\usepackage{courier}
\usepackage{amsmath,amssymb,amsthm,mathtools}
\usepackage{graphicx,xcolor,booktabs,array,enumitem}
\usepackage[authoryear,round]{natbib}
\usepackage[algo2e,ruled,vlined]{algorithm2e}
\usepackage{tikz}
\usetikzlibrary{arrows.meta}
\usepackage[colorlinks=true,linkcolor=blue,citecolor=blue,urlcolor=blue]{hyperref}
\usepackage{cleveref}
\newtheorem{theorem}{Theorem}[section]
\newtheorem{lemma}[theorem]{Lemma}
\newtheorem{proposition}[theorem]{Proposition}
\newtheorem{corollary}[theorem]{Corollary}
\theoremstyle{definition}
\newtheorem{definition}[theorem]{Definition}

\newtheorem{exmp}[theorem]{Example}
\theoremstyle{remark}
\newtheorem{remark}[theorem]{Remark}

\setcitestyle{authoryear,round,comma}
\crefname{theorem}{Theorem}{Theorems}
\crefname{lemma}{Lemma}{Lemmas}
\crefname{proposition}{Proposition}{Propositions}
\crefname{corollary}{Corollary}{Corollaries}
\crefname{definition}{Definition}{Definitions}
\crefname{remark}{Remark}{Remarks}
\crefname{example}{Example}{Examples}
\crefname{exmp}{Example}{Examples}
\crefname{section}{Section}{Sections}
\crefname{subsection}{Section}{Sections}
\crefname{subsubsection}{Section}{Sections}
\crefname{appendix}{Appendix}{Appendices}
\crefname{table}{Table}{Tables}
\crefname{figure}{Figure}{Figures}
\crefname{equation}{Equation}{Equations}
\crefname{algocf}{Algorithm}{Algorithms}
\crefname{algorithm}{Algorithm}{Algorithms}

\newcommand{\R}{\mathbb{R}}
\newcommand{\E}{\mathbb{E}}
\newcommand{\sign}{\operatorname{sign}}
\newcommand{\DIS}{\operatorname{DIS}}
\newcommand{\err}{\operatorname{err}}
\newcommand{\Sph}{S^{n-1}}
\newcommand{\eps}{\varepsilon}
\newcommand{\Dsym}{\mathcal D^{\mathrm{sym}}_{n,s}}
\newcommand{\Dns}{\mathcal D_{n,s}}

\title{When Can Unlabeled Data Pools Replace Membership Queries?}
\author{Dravyansh Sharma\thanks{Most of the work was done while the author was at the Toyota Technological Institute at Chicago (TTIC).}\\Microsoft}
\date{}
\newif\ifextendedproofs
\extendedproofstrue
\begin{document}
\maketitle
\begin{abstract}%
Membership queries let a learner label arbitrary points and give near-optimal $\tilde O(n\log(1/\eps))$ label complexity for halfspaces under every marginal. A pool learner may label only sampled points. We ask when unlabeled data can replace membership queries. For homogeneous halfspaces the answer depends only on the angular law of $X/\|X\|$. If this law assigns zero mass to every hyperplane through the origin and positive mass to every nonempty open set of directions together with its reflection, a pool learner uses $O(n\log^3n\log(n/(\eps\delta)))$ labels, uniformly over all such marginals and targets, without density, moment, concavity or isotropy assumptions. The same label bound holds for non-homogeneous halfspaces under every marginal with full support and no mass on affine hyperplanes, including Gaussians. 
Dropping either angular assumption admits $\Omega(1/\eps)$ pool versus $O(\log(1/\eps))$ membership-query separations, even with unlimited unlabeled data. The price is paid in unlabeled data: in dimension $n\ge3$, for every unlabeled budget fixed in advance, some marginal satisfying both assumptions forces $\Omega(1/\eps)$ labels.
For isotropic $s$-concave marginals in dimension $n\ge3$, unlimited pools keep logarithmic label rates for every finite $s$. With an unlabeled budget fixed in advance, class-uniform polylogarithmic rates hold exactly for $s>-1/(n+1)$, or $s>-1/n$ under central symmetry, and in the general class the optimal rate between these thresholds is $\tilde\Theta(\eps^{-(n-2)/n})$. This extends the range $s\ge-1/(2n+3)$ of previous sample-based guarantees, in the realizable setting.
\end{abstract}

\section{Introduction}

Membership queries are an unusually powerful form of active learning. The learner may choose any point and ask for its label. For halfspaces, \emph{point location}, which labels every point of a given finite set under an unknown halfspace, can be done for a passive sample of size $N$ with $O(n\log^2n\log N)$ synthesized queries \citep{HKLM20}, giving near-optimal membership-query learning under every marginal. An ordinary pool learner may only request labels of points it has sampled, and on worst-case finite point sets this can require $\Omega(N)$ labels \citep{Das04,BLMSS26}. For affine halfspaces under Gaussian marginals and a pool of $m$ points, \citet{DKM24} prove an information-theoretic $\tilde\Omega(n/(\eps\log m))$ lower bound. We ask the following distributional question (ignoring computational hardness).\looseness-1
\begin{quote}\itshape
For which marginals can an unlabeled pool substitute for synthesized membership queries without increasing label complexity?
\end{quote}
We count two resources separately, namely \emph{labels} and \emph{unlabeled examples}, the latter meaning every sampled point the learner inspects. For halfspaces through the origin $x\mapsto\sign\langle w,x\rangle$ only the direction of a point matters, so the relevant object is the \emph{angular law} $\mu=\mathcal L(X/\|X\|)$, the distribution of the direction of $X$. 

\subsection{Summary of Results}

\begin{theorem}[Pools match membership queries in labels, \cref{thm:main}]\label{thm:informal-main}
Let $n\ge2$ and let $D$ have no atom at the origin and an angular law that is \emph{hyperplane-null} (it gives probability zero to every hyperplane through the origin) and has \emph{symmetric full support} (every nonempty open set $O$ of directions has $\mu(O\cup-O)>0$, which holds in particular if every nonempty open set of directions has positive probability, \cref{sec:angular-laws}). A pool learner achieves error $\eps$ with probability $1-\delta$ using
\[
O\Bigl(n\log^3n\,\log\frac{n}{\eps\delta}\Bigr)
\]
labels, uniformly over all such $D$ and all targets. This matches the $\Omega(n\log(1/\eps))$ lower bound under the uniform distribution up to a $\log^3n$ factor. The number of unlabeled examples is finite almost surely, necessarily depends on $D$ (\cref{thm:informal-lower}), and for our simulation is exponential in $n$ even for the uniform distribution (\cref{rem:pathwise}).
\end{theorem}
Like the lower bounds below, \cref{thm:informal-main} is information-theoretic. It counts labels and unlabeled examples, not running time (\cref{sec:explicit-floor}). The two assumptions can also be placed on different distributions. Error can be measured under a hyperplane-null law while labels come from points of a second law with symmetric full support, which may be atomic. The evaluation law itself must remain hyperplane-null (\cref{thm:two-laws}).

A simpler learner, \emph{directional search}, approximates each membership query at a fixed angular resolution. It uses $O(n\log(n/\rho_\mu^{-1}(\eps/16)))$ labels, where $\rho_\mu^{-1}(\eps)$ is the largest width $\theta$ such that every band $\{u:|\langle u,w\rangle|<\theta\}$ has $\mu$-mass at most $\eps$ (\cref{prop:baseline}). A mixture of three centered Gaussians makes this procedure's label count arbitrarily large at fixed $\eps$, even after an invertible linear change of coordinates (\cref{ex:mixture}). In contrast, \cref{thm:informal-main} estimates query precision from unlabeled data and uses robust point location to confine its marginal dependence to the unlabeled cost.

\begin{theorem}[Both angular assumptions are necessary, \cref{prop:need-support,prop:need-null}]\label{thm:informal-necessity}
For every $\eps$ there are two marginals, each depending on $\eps$, for which every pool learner needs $\Omega(1/\eps)$ labels even with unlimited unlabeled data, while membership queries need $O(\log(1/\eps))$. The first is a smooth, centrally symmetric, hyperplane-null marginal on $\R^3$ without symmetric full support, with a finite family of targets. The second is a full-support marginal on $\R^4$ that charges a hyperplane, with a family of targets containing arbitrarily small tilts of one target, and there the lower bound holds in expectation.
\end{theorem}

\noindent When both assumptions hold, pools recover the membership-query label rate, but the number of unlabeled examples in \cref{thm:informal-main} depends on the marginal, and this dependence is unavoidable.\looseness-1

\begin{theorem}[Unlabeled cost cannot be uniform, \cref{prop:finite-budget}]\label{thm:informal-lower}
Let $n\ge3$. For every unlabeled budget $U(\eps)$, some marginal with a smooth, everywhere-positive, centrally symmetric density, so that both angular assumptions hold, forces every pool learner that inspects at most $U(\eps)$ examples to use $\Omega(1/\eps)$ labels, information-theoretically. For fixed $n$ this is the passive rate.
\end{theorem}
Here the hard distribution may depend on the budget and on $\eps$. We do not claim that a single distribution defeats every finite pool. Together with \cref{thm:informal-main}, this shows that labels admit a bound uniform over the marginal class, whereas the number of unlabeled examples does not. Inside structured classes the same mechanism gives refined exponents. When homogeneous halfspaces act on a high-probability core as affine halfspaces of a lower-dimensional ball (\cref{def:affine-core}), every finite budget forces $\Omega_n(\eps^{-(n-2)/n})$ labels (\cref{thm:affine-core}), which is tight for $s$-concave marginals with $-1/n<s\le-1/(n+1)$ (\cref{sec:sconcave}).

\begin{theorem}[$s$-concave marginals, \cref{sec:sconcave}]\label{thm:informal-phase}
For isotropic (mean zero, identity covariance) $s$-concave marginals in dimension $n\ge3$ (\cref{def:sconcave}, where $s=0$ is log-concavity, and more negative $s$ allows heavier tails):
(i) with an unlimited pool, $O(n\log^3n\log(n/(\eps\delta)))$ labels suffice for every finite $s$ (\cref{thm:main}), compared with the generic $\Omega(n\log(1/\eps))$ lower bound;
(ii) with an unlabeled budget fixed in advance for the class, polylogarithmic labels are achievable, via the disagreement-based learner CAL \citep{CAL94} and a class-uniform disagreement bound $O_{n,s}(\log(1/\eps))$ (\cref{sec:disagreement}), if and only if $s>-1/(n+1)$, or $s>-1/n$ in the centrally symmetric subclass ($f(x)=f(-x)$);
(iii) for $-1/n<s\le-1/(n+1)$ in the general class, the worst-case disagreement coefficient is $\Theta_{n,s}(\eps^{-(n-2)/n})$, and $\tilde\Theta_{n,s}(\eps^{-(n-2)/n})$ labels are necessary and sufficient with an unlabeled budget fixed in advance, at constant confidence;
(iv) for $s\le-1/n$ both the disagreement coefficient and the label count are $\Theta(1/\eps)$, even in the symmetric subclass.
Constants depend on $n$ and $s$. In regime (ii) we prove no lower bound beyond $\Omega(n\log(1/\eps))$. All statements concern the realizable setting.
\end{theorem}
\citet{BZ17} give polynomial-time, noise-tolerant learners for $s\ge-1/(2n+3)$. \Cref{thm:informal-phase} is information-theoretic apart from the CAL implementation (\cref{cor:cal}). It locates the thresholds for class-uniform rates and the sharp rate between them, rather than extending their computational or noise guarantees. Item~(i) follows from \cref{thm:main}. Items~(ii)--(iv) use CAL and \cref{lem:star}, with new arguments for the exponent in~(iii) (\cref{sec:exponent-proof}) and the symmetric construction in~(iv) (\cref{lem:radial-mixture}). %

\subsection{Techniques}\label{sec:techniques}
\paragraph{A target-independent margin scale.} A compactness argument implies that hyperplane-nullness is equivalent to the band mass $\rho_\mu(\theta)=\sup_{\|w\|=1}\mu\{u:|\langle u,w\rangle|<\theta\}$ vanishing with $\theta$ (\cref{lem:anticonc}). A uniform relative-deviation (VC) bound then finds, from unlabeled data alone, a width $\theta_0>0$ (a \emph{margin floor}) with the following property. The learner draws an independent \emph{passive sample}, an i.i.d.\ sample of size $N=\tilde O(n/\eps)$ that it will label in full by point location, so that every consistent halfspace is $\eps$-accurate. With high probability, every passive point has normalized margin at least $\theta_0$ from the \emph{fixed} unknown target. No rate at which the band mass $\rho_\mu$ vanishes is assumed, which is why labels are uniform while precision and unlabeled cost are not. This geometric margin is unrelated to the Tsybakov margin condition on label noise.

\paragraph{Answering queries with pool points.} The point-location algorithm of \citet{HKLM20} repeatedly restricts to a subspace $V$ containing many sample points and applies a linear map $T$ that puts those points in approximately isotropic position (their second-moment matrix becomes nearly a multiple of the identity). Every such subspace contains a passive point, which lower-bounds the norm of the projected target. Hence a pool point within angle $\gamma$ of a requested direction (or of its antipode, with the label negated) returns the correct label unless the transformed query $q$ has relative margin $|\langle q,w'\rangle|/(\|q\|\|w'\|)$ at most $\gamma\kappa(T)/\theta_0$ with respect to the transformed target normal $w'$, where $\kappa(T)=\|T\|\|T^{-1}\|$ is the condition number (\cref{lem:pool-oracle}). Because the learner chooses $\gamma$ after seeing $T$, the pool answers form a \emph{subspace-relative} oracle (\cref{def:sr-oracle}). This is stronger than a single margin-tolerant oracle in fixed coordinates, and it is what robust point location needs.

\paragraph{Robust point location.} We modify \citet{HKLM20} to tolerate such an oracle. Their margin comparisons already carry multiplicative slack. The one step that needs exact comparisons is deterministic median selection, which we replace by a round-robin tournament robust to near-ties, and we enlarge one acceptance threshold. This costs one $\log n$ factor (\cref{prop:robust-pl}). \Cref{tab:hklm} lists, with their hypotheses, all results of \citet{HKLM20} that we use.

\paragraph{Hard instances for the lower bounds.} The lower bounds place the informative directions in a region of vanishing mass. In the general $s$-concave class, a rare tail moves the mean, while a high-probability core is, after a projective change of coordinates, a uniform ball. In the symmetric class, successive logarithmic radial scales are dominated by different norms, so the angular law can approximate a mixture of the angular laws of elongated ellipsoids.

\paragraph{Relation to prior work.} The angular law (radially symmetric marginals were already exploited for origin-centered separators by \citealt{Kea98}), exponential active improvements, pool-versus-synthesis gaps, and the splitting-index distinction between how informative a query is and how likely the unlabeled distribution is to produce it \citep{Das05,Han14} are all classical. That a sufficiently rich pool approximates membership queries is folklore. The infinite full-support pool limit is discussed in \citet{GSS13}, and \citet{DKM24} remark that pools approach membership queries as they grow exponentially. Dasgupta's label bounds carry logarithmic factors that depend on the marginal, as does the directional search of \cref{prop:baseline}. Our contributions are threefold. \Cref{thm:main} removes every marginal-dependent factor from the label count. We show that both angular assumptions are needed. Finally, we show that the remaining marginal dependence, in unlabeled data, cannot be removed (\cref{thm:informal-lower}). \Cref{sec:related} gives detailed comparisons, including with universal and marginal-nonuniform rates, and separates new arguments from results of prior work.

\section{Setup}

Throughout, $\tilde O$ hides polylogarithmic factors, and subscripts such as $O_n$, $\Omega_{n}$, $O_{n,s}$ indicate constants depending only on those parameters.

\subsection{Pool and membership-query models}\label{sec:models}
Let $h_w(x)=\sign\langle w,x\rangle$, $w\in\Sph$, be the class of homogeneous halfspaces, with the convention $\sign(0)=+1$ throughout. The target normal is $w^\star$ and the target is $h^\star=h_{w^\star}$. In the membership-query model, the learner may choose any $q\in\R^n$ and request $h^\star(q)$. In the ordinary i.i.d.\ pool model, the learner observes an unlabeled stream $X_1,X_2,\ldots\sim D$ and may request labels only for points that have appeared in that stream. We count label queries and unlabeled examples separately. An unlabeled budget $U(\eps)$ bounds the number of stream elements the learner may inspect, whether or not it requests their labels.

For a finite passive sample $\mathcal X\subset\R^n$, point location asks for the labels of all points in $\mathcal X$ under an unknown homogeneous halfspace. The synthesized-query algorithm of \citet{HKLM20} uses $\tilde O(n\log|\mathcal X|)$ sign queries. Combined with the standard realizable VC bound, this gives near-optimal membership-query active learning under every marginal.

\subsection{Angular laws}\label{sec:angular-laws}
Because $h_w(rx)=h_w(x)$ for every $r>0$, only direction matters. For a distribution $D$ that assigns zero mass to the origin, its angular law is the pushforward
\[
\mu=\mathcal L\Bigl(\frac{X}{\|X\|}\Bigr)\quad\text{on } \Sph,\qquad \mu(B)=\Pr\{X/\|X\|\in B\}.
\]
A spherical density $d\mu/d\sigma$ relative to normalized surface measure $\sigma$ need not exist. Consequently $D(h_v\ne h_w)=\mu(h_v\ne h_w)$, and every event determined solely by homogeneous-halfspace labels depends only on $\mu$. More generally, for halfspaces through a reference point $x_0$, replace $X$ by $X-x_0$.

For $\gamma>0$ define the worst directional cap mass
\[
a_\mu(\gamma)=\inf_{q\in\Sph}\mu\{u:\angle(u,q)<\gamma\}.
\]
For $\theta>0$ also define the worst band mass
\[
\rho_\mu(\theta)=\sup_{\|w\|=1}\mu\{u:|\langle u,w\rangle|<\theta\}.
\]
We call $\mu$ \emph{hyperplane-null} if $\mu(H\cap\Sph)=0$ for every linear hyperplane $H$. The sets $H\cap\Sph$ are the \emph{great subspheres}. It has \emph{full support} if every nonempty open subset of $\Sph$ has positive $\mu$-probability, equivalently $a_\mu(\gamma)>0$ for all $\gamma>0$. Write $\bar\mu=\tfrac12(\mu+\mu\circ(-\mathrm{id}))$ for the symmetrization of $\mu$. We say $\mu$ has \emph{symmetric full support} if $\bar\mu$ has full support, that is, if $\mu(O\cup-O)>0$ for every nonempty open $O\subseteq\Sph$. The name refers to the symmetrization, and $\mu$ itself need not be symmetric. Full support implies symmetric full support, and the two coincide for centrally symmetric laws. Hyperplane-nullness of $\mu$ and of $\bar\mu$ coincide, since great subspheres are symmetric. Because $h_w(-x)=-h_w(x)$ whenever $\langle w,x\rangle\ne0$, a sampled direction near $-q$ is as informative as one near $q$. Hyperplane-nullness will imply $\rho_\mu(\theta)\to0$ as $\theta\downarrow0$, which is the anti-concentration needed to make query precision independent of the unknown target. Singular angular laws are allowed.

\paragraph{Margins and linear algebra.} The \emph{normalized margin} of $x\ne0$ with respect to a normal $w\ne0$ is $|\langle x,w\rangle|/(\|x\|\|w\|)$, the absolute cosine of the angle between $x$ and $w$. For a subspace $V$, $P_V$ is orthogonal projection onto $V$. For an invertible linear map $T$, $T^{-\top}$ is the transpose of $T^{-1}$ and $\kappa(T)=\|T\|\|T^{-1}\|$ is its condition number.

\subsection{Disagreement and passive generalization}\label{sec:disagreement}
For unit normals $u,v$ write $d_D(u,v)=\Pr_{X\sim D}[h_u(X)\ne h_v(X)]$. For a target $h^\star$ let $B_D(h^\star,r)=\{h:D(h\ne h^\star)\le r\}$, and for a set $\mathcal H$ of hypotheses let $\DIS(\mathcal H)$ be the set of points on which some two hypotheses in $\mathcal H$ disagree. The \emph{disagreement coefficient} measures how much of the space is still ambiguous among hypotheses of error at most $r$, relative to $r$. It governs the label complexity of disagreement-based active learners such as CAL, the algorithm of \citet{CAL94} \citep[see also][]{Han14}, which query only points on which the surviving hypotheses disagree.\looseness-1
\[
\theta_D(\eps;h^\star)=\sup_{r\ge\eps}\frac{D(\DIS(B_D(h^\star,r)))}{r}.
\]
A class has \emph{logarithmic disagreement} if $\theta_D(\eps;h^\star)=O_{n,s}(\log(1/\eps))$ uniformly over its laws and targets. We use only the standard realizable VC fact that a correctly labeled i.i.d.\ sample of size $N=O\bigl(\frac{n\log(1/\eps)+\log(1/\delta)}{\eps}\bigr)$ suffices for every consistent homogeneous halfspace to have error at most $\eps$ with probability at least $1-\delta$.

\section{From a pool to membership queries}\label{sec:simulation}
Hyperplane-nullness gives a target-independent margin scale, and symmetric full support places pool points in every angular cap or its reflection. The robust point location connecting them is summarized in \cref{sec:pointloc}. \ifextendedproofs Its proof is given in \cref{sec:robust-main-proof}, with supporting lemmas in \cref{app:robust}.\fi

\subsection{A target-independent margin floor}
A \emph{margin floor} is a positive scale $\theta_0$, computed from unlabeled data without knowledge of the target, such that very little angular probability lies within $\theta_0$ of \emph{any} origin-centered hyperplane.
\[
\rho_\mu(\theta_0)=\sup_{\|w\|=1}\Pr_{Y\sim\mu}\bigl(|\langle w,Y\rangle|<\theta_0\bigr)\le\alpha.
\]
For an independent sample of size $N$ and any \emph{fixed} target $w^\star$, a union bound then gives\[\Pr\bigl(\min_{j\le N}|\langle w^\star,Y_j\rangle|<\theta_0\bigr)\le N\alpha.\] The floor is not a lower bound on every margin, only a margin scale that holds with high probability for a fixed target.

\begin{lemma}[Uniform anti-concentration]\label{lem:anticonc}
For $n\ge2$, the following are equivalent: $\mu$ assigns zero mass to every great subsphere, $\rho_\mu(\theta)\to0$ as $\theta$ decreases to $0$, and every $n$ independent sampled directions are linearly independent almost surely.
\end{lemma}
The proof (\cref{app:deferred}) is a compactness argument over the normal sphere.

\begin{lemma}[Computing a margin floor]\label{lem:floor}
Fix a passive-sample size $N$ and confidence $\delta$. From
\[
M_0\ge C\frac{N}{\delta}\bigl(n\log M_0+\log(8/\delta)\bigr),\qquad M_0\ge\frac{32nN}{\delta},
\]
independent unlabeled directions, one can compute $\theta_0>0$ such that with probability at least $1-\delta/8$, $\rho_\mu(\theta_0)\le\delta/(8N)$. The computation is finite. \Cref{sec:explicit-floor} gives an explicit implementation in time $M_0^{O(n)}$.
\end{lemma}

\begin{corollary}[Passive-sample margin floor]\label{cor:passive-floor}
Let $\mathcal X$ be an independent passive sample of size $N$ and let $w^\star$ be a fixed unit target normal. With probability at least $1-\delta/4$, every $x\in\mathcal X$ has normalized margin $|\langle x,w^\star\rangle|/\|x\|\ge\theta_0$.
\end{corollary}
\begin{proof}
Condition on a successful computation of $\theta_0$. Every point lies in the target band of width $\theta_0$ with probability at most $\delta/(8N)$, so a union bound over the $N$ passive points costs at most $\delta/8$. Add the failure probability of \cref{lem:floor}.
\end{proof}

\subsection{Approximating a synthesized query}
\begin{definition}[Margin-tolerant oracles]\label{def:mt-oracle}
Let $w^\star$ be the target normal and $\eta\in(0,1)$. An \emph{$\eta$-margin-tolerant} oracle answers a query $q\ne0$ with $\sign\langle w^\star,q\rangle$ whenever $|\langle w^\star,q\rangle|>\eta\|w^\star\|\|q\|$, and arbitrarily, even inconsistently, otherwise. A \emph{query instance} is a pair $(V_0,T)$ of a subspace $V_0$ with $P_{V_0}w^\star\ne0$ and an invertible linear map $T:V_0\to V_0$. Its transformed target normal is $w'=T^{-\top}P_{V_0}w^\star$.\phantomsection\label{def:sr-oracle} An \emph{$\eta$-subspace-relative} oracle answers a query $z'\in V_0\setminus\{0\}$ in instance $(V_0,T)$ with $\sign\langle z',w'\rangle=\sign\langle T^{-1}z',w^\star\rangle$, the label of $T^{-1}z'$, whenever $|\langle z',w'\rangle|>\eta\|z'\|\|w'\|$, and arbitrarily otherwise.
\end{definition}
The parameter $\eta$ is a geometric tolerance. The Massart flip rate in \cref{sec:noise} is written $\beta$. A single margin-tolerant oracle is too weak for point location, which works in transformed coordinates. For $\eta>2a>0$ and $w_\pm=(\pm a,\sqrt{1-a^2})$, the oracle $q\mapsto\sign(q_2)$ is $\eta$-margin-tolerant for both targets, which label $e_1$ differently. A subspace-relative oracle must label $e_1$ correctly in the instance $(\mathrm{span}(e_1),I)$, where its relative margin is $1$.

\begin{lemma}[Pool realization of the subspace-relative oracle]\label{lem:pool-oracle}
Let $(V_0,T)$ be a query instance such that $V_0$ contains a passive-sample point whose normalized target margin is at least $\theta_0$, let $z'\in V_0\setminus\{0\}$ be a query, and put $z=T^{-1}z'$ and $w'=T^{-\top}P_{V_0}w^\star$. If a pool point $p$ satisfies $\angle(p,z)\le\gamma$, then its label can differ from $\sign\langle z',w'\rangle$ only when
\[
|\langle z',w'\rangle|\le\eta\|z'\|\|w'\|,\qquad \eta=\frac{\gamma\kappa(T)}{\theta_0}.
\]
Hence answering each query with the label of a pool point within angle $\eta\theta_0/\kappa(T)$ of $T^{-1}z'$ realizes an $\eta$-subspace-relative oracle on all such instances.
\end{lemma}
\begin{proof}
A wrong label forces $|\langle z,w^\star\rangle|\le\gamma\|z\|\|w^\star\|$, while $\langle z',w'\rangle=\langle z,P_{V_0}w^\star\rangle=\langle z,w^\star\rangle$ because $z\in V_0$. If $x\in V_0$ has normalized target margin at least $\theta_0$, then
$\|w'\|\ge\|P_{V_0}w^\star\|/\|T\|\ge\theta_0\|w^\star\|/\|T\|$ and $\|z\|\le\|T^{-1}\|\|z'\|$. Combining the inequalities gives the claim.
\end{proof}
The same conclusion holds for the negated label of a pool point $p$ with $\angle(-p,z)\le\gamma$. Outside the tolerance band $|\langle z,w^\star\rangle|>\gamma\|z\|\|w^\star\|$, so every direction within angle $\gamma$ of $z$ lies strictly on one side of the target hyperplane. Then $-p$ has the correct label, and since $\langle w^\star,p\rangle\ne0$ it equals minus the label of $p$. Neither argument uses any property of the law from which $p$ was drawn. The point $p$ may be an atom, and points on the target hyperplane can occur only for queries inside the band, where any answer is admissible.

\subsection{Margin-tolerant point location}\label{sec:pointloc}
The reduction needs a point-location algorithm that stays correct under the subspace-relative oracle of \cref{def:sr-oracle}, where queries of very small relative margin in the current transformed instance may receive arbitrary answers.

\paragraph{The algorithm of \citet{HKLM20} in brief.} Its weak learner, PartialLearn, labels a constant fraction of a weighted point set $(X,\omega)$ with few queries. A boosting loop reweights the unlabeled points and repeats until all of $X$ is labeled. PartialLearn first finds a subspace $V$ carrying many points and a linear map $T$ that puts them in approximately \emph{isotropic position}. As in \citet{HKLM20}, normalized points $(X,\omega)$ in $\R^d$ are \emph{$c$-isotropic} if $(1-c)/d\le\E_\omega\langle x,v\rangle^2\le(1+c)/d$ for every unit $v$. It then runs IsoLearn, in which StructureSearch compares the target margins of random sample \emph{representatives} (Gaussian combinations $x_S=\sum_{x\in S}g_xx$ of a sample $S$, whose margin tracks $(\sum_{x\in S}m(x)^2)^{1/2}$, where $m(x)$ is the margin of $x$ with respect to the normalized transformed target) across sample sizes until it finds a gap separating a few large-margin points from many small-margin ones. DimReduce estimates, from accepted small-margin subsets, a large subspace on which the target's margin is tiny, and on the remaining few directions the target is recovered by binary search, after which every point whose margin is clearly decided is labeled.\looseness-1 

\paragraph{Prior robustness guarantees.} The guarantees of \citet{HKLM20} allow the algorithm itself to fail with small probability (their ``bounded-error'' and ``reliable'' notions refer to this failure probability), but they assume every query is answered correctly. Their companion work with comparison queries \citep{HKLM20noise} handles stochastic bounded (Massart) label noise rather than arbitrary answers inside a relative-margin band. The closest model is the unreliable boundary queries of \citet{BCGS95}, where membership queries near the boundary may be answered wrongly and the distribution puts no mass there, but their goal is polynomial-time PAC learning of specific classes rather than near-optimal query complexity for all halfspaces. We are not aware of a near-optimal point-location algorithm with guarantees in this setting. Since a nearby sampled direction can be wrong deterministically, majority votes do not help.

\paragraph{Our modifications.} The isotropic-transform arguments of \citet{HKLM20}, which rest on the theorems of \citet{For02} and \citet{Bar98}, and their matrix-concentration, structural and boosting arguments do not involve the oracle. \Cref{app:robust} checks the hypotheses of each result we use under the new oracle. Only three oracle-sensitive steps change. \emph{Comparison.} $|m(a)|$ and $|m(b)|$ are compared via the signs of $a$, $b$ and a signed difference, which is correct whenever the magnitudes differ by more than the tolerance-induced error (\cref{lem:comparator}). \emph{Selection.} Deterministic median selection is fragile under such a comparator, so StructureSearch instead compares all pairs in a small representative set and picks a middle win count, which controls the rank of the selected representative from both sides up to one comparator error (\cref{lem:roundrobin}). This is the classical paired-comparison method in the bounded-error model of \citet{AFHN15}, and it costs one extra $\log n$ factor. \emph{Dimension reduction.} DimReduce accepts a sampled subset when its representative's margin is small relative to a scaled reference point, tested with three sign queries, and spurious accepted subsets only add positive semidefinite terms to the covariance estimate \citep[Lemma~4.15]{HKLM20}. We only enlarge the acceptance threshold so that subsets of small-margin points are still accepted under comparison error (\cref{lem:dimreduce}).

\begin{proposition}[Robust point location]\label{prop:robust-pl}
Let $X\subset\R^n$ be a set of $N$ points, none on the target hyperplane. There exists a tolerance $\eta^*=n^{-O(\log n)}$ such that the point-location learner of \citet{HKLM20}, with the modifications of \cref{app:robust} and given an $\eta^*$-subspace-relative oracle, labels every point of $X$ correctly with probability at least $1-\delta'$ using $O(n\log^3n\log(N/\delta'))$ queries. Every instance $(V_0,T)$ it queries has $V_0$ containing a point of $X$, so $P_{V_0}w^\star\ne0$. (With exact queries, \citet{HKLM20} use $O(n\log^2n\log(N/\delta'))$.)
\end{proposition}

\ifextendedproofs
\subsection{Proof of robust point location}\label{sec:robust-main-proof}
We prove \cref{prop:robust-pl} by tracking the errors in comparison, selection, dimension reduction, and reconstruction. The oracle-free isotropic reduction and the final boosting theorem are results of \citet{HKLM20}, listed in \cref{tab:hklm}. The local probability estimates and pseudocode are given in \cref{app:robust}.

\begin{proof}[Proof of \cref{prop:robust-pl}]
Fix a current instance of dimension $d\ge2$. Normalize its transformed target to a unit vector and write $m(x)$ for its inner product with $x$. The normalized sample points have unit norm. An $\eta$-subspace-relative oracle in these coordinates answers correctly whenever $|m(q)|>\eta\|q\|$.

\emph{Comparison and selection.} To compare $|m(a)|$ and $|m(b)|$, query their signs $\sigma_a,\sigma_b$ and the sign of $\sigma_aa-\sigma_bb$. Suppose $|m(a)|>|m(b)|+2\eta(\|a\|+\|b\|)$. Then $\sigma_a$ is correct. If $\sigma_b$ is correct, the final margin is $|m(a)|-|m(b)|$. Otherwise it is $|m(a)|+|m(b)|$. Both exceed $\eta\|\sigma_aa-\sigma_bb\|$. Interchanging $a,b$ proves correctness whenever the absolute margins differ by more than $2\eta(\|a\|+\|b\|)$. A zero query can be answered arbitrarily.

Compare every pair among $r$ representatives and choose a representative with median win count. Write $i_0=\lfloor r/2\rfloor+1$. The lowest $i_0$ scorers have $\binom{i_0}{2}$ internal wins, so the selected score is at least $(i_0-1)/2$. Among the highest $k_0=r-i_0+1$ scorers, the selected candidate has the lowest score and hence at least $(k_0-1)/2$ losses. Counting the selected candidate itself, at least $r/4$ values lie below its value plus the comparator error, and at least $r/4$ lie above its value minus that error. This rank statement holds even when every comparison within the error band is chosen adversarially.

\emph{Parameters and the reference margin.} Let $\delta_0\in(0,1/2]$ be a local failure probability. Choose sufficiently large absolute constants $C_0,c_1$ and set
\[
\lambda=(d/\delta_0)^{c_1},\quad \ell=C_0d^3\lambda^4,\quad
L=\lfloor\log_2d\rfloor,\quad N_0=4\lambda\sqrt d,
\]
\[
\tau=(4\lambda\sqrt d)^{-1}(2\ell)^{-(L+1)},\quad
\alpha_0=(96\lambda d^{3/2})^{-1},\quad
\eta\le\frac{\alpha_0\tau^2}{16\ell N_0}.
\]
At level $i$, draw $r=\lceil800\ln(8\log_2(2d)/\delta_0)\rceil$ samples of size $z_i=\lfloor16d/2^i\rfloor$. For each sample $S$, use the Gaussian representative $x_S=\sum_{x\in S}g_xx$. If $A_S=(\sum_{x\in S}m(x)^2)^{1/2}$, then $m(x_S)$ is Gaussian with variance $A_S^2$ and $\E\|x_S\|^2=|S|$. Gaussian small-ball and Markov bounds give
\[
\lambda^{-1}A_S\le |m(x_S)|\le\lambda A_S,
\qquad \|x_S\|\le\lambda\sqrt{|S|}
\]
except with probability $3/\lambda$. Reject observed norms above $N_0$.

The rank statement and a Dvoretzky--Kiefer--Wolfowitz (DKW) error of $1/20$ imply that the selected value $\widetilde M_i$ satisfies
\[
\Pr\{A_S\ge(\widetilde M_i-\bar\eta)/\lambda\}\ge1/10,
\qquad
\Pr\{A_S\le\lambda(\widetilde M_i+\bar\eta)\}\ge1/10,
\quad \bar\eta=4\eta N_0.
\]
These inequalities hold simultaneously over levels with the probability stated in \cref{lem:structure}. To see why the first scale is bounded away from zero, $1/4$-isotropy implies $\E m(x)^2\ge3/(4d)$. Since $|m(x)|\le1$, a sample of $16d$ points has $A_S\ge1/(2\sqrt d)$ except with probability at most $e^{-6}$. The lower representative bound and the lower empirical quantile therefore imply $\widetilde M_0\ge1/(4\lambda\sqrt d)$.

Test adjacent levels by comparing $\widehat x_i$ with $\ell\widehat x_{i+1}$. Its error is at most $2\eta(1+\ell)N_0\le\tau/2$. Before the first declared gap,
\[
\widetilde M_{i+1}\ge\frac{\widetilde M_i-\tau/2}{\ell}
\ge\frac{\widetilde M_i}{2\ell}.
\]
Thus the chosen reference has margin $M\ge\tau>\eta N_0$ and its queried sign is exact. If a gap is found at $i^*$, the two tail inequalities above and \cref{lem:norm-tail} give
\[
\Pr\{|m(x)|\ge t\}\ge k/d,\qquad
\Pr\{|m(x)|\le t/\varsigma\}\ge1-50k/d,
\]
where $k=2^{i^*}/160$, $t=M/(8\lambda\sqrt d)$ and $\varsigma=\ell/(24\lambda^2\sqrt d)$. If there is no gap, the last sample size lies between $16$ and $31$, giving mass at least $1/320$ with margin at least $M/(12\lambda)$. The exact tail calculations are in \cref{lem:structure}.

\emph{Dimension reduction under incorrect near-boundary answers.} In the gap case with $50k\le d/2$, put $X_{\mathrm{small}}=\{x:|m(x)|\le t/\varsigma\}$ and $z=\lfloor d/(50k)\rfloor$. Accept a sampled subset by comparing its representative with $cx_{\mathrm{ref}}$, where
\[
c=6\lambda^2\sqrt z/\ell,
\qquad |m(cx_{\mathrm{ref}})|=2\lambda\sqrt z\,t/\varsigma.
\]
On the representative good event, every subset contained in $X_{\mathrm{small}}$ has representative margin at most half this threshold. The comparator error fits within the remaining half, so it is accepted. Conversely, every accepted subset has representative margin at most three halves of the threshold. Each of its points consequently has margin at most
\[
\bar m=18\lambda^3\sqrt d\,M/\ell.
\]
Widening the acceptance threshold in this way keeps every subset of small-margin points while still controlling every other accepted point.

Let $\Sigma_{\mathrm{far}}=\E[xx^\top\mathbf1_{x\notin X_{\mathrm{small}}}]$. Its trace is at most $50k/d$. On a subspace $W$ of dimension at least $d-100k$, its quadratic form is at most $1/(2d)$, while isotropy gives
\[
\E[xx^\top\mathbf1_{x\in X_{\mathrm{small}}}]\big|_W\succeq I_W/(4d).
\]
A trial subset lies in $X_{\mathrm{small}}$ with probability at least $1/4$. With $R=C(k+1)\log(d/\delta_0)$ independent trials, the subsets contained in $X_{\mathrm{small}}$ therefore give enough independent conditional draws for a matrix Chernoff bound. As proved in \cref{lem:dimreduce}, their covariance, normalized by $Rz$, is at least $I_W/(64d)$ with the required probability. All extra accepted outer products are positive semidefinite. The full accepted covariance $\widehat C$ consequently has at least $d-100k$ eigenvalues above $1/(128d)$. Let $V$ be their span. Since every contributing point has margin at most $\bar m$ and the total weight is at most one,
\[
\|\widehat Cw'\|\le\bar m,
\qquad
|m(v)|=|\langle\widehat Cw',\widehat C^{-1}v\rangle|
\le128d\bar m\|v\|\quad(v\in V).
\]
Here the inverse is restricted to $V$. Thus $\dim V^\perp\le100k$ and $|m(v)|\le C_2\lambda^3d^{3/2}M\|v\|/\ell$ on $V$. If dimension reduction is not used, take $V=\{0\}$. In that case either $k>d/100$ or the no-gap case applies with $k=d/320$, so there are still at most $320(k+1)$ remaining coordinates.

\emph{Reconstruction with a supplied error tolerance.} For an orthonormal basis $w_i$ of $V^\perp$, write $\xi_i=m(w_i)/M$. The exact reference sign $\sigma_{\mathrm{ref}}$ lets the learner query
\[
m(w_i-\alpha\sigma_{\mathrm{ref}}x_{\mathrm{ref}})=M(\xi_i-\alpha).
\]
Since $M\ge\tau$, every $\xi_i$ lies in $[-1/\tau,1/\tau]$. Queries in that interval have norm at most $2N_0/\tau$. Hence answers are correct outside an interval of radius $2\eta N_0/\tau^2\le\alpha_0/4$ around $\xi_i$. Bisection to width $\alpha_0/2$ returns $\widehat\xi_i$ with error at most $\alpha_0/2$, using $O(\log d\log(d/\delta_0))$ queries. This follows by maintaining lower and upper endpoints within that tolerance of the true root, as in \cref{lem:fbs}.

For a unit point $y=v+\sum_i b_iw_i$, set $\widehat E=\sum_i b_i\widehat\xi_i$. Cauchy--Schwarz and the residual bound give
\[
\left|\frac{m(y)}M-\widehat E\right|
\le\frac{C_2\lambda^3d^{3/2}}\ell+\frac{\sqrt d\alpha_0}{2}
\le\varepsilon_{\mathrm{rec}},\qquad \varepsilon_{\mathrm{rec}}=(48\lambda d)^{-1}.
\]
Label only points with $|\widehat E|\ge2\varepsilon_{\mathrm{rec}}$. Every such label is correct. All points of margin at least $t$ are labeled in the gap case, and all points of margin at least $M/(12\lambda)$ are labeled in the no-gap case. The coverage is therefore at least $k/d$. Selection costs $O(r^2\log d)$ queries, dimension reduction costs $O((k+1)\log(d/\delta_0))$, and reconstruction costs $O((k+1)\log d\log(d/\delta_0))$. Their total is $O((k+1)\log d\log^2(d/\delta_0))$. By the probability accounting in \cref{lem:structure,lem:dimreduce,lem:isolearn}, every assigned label is correct and the coverage bound holds, except with probability $\delta_0$.

\emph{A uniform tolerance and a deterministic query budget.} Set $\delta_0=n^{-C}$ for a sufficiently large absolute $C$ and use the minimum of the displayed tolerances over $2\le d\le n$. The explicit formulas give $\eta_n^*=n^{-O(\log n)}$. The oracle-free isotropic reduction gives an instance containing at least $d/n$ of the current weight, so a successful local call removes at least $k/n$ of that weight with at most $c(k+1)\log^3n$ queries. Every instance contains an original sample point. Thus its projected target is nonzero, as required by the subspace-relative oracle.

Run these calls on residual weights with a deterministic budget $Q_W=C_Wn\log^3n$. Discard any call interrupted by the budget. A call costs at most a constant times $n\log^3n$, and $k\ge1/160$ implies $k+1\le161k$. Taking $C_W$ sufficiently large ensures that, unless all points have already been labeled, the completed calls satisfy $\sum_i k_i\ge5n$. On their joint success event the remaining mass is at most $\prod_i(1-k_i/n)\le e^{-5}<0.01$. There are at most $Q_W$ calls. Conditional failure bounds and a union bound give total failure probability at most $2Q_Wn^{-C}\le0.01$ after choosing $C$ sufficiently large. This yields a weak learner with a deterministic $O(n\log^3n)$ query budget, reliability $0.99$, and coverage $0.99$ with probability $0.99$ for every weighting of $X$.

The oracle-free boosting reduction of \citet[Lemma~5.3]{HKLM20}, with its hypotheses listed in \cref{tab:hklm}, now labels all $N$ points with failure probability at most $\delta'$ using $O(n\log^3n\log(N/\delta'))$ queries. Only the number of boosting rounds depends on $N$ and $\delta'$, so the tolerance remains $\eta_n^*$. One-dimensional instances require one exact sign query of relative margin one. This completes the proof.
\end{proof}

\fi

\subsection{Uniform label complexity}
\begin{theorem}[Pool simulation under hyperplane-null angular laws]\label{thm:main}
Let $n\ge2$ and let $D$ assign zero mass to the origin. Suppose its angular law $\mu$ satisfies $\mu(H\cap\Sph)=0$ for every linear hyperplane $H$ and has symmetric full support (\cref{sec:angular-laws}), for instance full support. Then homogeneous halfspaces are learnable from an i.i.d.\ pool to error $\eps$ with probability at least $1-\delta$ using
\[
O\Bigl(n\log^3n\,\log\frac{n}{\eps\delta}\Bigr)
\]
labels, uniformly over $D$ and the target. No moment, concavity, isotropy, or bounded angular-density assumption is required. The number of unlabeled examples is finite almost surely but may depend on $D$.
\end{theorem}
\begin{proof}
Take $N=C\frac{n\log(1/\eps)+\log(1/\delta)}{\eps}$. Use an independent unlabeled batch to compute $\theta_0$ by \cref{lem:floor}, and then draw a passive sample $\mathcal X$ of size $N$. On the event of \cref{cor:passive-floor}, every passive point has normalized target margin at least $\theta_0$.

Run the point-location learner of \cref{prop:robust-pl} on $\mathcal X$. On the event above no point of $\mathcal X$ lies on the target hyperplane, and every instance $(V_0,T)$ it queries has $V_0$ containing a point of $\mathcal X$, whose normalized target margin is at least $\theta_0$. When it requests $z'$ in instance $(V_0,T)$, wait for a fresh pool point $p$ with $\angle(p,T^{-1}z')\le\gamma$ or $\angle(-p,T^{-1}z')\le\gamma$, where $\gamma=\eta^*\theta_0/\kappa(T)$, and return its label, negated in the second case. If $C$ is the angular $\gamma$-cap around $T^{-1}z'$, each draw lands in the double cap $C\cup-C$ with probability $\mu(C\cup-C)\ge\bar\mu(C)\ge a_{\bar\mu}(\gamma)>0$, so symmetric full support makes this waiting time finite almost surely. By \cref{lem:pool-oracle} and the remark after it, the answers form an $\eta^*$-subspace-relative oracle. \Cref{prop:robust-pl} therefore labels all of $\mathcal X$ correctly with probability at least $1-\delta/4$ using $O(n\log^3n\log(N/\delta))$ labels. Return any homogeneous halfspace consistent with those labels. The realizable VC bound makes its error at most $\eps$ with probability at least $1-\delta/4$. Combining failure events and using $\log(N/\delta)=O(\log(n/(\eps\delta)))$ proves the theorem.
\end{proof}

The two angular assumptions play separate roles, and they can be placed on different distributions. Hyperplane-nullness is needed where error is measured and the margin floor is computed. Symmetric full support is needed only where answers are drawn (proof in \cref{app:baseline-proofs}).
\begin{theorem}[Separate evaluation and query laws]\label{thm:two-laws}
Let $n\ge2$. Suppose the learner can draw independent unlabeled samples from two distributions on $\R^n\setminus\{0\}$: an evaluation law $D$ whose angular law is hyperplane-null, and a query law $\mathcal Q$ whose angular law has symmetric full support. Labels may be requested only for points drawn from $\mathcal Q$, and error is measured under $D$. Then homogeneous halfspaces are learnable to $D$-error $\eps$ with probability at least $1-\delta$ using $O(n\log^3n\log\frac n{\eps\delta})$ labels, uniformly over $D$, $\mathcal Q$, and the target. The law $\mathcal Q$ may be atomic or charge the target hyperplane, and no absolute-continuity relation between $D$ and $\mathcal Q$ is needed. Both unlabeled counts are finite almost surely.
\end{theorem}
\Cref{thm:main} is the case $\mathcal Q=D$.

Lifting via $x\mapsto(x,1)$ makes an affine halfspace homogeneous. Because only the symmetrized angular law matters, the theorem then covers general halfspaces under every full-support marginal (proof in \cref{app:baseline-proofs}).
\begin{corollary}[General halfspaces]\label{cor:affine}
Let $d\ge1$ and let $D'$ be a distribution on $\R^d$ with support $\R^d$ that gives zero mass to every affine hyperplane, for instance any nondegenerate Gaussian or any everywhere-positive density. Halfspaces $x\mapsto\sign(\langle a,x\rangle+b)$ are learnable from an i.i.d.\ pool to error $\eps$ with probability $1-\delta$ using $O\bigl((d+1)\log^3(d+1)\log\frac{d+1}{\eps\delta}\bigr)$ labels, uniformly over all such $D'$ and all targets.
\end{corollary}

\subsection{A quantitative unlabeled-sample bound}
\Cref{thm:main} bounds labels uniformly. Its unlabeled cost can still be quantified in terms of the angular cap-mass function. In the language of splitting indices \citep{Das05,Han14}, a \emph{splitter} is a query that separates many pairs of candidate hypotheses. The function $a_\mu$ measures the \emph{availability} of approximate splitters under the unlabeled distribution, while point location determines their \emph{informativeness}.

\begin{theorem}[Pathwise unlabeled cost]\label{thm:unlabeled}
Answer the queries from a dedicated independent stream of unlabeled examples, and let point location use only accepted points and labels, not rejected draws or waiting lengths. Condition on the margin-floor batch, passive sample, internal random seed, and a successful transcript of requested caps, accepted points, and labels. Suppose the execution issues $Q$ synthesized queries. For the $j$th query let $T_j$ be the current transform and $\gamma_j=\eta^*\theta_0/\kappa(T_j)$. If each query is realized by drawing fresh unlabeled examples until one falls in the corresponding double cap $C_j\cup-C_j$, where $C_j$ is the angular $\gamma_j$-cap around the requested direction, then for every $\zeta\in(0,1)$, with conditional probability at least $1-\zeta$ the number of additional unlabeled draws is at most
\[
\sum_{j=1}^Q\Bigl\lceil\frac{\log(Q/\zeta)}{a_{\bar\mu}(\gamma_j)}\Bigr\rceil.
\]
In particular, if $\gamma_j\ge\Gamma$ throughout the execution, it is at most $O\bigl(\frac{Q}{a_{\bar\mu}(\Gamma)}\log\frac Q\zeta\bigr)$. Adding the $M_0$ margin-floor samples and the $N$ passive examples gives the total unlabeled-sample bound.
\end{theorem}

\begin{remark}[Scope of the unlabeled bound]\label{rem:pathwise}
\emph{Controlled quantities.} The precisions $\gamma_j=\eta^*\theta_0/\kappa(T_j)$ depend on the condition numbers of the data-dependent isotropic transforms (\cref{sec:pointloc}), which we do not bound a priori. So \cref{thm:unlabeled} is pathwise. The unlabeled cost is finite almost surely, and therefore has finite quantiles for each fixed marginal and target. We do not derive explicit quantile bounds from $\mu$ alone, or establish that its expectation is finite. The margin-floor batch grows polynomially (not logarithmically) in $1/\delta$.

\emph{Cost of this simulation.} For $0<\gamma<\pi/2$, waiting for a point of a $\gamma$-cap $C$ or its reflection costs $1/\mu(C\cup-C)=1/(2\bar\mu(C))$ draws in expectation, at most $1/(2a_{\bar\mu}(\gamma))$. This bounds the cost of the cap-based simulation in the worst direction, and the execution need not request the least accessible direction. Averaging over directions gives $a_{\bar\mu}(\gamma)\le\sigma\{u:\angle(u,q)<\gamma\}\asymp_n\gamma^{n-1}$ for every $q$. For the uniform law $\sigma$, where every direction is equally accessible, $\gamma_j\le\eta^*\theta_0$ with $\eta^*=n^{-O(\log n)}$ and $\theta_0\le\eps\delta/n$, so every query waits $(n/(\eps\delta))^{\Omega(n)}$ draws in expectation, whereas margin-based learners use $\tilde O(n/\eps)$ unlabeled examples there \citep{BL13}.

\emph{Lower bounds for every learner.} No unlabeled budget fixed in advance works for all admissible marginals (\cref{prop:finite-budget}). For general halfspaces under Gaussian marginals, where \cref{cor:affine} applies, the lower bound of \citet{DKM24} forces every learner with $\tilde O(d\log(1/\eps))$ labels to inspect $\exp(\tilde\Omega(1/\eps))$ unlabeled examples on their instances (\cref{sec:related}).
\end{remark}

\subsection{Comparison with directional search}\label{sec:baseline}
The natural alternative to \cref{thm:main}, \emph{directional search}, approximates membership queries at a fixed angular resolution. That resolution is set by the \emph{anti-concentration scale}
\[
\rho_\mu^{-1}(\eps)=\sup\{\theta\in(0,1]:\rho_\mu(\theta)\le\eps\},
\]
the generalized inverse of the band mass, which is positive for every $\eps>0$ by \cref{lem:anticonc}. Proofs are in \cref{app:baseline-proofs}.

\begin{proposition}[Directional search at the anti-concentration scale]\label{prop:baseline}
Let $n\ge2$ and let $D$ have no atom at the origin and an angular law $\mu$ that is hyperplane-null with symmetric full support. There is a pool learner that, with probability at least $1-\delta$, outputs a homogeneous halfspace of error at most $\eps$ using
\[
O\Bigl(n\log\frac{n}{\rho_\mu^{-1}(\eps/16)}\Bigr)
\]
labels. It inspects $M_1=O\bigl((n\log(n/\eps)+\log(1/\delta))/\eps\bigr)$ unlabeled examples to choose a precision $\hat\theta\in[\rho_\mu^{-1}(\eps/16)/2,\,\rho_\mu^{-1}(\eps)]$. With probability at least $1-\delta/2$ it then inspects at most $Q\lceil\log(2Q/\delta)/a_{\bar\mu}(c\hat\theta/n^{3/2})\rceil$ further unlabeled examples, where $Q=O(n\log(n/\hat\theta))$ is its number of label requests and $c>0$ is an absolute constant. It uses neither a passive sample nor point location.
\end{proposition}
\Cref{cor:bounded} specializes this bound to bounded one-dimensional densities, and \cref{cor:combined} runs whichever of the two learners has the smaller label bound. The scale $\rho_\mu^{-1}$ cannot be removed from the label count of directional search by a change of coordinates.

\begin{exmp}[Flat Gaussian components]\label[exmp]{ex:mixture}
For $n\ge2$, $0<b<1$ and $0<\eps<1/6$, let $D_b=(1-6\eps)\mathcal N(0,I_n)+3\eps \mathcal N(0,\Sigma_1)+3\eps \mathcal N(0,\Sigma_2)$ with $\Sigma_i=I_n-(1-b^2)e_ie_i^\top$. \Cref{thm:main} applies to $D_b$ with a label bound independent of $b$, while $\rho_{\mu_A}^{-1}(\eps)\le5b$ for the angular law $\mu_A$ of $AX$ under every invertible $A$. So the label count of directional search grows like $n\log(1/b)$ in every linear coordinate system. This lower bound applies only to directional search.
\end{exmp}

\section{When the angular assumptions or a finite pool are not enough}\label{sec:lower}
The positive theorem uses two angular assumptions for two different reasons. Hyperplane-nullness gives anti-concentration near every great subsphere, and symmetric full support places points near every desired query direction or its antipode. \Cref{sec:necessity} shows that each is necessary for a general equivalence, even with unlimited unlabeled data. Neither is necessary for every individual distribution. Many marginals with smaller angular support, or with hyperplane mass, remain easy. \Cref{sec:affine-core} then shows that even with both assumptions, the amount of unlabeled data can be forced to be arbitrarily large.

\subsection{Necessity of the angular assumptions}\label{sec:necessity}
Both constructions place mass near the points $p_i=(t_i,t_i^2,1)$, $t_i=i/(m+1)$, on the moment curve $t\mapsto(t,t^2,1)$. For each $i$ some homogeneous halfspace labels $p_i$ negative and every other $p_j$ positive, so each $p_i$ can only be informative about its own alternative target, while membership queries off the support can binary-search over $i$. Proofs are in \cref{app:necessity}, and \cref{tab:necessity} summarizes the role of each assumption.

\begin{proposition}[Symmetric full support is necessary]\label{prop:need-support}
For every $m\ge2$ there are a distribution $D_m$ on $\R^3$ and $m+1$ homogeneous halfspaces such that, at accuracy $\eps=1/(4m)$ and success probability $3/4$:\looseness-1
\begin{enumerate}[label=(\roman*)]
\item every pool learner needs at least $m/2$ labels in the worst case, even with unlimited unlabeled examples, knowledge of $D_m$, and improper output,
\item a membership-query learner identifies the target exactly with at most $1+\lceil\log_2m\rceil$ labels.
\end{enumerate}
Moreover $D_m$ may be taken to have a smooth density (hence a hyperplane-null angular law), and to be centrally symmetric. Only symmetric full support fails.
\end{proposition}
\Cref{prop:need-support} is the distributional form of the classical one-point-per-target obstruction \citep{Das04}. It shows that hyperplane-nullness does not suffice even with a smooth, symmetric density.

\begin{proposition}[Hyperplane-nullness is necessary]\label{prop:need-null}
For every $m\ge2$ there are a full-support distribution $D_m$ on $\R^4$ and an infinite family of homogeneous halfspaces, containing arbitrarily small tilts of one target, such that, for $\eps=1/(16m)$:
\begin{enumerate}[label=(\roman*)]
\item membership queries achieve error at most $\eps$ using $1+\lceil\log_2m\rceil$ labels,
\item any pool learner that terminates almost surely and succeeds with probability at least $3/4$ for every target uses at least $m/2$ labels in expectation on some target, even with unlimited unlabeled data.
\end{enumerate}
\end{proposition}
The construction places mass on the target hyperplane and uses arbitrarily small tilts (\cref{rem:tilts}). Hyperplane-nullness excludes such mass, and this is what makes a target-independent margin floor possible.

\paragraph{Lifted affine problems.} Lifting an affine halfspace in $\R^{n-1}$ via $x\mapsto(x,1)$ makes it homogeneous and confines all directions to an open hemisphere. Since $h_w(-x)=-h_w(x)$ whenever $\langle w,x\rangle\ne0$, only the symmetrized angular law matters, and for a lifted marginal it has full support exactly when the original marginal has full support on $\R^{n-1}$ (\cref{cor:affine}). Marginals with bounded or otherwise incomplete support, such as the uniform ball, leave some neighborhood of the equator uncharged. \Cref{prop:need-support} is an explicit instance, and \cref{thm:affine-core} below shows that the same obstruction can reappear inside a high-probability core of a full-support law when the unlabeled budget is bounded.

\subsection{Finite unlabeled budgets}\label{sec:affine-core}
Even when both angular assumptions hold, no unlabeled budget fixed in advance suffices for all marginals, and the label cost then returns to the passive rate.

\begin{proposition}[A fixed unlabeled budget forces passive label rates]\label{prop:finite-budget}
Let $n\ge3$. For every $U:(0,1/32]\to\mathbb N$ and every $0<\eps\le1/32$ there is a distribution $D$ on $\R^n$ with a smooth, everywhere-positive, centrally symmetric density (so its angular law is hyperplane-null with full support) for which the following holds. Every possibly randomized pool learner that inspects at most $U(\eps)$ unlabeled examples, requests at most $q$ labels, and outputs a hypothesis of error at most $\eps$ with probability at least $2/3$ for every homogeneous target satisfies $q\ge1/(10\eps)$. Membership queries need only $\tilde O(n\log(1/\eps))$ labels under every marginal.
\end{proposition}
The hard law mixes the smoothed symmetric bumps of \cref{prop:need-support} with a Gaussian component of mass $u=1/(100\max\{U(\eps),1\})$, which gives full support. With probability $0.99$ the first $U(\eps)$ examples all fall in the bumps, where each label is informative about only one target (proof in \cref{app:deferred}). Within the symmetric $s$-concave class, \cref{thm:sym-pool} gives the same bound with a more delicate construction. \Cref{prop:finite-budget} settles the question for the class of \cref{thm:main}. The following refinement is needed only to locate thresholds inside structured classes such as $s$-concave laws, whose members cannot carry isolated bumps.

Let $\nu_d$ denote the uniform probability measure on the Euclidean unit ball $B_d\subset\R^d$.

\begin{definition}[Affine-core richness]\label{def:affine-core}
A class $\mathcal D_n$ of distributions on $\R^n$ is affine-core rich if for every $u>0$ there are $D\in\mathcal D_n$, a measurable set $G$, and a measurable map $\Pi:G\to B_{n-1}$ such that (i) $D(G)\ge1-u$, (ii) conditional on $G$, $\Pi(X)$ is exactly $\nu_{n-1}$, (iii) for every affine halfspace $h$ on $B_{n-1}$ there exists a homogeneous halfspace $H$ on $\R^n$ with $H(x)=h(\Pi(x))$ for every $x\in G$.
\end{definition}
Almost all data look like an affine halfspace problem in one lower dimension, and the behavior outside the core is unrestricted.

\begin{theorem}[Fixed-budget lower bound for affine-core-rich classes]\label{thm:affine-core}
Let $n\ge3$ and let $\mathcal D_n$ be affine-core rich. There are constants $c_n,\eps_n>0$ such that for every function $U:(0,\eps_n]\to\mathbb N$ and every $0<\eps\le\eps_n$, there exists $D\in\mathcal D_n$ for which the following holds. Any possibly randomized, computationally unbounded pool learner that (i) inspects at most $U(\eps)$ i.i.d.\ unlabeled examples from $D$, (ii) requests at most $q$ labels among them, and (iii) for every homogeneous target outputs a hypothesis of $D$-error at most $\eps$ with probability at least $2/3$, must satisfy $q\ge c_n\eps^{-(n-2)/n}$.
\end{theorem}
\noindent\emph{Proof idea} (full proof in \cref{app:deferred}). Choose the core mass so that the first $U(\eps)$ stream elements all lie in the core with probability $0.99$. Pack $M\gtrsim\eps^{-(n-2)/n}$ disjoint caps of mass $4\eps$ in the uniform ball (\cref{lem:caps}). Each cap is the negative region of an affine, hence (on the core) homogeneous, halfspace, and these targets are pairwise $7\eps$-separated. Every sampled point lies in at most one cap, so each label reveals whether the target is one specific candidate. Along the \emph{transcript} (the sequence of queried points and answers) in which every answer is positive, $q$ labels rule out at most $q$ of the $M$ candidates.

\noindent In particular, since a polynomial-time pool algorithm inspects only polynomially many examples, no algorithm for the class of \cref{thm:main} (by \cref{prop:finite-budget}), or for any affine-core-rich class, can have both a distribution-uniform polynomial running time and label complexity polylogarithmic in $1/\eps$.\phantomsection\label{cor:comp}

\section{Isotropic \texorpdfstring{$s$}{s}-concave label complexity phase diagram}\label{sec:sconcave}
\begin{definition}[$s$-concave densities, {\citealp{Bor75,BZ17}}]\label{def:sconcave}
For $s<0$, a density $f$ on $\R^n$ is \emph{$s$-concave} if for all $x,y$ in its support and $\lambda\in[0,1]$,
\[
f(\lambda x+(1-\lambda)y)\ \ge\ \bigl(\lambda f(x)^s+(1-\lambda)f(y)^s\bigr)^{1/s},
\]
equivalently, $f^s$ is convex on the support (and the support is convex). For $s=0$ the condition is read as log-concavity, $f(\lambda x+(1-\lambda)y)\ge f(x)^\lambda f(y)^{1-\lambda}$.
\end{definition}
An $s$-concave density is $s'$-concave for every $s'\le s$, so the condition weakens as $s$ decreases. Negative $s$ admits polynomial tails. For example, $f(x)\propto(1+\|x\|)^{-p}$ is $(-1/p)$-concave because $f^{-1/p}\propto1+\|x\|$ is convex, and the multivariate Student-$t$ law with $r$ degrees of freedom is $(-2/(r+n))$-concave. A law is \emph{isotropic} if it has mean zero and identity covariance. We always center halfspaces at the mean, so in isotropic position they are homogeneous. Let $\Dns$ be the full-dimensional isotropic laws with an $s$-concave density (the \emph{general class}) and $\Dsym\subseteq\Dns$ its \emph{centrally symmetric} subclass ($f(x)=f(-x)$, not necessarily rotationally symmetric). The \emph{slack} $-1/s-(n+2)$ is positive exactly when the extreme tail $(1+\|x\|)^{1/s}$ allowed by $s$-concavity still has a finite second moment. \citet{BZ17} developed sample-based active learners for $-1/(2n+3)\le s\le0$. Logarithmic disagreement is defined in \cref{sec:disagreement}. \ifextendedproofs The sharp intermediate exponent is proved in \cref{sec:exponent-main}, and the remaining proofs are in \cref{app:sconcave,app:exponent}.\fi{} \Cref{tab:before-after} compares each row of \cref{fig:phase} with what was previously established, and \cref{tab:regimes} gives the status by regime.

\begin{figure}[t]
\centering
\resizebox{\linewidth}{!}{%
\begin{tikzpicture}[x=1cm,y=0.8cm,font=\small]
  \def\xa{1.6}\def\xb{4.4}\def\xc{7.2}\def\xd{10.0}\def\xe{12.4}
  \newcommand{\phasebar}[6]{%
    \fill[#4] (#1,#3-0.3) rectangle (#2,#3+0.3);
    \node[anchor=east,align=right,font=\footnotesize] at (-0.2,#3) {#5};
    \node[font=\footnotesize] at (#1/2,#3) {#6};}
  \draw[-{Latex}] (0,0) -- (13.1,0) node[right]{$s$};
  \foreach \x/\l in {\xa/{-\tfrac1n},\xb/{-\tfrac1{n+1}},\xc/{-\tfrac1{n+2}},\xd/{-\tfrac1{2n+3}},\xe/{0}}{
    \draw (\x,0.12)--(\x,-0.12) node[below]{$\l$};
    \draw[gray!50,densely dotted] (\x,0.15)--(\x,4.5);}
  \node[below] at (0.15,-0.12) {$-\infty$};
  \phasebar{0}{\xe}{4.0}{blue!30}{Unlimited pool}{}
  \node[font=\footnotesize] at (5.5,4.0) {every finite $s$ (\cref{cor:all-s})};
  \phasebar{\xc}{\xe}{3.0}{orange!45}{Balcan--Zhang tools}{no uniform density/band bounds (\cref{prop:struct-sharp})}
  \phasebar{\xb}{\xe}{2.0}{green!45}{Disagreement and\\fixed-pool labels (general)}{}
  \fill[green!18] (\xa,1.7) rectangle (\xb,2.3);
  \node[font=\footnotesize] at (\xa/2,2.0) {$\Theta(\frac1\eps)$};
  \node[font=\footnotesize] at ({(\xa+\xb)/2},2.0) {$\Theta(\eps^{-(n-2)/n})$};
  \node[font=\footnotesize] at ({(\xb+\xe)/2},2.0) {$O(\log\frac1\eps)$, polylog labels};
  \phasebar{\xa}{\xe}{1.0}{green!45}{Disagreement and\\fixed-pool labels (symmetric)}{$\Theta(\frac1\eps)$}
  \node[font=\footnotesize] at ({(\xa+\xe)/2},1.0) {$O(\log\frac1\eps)$, polylog labels};
  \draw[very thick,red!70!black] (\xd,4.55)--(\xd,0.15);
  \draw[red!70!black,-{Latex}] (\xd,4.75)--(\xe+0.2,4.75);
  \node[red!70!black,font=\footnotesize,anchor=east] at (\xd-0.1,4.75) {previous sufficient range $s\ge-\tfrac1{2n+3}$};
\end{tikzpicture}}
\caption{Realizable regimes for isotropic $s$-concave marginals in dimension $n\ge3$ (axis not to scale). In the two lower rows, darker green marks class-uniform logarithmic disagreement with polylogarithmic labels, and lighter green the sharp polynomial rate. Each $\Theta$ entry is the sharp worst-case rate of the disagreement coefficient and, up to logarithmic factors and at constant confidence, of the labels needed with an unlabeled budget fixed in advance for the class. In the regimes marked $O(\log\frac1\eps)$ we prove only upper bounds. Label upper bounds go through CAL, and both lower bounds come from one packing lemma (\cref{lem:star}). The red line marks the range $s\ge-1/(2n+3)$ covered by \citet{BZ17}. An unlimited pool keeps near-optimal labels for every finite $s$. Only the first row uses \cref{thm:main}. %
}
\label{fig:phase}
\end{figure}

\paragraph{Two thresholds with different causes.} The threshold $-1/(n+1)$ concerns the reference point. Above it, a Gr\"unbaum-type inequality bounds the Tukey depth of the mean below, uniformly over the class (\cref{thm:disagree-gen}). At and below it, a tail of vanishing probability can move the mean to depth tending to zero while the bulk is unchanged (\cref{sec:rare-tail}). \Cref{prop:depth} shows that logarithmic disagreement persists for every $s>-1/n$ at reference points of uniformly positive depth, so for halfspaces through a deep point the general threshold disappears. The threshold $-1/n$ is intrinsic to the angular geometry. It persists at depth $1/2$ (\cref{thm:sym-dis}), it is where the level-set bound of \cref{lem:level-int} degenerates, and below it \cref{lem:bobkov} no longer guarantees convexity of $K_{n-1}$. We follow \citet{BZ17} in centering halfspaces at the mean. For $s\le-1/(n+2)$, members of the class have finite covariance only because their tails are lighter than the extremal $s$-concave tail, and all our lower-bound families are of this kind.

\begin{enumerate}[leftmargin=*,label=(\alph*)]
\item \emph{Unlimited pool} (\cref{cor:all-s}). A full-dimensional $s$-concave law has a positive density near its interior mean, so its centered angular law is absolutely continuous with full support, and \cref{thm:main} applies for every finite $s$.
\item \emph{Limits of prior techniques below $s=-\frac1{n+2}$} (\cref{thm:toolkit,prop:struct-sharp}). The tools of \citet{BZ17} bound band masses $\Pr[|\langle u,X\rangle|\le t]$ from both sides, disagreement from below by the angle between normals, disagreement outside a band, and conditional second moments in a band. Using the classical centroid-to-mode inequality of \citet{AGLLOPT12}, which bounds the density at the mean below by a constant fraction of its maximum, all of these hold with constants depending only on the slack whenever $s>-1/(n+2)$. At and below $-1/(n+2)$, truncations of $(1+\|x\|)^{-(n+2)}$ make the density and band upper bounds unbounded, even for rotationally symmetric laws, so the tools cannot hold with uniform constants.\looseness-1
\item \emph{General angular threshold} (\cref{thm:disagree-gen,prop:dis-blowup,cor:gen-pool}). For $s>-1/(n+1)$, integrating the density along rays defines a body $K=\{ru:u\in\Sph,\ 0\le r\le\varrho(u)\}$ with $\varrho(u)^n=n\int_0^\infty r^{n-1}f(ru)\,dr$, which is convex by a theorem of Ball and Bobkov (\cref{lem:bobkov}), and the angular law of $f$ is the direction law of a uniform point of $K$. Convex-body geometry gives $\theta_D(\eps)=O_s(\sqrt n\log(1/\eps))$. For $s\le-1/(n+1)$, a tail of vanishing probability that shifts the mean yields $\theta_D(\eps)=\Omega_n(\eps^{-(n-2)/n})$ and, via \cref{thm:affine-core}, the same finite-pool label lower bound.
\item $s\in\bigl(-\frac1n,-\frac1{n+1}\bigr]$ (\cref{thm:exp-upper,cor:exp-cal,cor:exp-match}). For $-1/n<s\le-1/(n+1)$, every $s$-concave $D$ (isotropic or not) and homogeneous target satisfy $D(\DIS(B_D(h^\star,r)))\le C_n(1+ns)^{-1}r^{2/n}$. So $\theta_D(\eps)=O_{n,s}(\eps^{-(n-2)/n})$, and CAL on a pool of $\tilde O(n/\eps)$ points fixed in advance uses $\tilde O_{n,s}(\eps^{-(n-2)/n})$ labels, matching the lower bounds. The proof has four steps. (1) \emph{Projective marginal.} If $(Z,T)$ is uniform on a convex body in $\{t\ge0\}$, then $(Z/T,1/T)$ has density $\propto v^{-(n+1)}$ on a convex set, hence is $(-1/(n+1))$-concave, so $Y=Z/T$ has a $(-1/n)$-concave density on $\R^{n-1}$. (2) \emph{Small-mass homogeneous halfspace regions.} Homogeneous halfspaces $\{\langle a,z\rangle+bt\ge0\}$ become affine halfspaces $\{\langle a,y\rangle+b\ge0\}$, so the union of all homogeneous halfspaces of mass at most $u$ becomes the small-mass halfspace region of $Y$. The classical bound $O_d(u^{2/(d+1)})$ on this region for a uniform convex body in $\R^d$ \citep{BarL88}, applied to the level sets of the density of $Y$, gives $O_n(u^{2/n})$. (3) \emph{Density level sets.} For a general $(-1/p)$-concave density $f$ with $p>n$, a halfspace of mass $u$ has mass at most $u/(tV(t))$ under the uniform law on $\{f\ge t\}$, whose volume is $V(t)$, and $\int t^{-\alpha}V(t)^{1-\alpha}dt\le C_n/(1+ns)$ because level sets grow at most like $t^{-n/p}$. (4) \emph{Disagreement.} On each side of the target hyperplane, the disagreement region of the error ball of radius $r$ is a small-mass homogeneous halfspace region at level $r/\omega_\pm$, where $\omega_\pm$ are the masses of the two sides. Homogeneity reduces caps to dimension $n-1$, which yields $2/n$ instead of $2/(n+1)$, and $s>-1/n$ is the condition that keeps the integral in step (3) finite. At reference points of uniformly positive depth the bound improves to $O(\log\frac1\eps)$ for all $s>-1/n$ (\cref{prop:depth}).
\item \emph{Symmetric angular threshold} (\cref{thm:sym-disagree,thm:sym-dis,thm:sym-pool}). For symmetric laws with $s>-1/n$, the body built from $(n-1)$st radial moments is still convex, and the Fradelizi--Gu\'edon--Pajor inequalities \citep{FGP14} compare it with the $n$th-moment body up to a constant, giving $\theta_D(\eps)=O_{n,s}(\log(1/\eps))$. For $s\le-1/n$, choose $m$ antipodal pairs $\pm v_i$ and halfspaces $h_i$ that disagree with a base target exactly near $\pm v_i$. \Cref{lem:radial-mixture} realizes an equal mixture of the angular laws of ellipsoids elongated along the $v_i$, up to small total variation, as the angular law of a symmetric $(-1/n)$-concave density. This gives $\theta_D(2/m)\ge3m/8$ and, for every unlabeled budget fixed in advance, $\Omega(1/\eps)$ labels, also for the general class.
\end{enumerate}

\ifextendedproofs
\subsection{The sharp general-class exponent}\label{sec:exponent-main}\label{sec:exponent-proof}
We prove the sharp general-class exponent in the intermediate range $-1/n<s\le-1/(n+1)$. All halfspaces in the learning statements are homogeneous with respect to the fixed origin, and concavity means concavity of the density (\cref{def:sconcave}). The upper bound needs no isotropy, centering, symmetry, full angular support, or finite moments. Write $s=-1/p$, so $s>-1/n$ means $p>n$.

\providecommand{\expsubsection}{}\renewcommand{\expsubsection}{\subsubsection}
\begin{theorem}[Uniform disagreement upper bound]\label{thm:exp-upper}
Let $n\ge3$ and $p>n$, and let $D$ have a probability density $f$ on $\R^n$ such that $f^{-1/p}$ is convex on its convex support. There is $K_{n,p}\ge1$, depending only on $n$ and $p$, such that for every homogeneous target $h^\star$ and $0<r\le1$,
\[
D\bigl(\DIS(B_D(h^\star,r))\bigr)\le K_{n,p}\,r^{2/n},\qquad\theta_D(\eps;h^\star)\le K_{n,p}\,\eps^{-(n-2)/n}\quad(0<\eps\le1).
\]
For $s=-1/p\in(-1/n,-1/(n+1)]$ one can take $K_{n,p}\le C_n/(1+ns)$.
\end{theorem}
Together with \cref{prop:dis-blowup} this gives the exact worst-case disagreement exponent $(n-2)/n$, and \cref{cor:exp-cal,cor:exp-match} give the matching label exponent under an unlabeled budget fixed in advance, up to logarithms. The proof first treats uniform convex bodies by a projective change of variables and then integrates over the level sets of $f$. Projecting $f$ itself would lose too much concavity near $s=-1/n$.

\expsubsection{Small-mass halfspace regions and integration over level sets}
Let $\nu$ be an absolutely continuous probability measure on $\R^d$ and $u>0$. The \emph{small-mass halfspace region} $W_\nu(u)$ is the union of all closed affine halfspaces with $\nu$-mass at most $u$,
\[
W_\nu(u)=\bigcup\{H:\ H\text{ a closed affine halfspace},\ \nu(H)\le u\}.
\]
This region is traditionally called the \emph{wet part}. We write $W^0_\nu(u)$ when the union is restricted to halfspaces whose boundaries pass through the origin. The union itself can have much more than $u$ probability mass, and the bounds below control how large it can be. These sets are measurable for the completed measure, being projections of Borel sets in the point and halfspace parameters, and since hyperplanes are null the choice of open or closed sides does not matter. Such regions for arbitrary probability measures are studied in \citet{BFGHR20}. We use the classical estimate for convex bodies \citep{BarL88,Bar08}. For $d\ge2$ there is $A_d\ge1$ such that for every convex body $K\subset\R^d$, with $\nu_K$ the uniform law on $K$,
\begin{equation}\label{eq:wet-body}
\nu_K(W_{\nu_K}(u))\le A_d\,u^{2/(d+1)}\qquad(u>0).
\end{equation}
The bound of \citet[Cor.~11.4]{Bar08} holds for $u$ below a dimensional threshold with constants depending only on $d$. Enlarging $A_d$ covers the remaining $u$, including $u\ge1$.

\begin{lemma}[A uniform level-set integral]\label{lem:level-int}
Let $p>d$ and let $f$ be a probability density on $\R^d$ with $f^{-1/p}$ convex and a finite attained maximum $f_{\max}$. Set $K_t=\{f\ge t\}$ and $V(t)=\mathrm{vol}_d(K_t)$. For $0<\alpha<1$,
\[
\int_0^{f_{\max}}t^{-\alpha}V(t)^{1-\alpha}\,dt\le L_{d,p,\alpha}:=\frac{2^{p(1-\alpha)}}{(1-\alpha)(1-d/p)}.
\]
\end{lemma}
\begin{proof}
Choose a mode $x_0$ and put $t_0=2^{-p}f_{\max}$. Since $\int f=1$, $V(t_0)\le2^p/f_{\max}$. For $t\le t_0$ let $v=(f_{\max}/t)^{1/p}\ge2$. If $x\in K_t$, convexity of $\phi=f^{-1/p}$ gives
\[
\phi\Bigl(x_0+\frac{x-x_0}{v-1}\Bigr)\le\Bigl(1-\frac1{v-1}\Bigr)f_{\max}^{-1/p}+\frac{t^{-1/p}}{v-1}=2f_{\max}^{-1/p}.
\]
Thus $K_t\subseteq x_0+(v-1)(K_{t_0}-x_0)$ and $V(t)\le(2^p/f_{\max})(f_{\max}/t)^{d/p}$. The same bound holds for $t\ge t_0$ by monotonicity of $V$. Substituting $t=f_{\max}y$, the powers of $f_{\max}$ cancel and
\[
\int_0^{f_{\max}}t^{-\alpha}V(t)^{1-\alpha}\,dt\le2^{p(1-\alpha)}\int_0^1y^{-\alpha-(d/p)(1-\alpha)}\,dy=L_{d,p,\alpha},
\]
which is finite because $p>d$ and $\alpha<1$.
\end{proof}

\begin{proposition}[Small-mass halfspace regions of $(-1/p)$-concave densities]\label{prop:wet-density}
For $d\ge2$, $p>d$ and every probability density $f$ on $\R^d$ with $f^{-1/p}$ convex, its law $\nu$ satisfies
\[
\nu(W_\nu(u))\le A_dL_{d,p,\alpha}\,u^\alpha,\qquad\alpha=\frac2{d+1},\qquad u>0.
\]
\end{proposition}
\begin{proof}
First suppose $f$ has a finite attained maximum $f_{\max}$. For $0<t<f_{\max}$ the superlevel set $K_t$ is convex with positive finite volume, hence bounded, and its closure is a convex body of the same volume. If $\nu(H)\le u$ then $t\,\mathrm{vol}_d(H\cap K_t)\le\int_Hf\le u$, so $H$ has mass at most $u/(tV(t))$ under the uniform law on $K_t$. By \eqref{eq:wet-body},
\[
\mathrm{vol}_d(K_t\cap W_\nu(u))\le A_dV(t)\Bigl(\frac u{tV(t)}\Bigr)^\alpha.
\]
The layer-cake formula $\nu(W)=\int_0^{f_{\max}}\mathrm{vol}_d(K_t\cap W)\,dt$ and \cref{lem:level-int} give the claim.

In general, for $0<c<\operatorname{ess\,sup}f$ put $g_c=\min(f,c)$, $Z_c=\int g_c$ and $f_c=g_c/Z_c$. Then $f_c^{-1/p}=Z_c^{1/p}\max(f^{-1/p},c^{-1/p})$ is convex, and $f_c$ has an attained finite maximum. As $c\uparrow\operatorname{ess\,sup}f$ (possibly $+\infty$), $Z_c\uparrow1$ and the laws $\nu_c$ of $f_c$ converge to $\nu$ in total variation. Since $\nu_c(H)\le\nu(H)/Z_c$ for every halfspace, $W_\nu(u)\subseteq W_{\nu_c}(u/Z_c)$. Applying the bound to $\nu_c$ and letting $c$ increase proves the claim. The same approximation works for any restricted family of halfspaces, in particular for $W^0$.
\end{proof}

\expsubsection{A projective estimate on one side of the origin}
\begin{lemma}[Small-mass homogeneous halfspace regions of a uniform convex body]\label{lem:cone-body}
For $n\ge3$ there is $A^0_n\ge1$ such that for every convex body $K\subset\{(z,t)\in\R^{n-1}\times\R:t\ge0\}$,
\[
\nu_K(W^0_{\nu_K}(u))\le A^0_n\,u^{2/n}\qquad(u>0).
\]
One may take $A^0_n=A_{n-1}L_{n-1,n,2/n}$.
\end{lemma}
\begin{proof}
The face $\{t=0\}$ is null. Use the bijection $(z,t)\mapsto(y,v)=(z/t,1/t)$ of $\{t>0\}$, with inverse $(y,v)\mapsto(y/v,1/v)$. The image $K'$ of $K\cap\{t>0\}$ is convex. Indeed, if $(y_i,v_i)$ correspond to $x_i\in K$, the preimage of $\lambda(y_1,v_1)+(1-\lambda)(y_2,v_2)$ is $(\lambda v_1x_1+(1-\lambda)v_2x_2)/(\lambda v_1+(1-\lambda)v_2)$, a convex combination of $x_1,x_2$. The inverse map has Jacobian of absolute value $v^{-(n+1)}$, so the transformed law has density
\[
F(y,v)=\frac{v^{-(n+1)}}{\mathrm{vol}_n(K)}\mathbf 1_{K'}(y,v).
\]
Since $F^{-1/(n+1)}$ is a positive multiple of $v$ on the convex set $K'$, $F$ is $(-1/(n+1))$-concave. Let $(Z,T)\sim\nu_K$. By the marginal rule (\cref{lem:marginals} with one coordinate integrated out, valid since $-1/(n+1)>-1$), the density of $Y=Z/T$ on $\R^{n-1}$ is $s'$-concave with $s'=\frac{-1/(n+1)}{1-1/(n+1)}=-\frac1n$. This needs only integrability and convex support, so the possible unboundedness of $K'$ is harmless. Apply \cref{prop:wet-density} in dimension $d=n-1$ with $p=n>d$. Its exponent is $2/(d+1)=2/n$.

Finally, on $\{t>0\}$ the homogeneous halfspace $\{\langle a,z\rangle+bt\ge0\}$ is exactly $\{\langle a,y\rangle+b\ge0\}$ in the new coordinates, and every affine halfspace of $\R^{n-1}$ arises this way. (Degenerate choices give the empty or full set and only matter for $u\ge1$, where the bound is trivial.) The map $(z,t)\mapsto z/t$ preserves masses and unions, which proves the claim.
\end{proof}

\begin{lemma}[Small-mass homogeneous halfspace regions of a one-sided density]\label{lem:cone-density}
Let $n\ge3$, $p>n$, and let $\nu$ have a density $f$ with $f^{-1/p}$ convex, supported in a closed halfspace through the origin. With $\alpha=2/n$,
\[
\nu(W^0_\nu(u))\le A^0_nL_{n,p,\alpha}\,u^\alpha\qquad(u>0).
\]
\end{lemma}
\begin{proof}
Rotate so that the supporting halfspace is the one in \cref{lem:cone-body}. If $f$ has an attained finite maximum $f_{\max}$, apply \cref{lem:cone-body} to the closure of every positive-volume superlevel set $K_t$, which lies in the supporting halfspace. A homogeneous halfspace of $\nu$-mass at most $u$ has mass at most $u/(tV(t))$ under the uniform law on $K_t$. Exactly as in \cref{prop:wet-density}, the layer-cake formula and \cref{lem:level-int} (in dimension $n$) give
\[
\nu(W^0_\nu(u))\le A^0_nu^\alpha\int_0^{f_{\max}}t^{-\alpha}V(t)^{1-\alpha}\,dt\le A^0_nL_{n,p,\alpha}u^\alpha.
\]
The capped-density approximation of \cref{prop:wet-density} preserves the supporting halfspace and gives the general case.
\end{proof}

\begin{proof}[Proof of \cref{thm:exp-upper}]
Put $\alpha=2/n$. The target hyperplane splits $\R^n$ into closed halfspaces $H_+$ and $H_-$ with masses $\omega_+$ and $\omega_-$ summing to one. Ignore a side of mass zero. The conditional laws $D_\pm=D(\cdot\mid H_\pm)$ still have densities whose $(-1/p)$ powers are convex, since restriction to a convex set and multiplication by a constant preserve this. On $H_\pm$ the target label is constant, so for $h\in B_D(h^\star,r)$ the set $\{h\ne h^\star\}\cap H_\pm$ is the intersection of $H_\pm$ with a homogeneous halfspace of $D_\pm$-mass at most $r/\omega_\pm$. Because $h^\star\in B_D(h^\star,r)$, the disagreement region is the union of these sets over $h\in B_D(h^\star,r)$. \Cref{lem:cone-density} on each side gives
\[
D\bigl(\DIS(B_D(h^\star,r))\bigr)\le\sum_{\pm}\omega_\pm A^0_nL_{n,p,\alpha}\Bigl(\frac r{\omega_\pm}\Bigr)^\alpha\le2^\alpha A^0_nL_{n,p,\alpha}\,r^\alpha,
\]
since $\omega_+^{1-\alpha}+\omega_-^{1-\alpha}\le2^\alpha$ by concavity. Take $K_{n,p}=2^\alpha A^0_nL_{n,p,\alpha}$. Dividing by $r$ and taking the supremum over $r\ge\eps$ gives the bound on $\theta_D$ (values $r>1$ contribute at most one). In the intermediate range $n<p\le n+1$ and $1-n/p=1+ns$, so $L_{n,p,\alpha}\le2^{(n+1)(1-2/n)}/((1-2/n)(1+ns))$, which gives the stated dependence on $s$.
\end{proof}

\fi

\section{Application to annotation and certification of learned verifiers}\label{sec:verifier-main}
A generator produces unlabeled reasoning states, while expert judgments of their validity are expensive labels. Let $\phi(Z)\in\R^n\setminus\{0\}$ be a fixed known representation and let $h^\star(Z)=\sign\langle w^\star,\phi(Z)\rangle$ be realizable. The angular assumptions concern which feature directions the generator produces, and \cref{sec:verifier-model} discusses the scope of this modeling choice.

\Cref{thm:main} applies to offline annotation, in which the learner selects which generated states to label. Coupling the learned and perfect-verifier executions until their first different verdict transfers error under the exploration law to deployment. If the laws of the states visited by the perfect verifier have densities at most $C_t$ relative to the exploration law, then for every $\xi\in(0,1)$, $O(n\log^3n\log(n\sum_tC_t/(\xi\delta)))$ labels give agreement of the whole deployment trajectory with probability $1-\delta-\xi$ (\cref{cor:offline}). \Cref{thm:coverage-profile} extends this to unbounded density ratios through a coverage profile, and \cref{prop:occupancy} replaces coverage of full state laws by angular occupancy.

\Cref{thm:two-laws} separates the assumptions on evaluation and annotation. Evaluation directions must be hyperplane-null, whereas the law of proposed states, which plays the role of the query law of \cref{thm:two-laws} and which we call the \emph{proposal law}, only needs symmetric full support and may be atomic (\cref{sec:two-laws}). A supplied positive margin permits exact certification of a known finite deployment set using $O(n\log^3n\log(M/\delta))$ labels, and directional nets extend certification to infinite sets (\cref{sec:certificate}). Bounds on the number of proposals and a zero-margin obstruction make the access requirements explicit (\cref{sec:generation-cost}).

\Cref{thm:interaction} compares three training interfaces on one two-dimensional family with $M=2^m$ targets. At fixed confidence, successful-trace behavioral cloning needs $\Theta(M)$ demonstrations. Active annotation uses $m$ expert verdicts but needs $\Omega(M)$ pool states. Interaction uses $m$ states and $m$ verdicts. The objective is to learn the verdicts, or to reproduce the perfect-verifier trajectory, rather than to succeed at a task by imitation. Proofs and interface definitions are in \cref{app:verifier}.

\section{Related work}\label{sec:related}
Our contribution is a target-independent, robust simulation of synthesized queries from ordinary sample access, together with quantitative accessibility bounds and sharp separations between the label complexity, which is uniform over marginals, and the number of unlabeled examples, which is not. Separately, we give the sharp $s$-concave phase diagram, including a matching intermediate exponent. We group prior work by the part of this contribution it bears on.

\paragraph{Pool access, membership queries, and point location.}
Point location in hyperplane arrangements \citep{Mei93,ES19,EHKS20} labels a finite set of $N$ points with a number of linear queries polynomial in $n$ and logarithmic in $N$, and \citet{HKLM20} give a near-optimal active-learning form that uses $\tilde O(n\log N)$ synthesized queries. We do not give a new exact-query point-location theorem. We make the algorithm of \citet{HKLM20} tolerate subspace-relative query uncertainty and show that an ordinary pool provides such an oracle with a target-independent precision. \citet{WD26} ask which classes are easy under synthesis, through adaptive identification complexity and the synthesized halving dimension. We ask when a marginal makes the needed queries accessible to a pool, where exact-query guarantees do not survive because a nearby pool point may lie across the boundary. Without synthesis, \citet{BLMSS26} show that on an arbitrary finite point set, restricting the normal to $K$ known directions gives query complexity governed by $K+\log N$, and \cref{prop:need-support,prop:need-null} are distributional versions of the classical obstruction. \citet{DKM24} prove an information-theoretic lower bound for affine halfspaces under $\mathcal N(0,I)$. For a random halfspace of bias $b$, any pool learner with $m\ge\mathrm{poly}(d/b)$ unlabeled examples needs $\tilde\Omega(d/(b\log m))$ labels to reach error $b/2$. So reaching that error with $k$ labels requires $\log m=\tilde\Omega(d/(bk))$, and in particular polylogarithmically many labels require a pool of size $\exp(\tilde\Omega(d/b))$. Membership queries, in contrast, give efficient agnostic learning with $\tilde O(\min\{1/b,1/\eps\}+d\,\mathrm{polylog}(1/\eps))$ queries. They remark that pools approach membership queries as they grow exponentially. \Cref{cor:affine} makes this precise in labels for general halfspaces under every full-support marginal with no mass on affine hyperplanes, including $\mathcal N(0,I)$. With an unlimited pool, $O(d\log^3d\log(d/(\eps\delta)))$ labels suffice. Their lower bound then forces any learner with this label count to inspect $\exp(\tilde\Omega(1/\eps))$ unlabeled examples on their instances. 

Other relaxed query models differ from ours in what an answer means. Comparison queries \citep{HKL20,HKLM20noise,KLMZ17,KLM18} return relative information, not a label at a nearby point. Unreliable boundary queries \citep{BCGS95} allow wrong or ``don't know'' answers near the boundary under a marginal with no mass there, and yield polynomial-time learners for intersections of two halfspaces, extending \citet{Bau90}, and for subclasses of monotone DNF. Our subspace-relative oracle with a margin floor is a relative, data-driven analogue aimed at near-optimal query complexity. \citet{AFK13} restrict synthesized queries to neighborhoods of random examples, whereas our locality is angular and every label is that of an actual sample. \citet{GSS13} analyze a margin-based pool algorithm, but a margin-dependent bound with a possibly tiny floor does not give a uniform label rate.

\paragraph{Disagreement, splitting, and star-based lower bounds.}
CAL \citep{CAL94} queries exactly where the version space disagrees, and the disagreement coefficient \citep{Han07,Han14} bounds how often. It is the route from our $s$-concave disagreement bounds to label bounds. The simulation theorem needs no small disagreement coefficient, and a large disagreement region is not by itself a lower bound for every learner, which is why our lower bounds are proved directly (\cref{lem:star}). Dasgupta's splitting index \citep{Das05} (see also \citealp[Sec.~8.2]{Han14}) separates \emph{informativeness}, how many hypothesis pairs a query splits, from \emph{availability}, the probability $\tau$ (in his notation) that the unlabeled distribution produces such a query. In our setting availability is the cap mass $a_{\bar\mu}$ of the symmetrized angular law, and $Q$ adaptive queries at radius $\gamma$ cost $O(Q\,a_{\bar\mu}(\gamma)^{-1}\log(Q/\delta))$ draws with high probability. The guarantees differ in two ways. Dasgupta's upper bound for a $(\rho,\eps,\tau)$-splittable class uses $\tilde O(n/\rho)$ labels whose hidden logarithmic factors include $\log(1/\tau)$, which depends on the marginal, whereas \cref{thm:main} has no marginal-dependent factor. And his splittability guarantee for halfspaces \citep[Theorem~5]{Das05} is local and for the uniform sphere, whereas we need no splitting guarantee for an arbitrary angular law. \citet{HY15} give distribution-free minimax bounds governed by the star number, which is infinite for halfspaces and gives $\Omega(1/\eps)$ over unrestricted marginals. \citet{WHE15} refine disagreement-based analyses through version-space compression, and \citet{Han24star} relates the star number to the VC dimension of version spaces and disagreement regions. Our lower bounds are star configurations whose cap masses are chosen to match the accuracy and the unlabeled budget.

\paragraph{Distribution-specific halfspaces and the $s$-concave phase diagram.}
Margin localization \citep{BL13,ABL17,BZ17} queries labels in a shrinking band around the current estimate and gives efficient learners under log-concave and $s$-concave laws, also with Massart and Tsybakov noise \citep{ZL21,LZ24}. These results give noise tolerance, polynomial unlabeled cost and polynomial runtime under stronger geometry. Their analysis needs band masses, disagreement outside the band and conditional moments, the inputs \citet{BZ17} develop for $s$-concave laws via Bobkov's heavy-tail concentration \citep{Bob07}. Our margin floor has a different role. It sets the accuracy at which a sampled direction can stand in for a synthesized query. For the phase diagram we use known radial-body and concavity results: Ball's bodies \citep{Bal88} and Bobkov's extension \citep{Bob10}, moment and centroid estimates \citep{AGLLOPT12,FLM20,MP21}, the Fradelizi--Gu\'edon--Pajor comparisons \citep{FGP14}, the entropy and convexity facts surveyed by \citet{MMX17}, and the B\'ar\'any--Larman bound on the small-mass halfspace region of a convex body \citep{BarL88}. Our new ingredients are the projective and level-set argument of \cref{sec:exponent-proof}, which gives the sharp intermediate exponent $(n-2)/n$, and the positive-depth refinement (\cref{prop:depth}), which shows that logarithmic disagreement holds for all $s>-1/n$ at reference points of uniformly positive depth. The results of \citet{BFGHR20} on these regions for general measures do not give the concavity-dependent bound. Symmetry gives depth $1/2$, which is why the symmetric and general thresholds differ.\looseness-1

\paragraph{Quantifier order.}
\citet{BHW10} separate achieving small error from certifying it, and activization \citep{Han12} improves asymptotic label complexity on fixed problems. Our bound is instead uniform over all targets and all admissible marginals, and only the number of unlabeled examples depends on the marginal. The two recent frameworks closest in spirit use different quantifiers (\cref{tab:hanneke}). \citet{HKMV24} fix the unlabeled budget before execution and let rate constants depend on the distribution, so our lower bounds, whose hard distribution varies with the budget and accuracy, do not bound a universal rate for one fixed distribution. \citet{HMT25} require uniformity over targets for a fixed marginal but concern passive sample size. Two reductions make the comparison concrete. If our algorithm uses $N_U$ unlabeled examples and otherwise fails with probability at most $\delta$, stopping it after $u$ examples gives
\[
\E[\err]\le\eps+\delta+\Pr(N_U>u),
\]
so an algorithm in the framework of \citet{HKMV24} needs a choice of $u$ that does not depend on the distribution, and almost-sure finiteness of $N_U$ does not provide one. Conversely, simulating a pool algorithm on $u$ fully labeled examples costs $u$ passive labels, so active label savings do not become passive sample savings. Partial concept classes \citep{AHHM21} are the natural formalism for the admissible queries of \cref{sec:verifier-model}, and \citet{HKMeV26} study language generation in the limit under query and mistake feedback.

\begin{table}[t]
\centering\small
\renewcommand{\arraystretch}{1.15}
\begin{tabular}{>{\raggedright\arraybackslash}p{3.6cm}>{\raggedright\arraybackslash}p{\dimexpr(\linewidth-3.6cm-6\tabcolsep)/2\relax}>{\raggedright\arraybackslash}p{\dimexpr(\linewidth-3.6cm-6\tabcolsep)/2\relax}}
\toprule
Work & Main resource and quantifiers & Relation to this paper\\
\midrule
\citet{HKMV24} & Label budget $k$. Unlabeled budget $u(k)$ fixed before execution. Rate constants may depend on the joint distribution $P$ & Unlabeled budget fixed in advance, whereas ours is adaptive and marginal-dependent\\
\citet{HMT25} & Passive labeled sample size. Guarantees may depend on the marginal but are uniform over targets & Uniform over targets, as our precision is, but about passive sample size rather than active labels\\
\bottomrule
\end{tabular}
\caption{Quantifier comparison with recent distribution-sensitive frameworks.}
\label{tab:hanneke}
\end{table}

\paragraph{Noise and verifier supervision.}
General agnostic active learners \citep{BBL09,DHM07}, importance weighting \citep{BDL09}, analyses beyond disagreement \citep{ZC14}, and experimental design over a pool \citep{KZJJ21} address noise rather than access. \citet{Han25} removes disagreement penalties from the leading agnostic term, with bound $O\bigl(\tfrac{\mathrm{OPT}^2}{\eps^2}(d+\log\tfrac1\delta)\bigr)+\tilde O(d/\eps)$, where $\mathrm{OPT}$ is the error of the best classifier in the class. The residual $d/\eps$ term means it does not give our realizable polylogarithmic rate. Related models for verifiers include reliable apprenticeship \citep{HYWS25} and online selective sampling \citep{HY21}. Robustly-reliable learners output, with each prediction, a level of training-data poisoning up to which the prediction is guaranteed to be correct, also in active learning \citep{BBHS22}, and \citet{BHPS23} give such guarantees under test-time attacks and distribution shift. \citet{BHPS25} study short certificates for reliable predictions. \citet{BBLS25} develop PAC objectives for learned chain-of-thought verifiers, and \citet{BBFLS26} analyze online verifier learning under generator--verifier feedback. We impose the coverage assumption in advance and do not show that feedback-induced shift satisfies it.

\section{Discussion and open problems}\label{sec:discussion}
Under two qualitative angular assumptions, each necessary, a pool learner matches the membership-query label rate uniformly over marginals and targets. The number of unlabeled examples depends on the marginal, and the lower bounds show that this dependence cannot be removed. The results therefore describe a tradeoff between labels and unlabeled examples, in the spirit of the splitting-index distinction between informativeness and availability, rather than an equivalence between the two query models. For $s$-concave marginals the thresholds are sharp, and so are the polynomial rates below them.

\paragraph{The label--unlabeled tradeoff for a fixed angular law.} Characterize the achievable pairs (unlabeled samples, labels) for a fixed angular law $\mu$, interpolating between unlimited unlabeled access (\cref{thm:main}) and a distribution-uniform budget (\cref{prop:finite-budget,thm:affine-core}). Our upper bound depends on $\mu$ through the cap masses $a_{\bar\mu}(\gamma_j)$ and on the conditioning $\kappa(T_j)$ of the data-dependent isotropic transforms (\cref{thm:unlabeled}). How do cap accessibility and transform conditioning enter the optimal tradeoff? A concrete first step is to bound $\kappa(T_j)$, with high probability over the passive sample, by a function of $n$, $N$, and a general-position parameter of the sample such as $\sigma_{\mathrm{gp}}=\min_{|S|=n}\sigma_{\min}(Y_S)$ from \cref{lem:explicit-floor}. Such a bound would turn \cref{thm:unlabeled} into a high-probability unlabeled bound in terms of $a_{\bar\mu}$ and the small-ball probabilities of $\mu$. Bounding the expected cost would further require integrability, for example control of $\E[1/a_{\bar\mu}(\gamma_j)]$. An almost surely finite cost can have infinite expectation.

\paragraph{Fixed-distribution rates.} Let $k$ be the label budget and $u(k)$ an unlabeled schedule that the learner fixes before seeing data. Our lower bounds let the marginal depend on the accuracy and on the schedule. Two fixed-distribution questions remain. First, for one fixed marginal $D$ satisfying our assumptions and worst-case targets, is there a polynomial schedule $u(k)$ under which the optimal error decays only polynomially in $k$, although it decays as $\exp(-\Omega(k/(n\log^3n)))$ with unlimited unlabeled data? Second, the same question for the universal rates of \citet{HKMV24}, in which $D$ and the target are both fixed and the rate constants may depend on them.

\paragraph{An efficient margin floor with a quality guarantee.} The explicit floor costs $M_0^{O(n)}$ arithmetic operations (\cref{sec:explicit-floor}). A positive floor alone is not enough, since an overly conservative floor makes the cap-waiting cost prohibitive. In the arithmetic model on a sample of size $M_0$, is there an algorithm running in time polynomial in $n$ and $M_0$ that outputs an admissible floor within a factor $\mathrm{poly}(n)$ of the largest width whose worst empirical band mass is at most $\delta/(32N)$?

\paragraph{Joint geometry of disagreement.} \Cref{thm:exp-upper} gives the uniform bound $\theta_D(\eps)\le C_n(1+ns)^{-1}\eps^{-(n-2)/n}$ for $-1/n<s\le-1/(n+1)$, and \cref{prop:depth} gives $O(\log(1/\eps))$ at reference points of positive depth. What is the optimal joint dependence of the disagreement coefficient on the accuracy, the depth of the reference point, the dimension, and $1+ns$? Matching logarithmic lower bounds in the regimes where we prove $O(\log(1/\eps))$ are also open, and more broadly one may ask which angular geometries between elliptical and arbitrary centrally symmetric laws have logarithmic disagreement.

\paragraph{Adversarial noise and agents.} Under Massart noise the pool simulation extends by majority votes (\cref{cor:massart-pool}). Adversarial noise can concentrate on the rare caps that the simulation uses, so an analogue of \cref{thm:main} needs point location that tolerates both relative query uncertainty and adversarially placed label errors. A related direction is to study active learning when the examples are agents who can improve their features in response to the deployed classifier~\citep{ABNSSW25,SS25}.

\paragraph{Data-driven active learning.} Our learners set their parameters, such as the margin floor, the tolerance $\eta$, the cap precision $\gamma$ and the unlabeled schedule, by worst-case analysis. Data-driven algorithm design \citep{Bal20} instead tunes such parameters on a sample of related tasks, with generalization guarantees for the tuned choice. For learning with few labels, this approach has mainly been developed for semi-supervised learning (also pool-based, but the learner cannot decide which points to label) where the similarity graph and the hyperparameters are learned across tasks \citep{BS21,SJ23,DHS25}. Our work motivates data-driven configuration of active learning algorithms. \Cref{prop:finite-budget,thm:affine-core} rule out an unlabeled budget that works uniformly over marginals, but they leave open whether a budget, margin floor or query rule tuned on tasks from an unknown task distribution can achieve near-optimal label and unlabeled costs on average over that distribution.

\paragraph{Secondary directions.} Robust point location costs one extra $\log n$ factor over the exact-query learner, and removing it would require a selection rule with the comparison cost of exact median selection under an imprecise comparator (\cref{app:robust}). For verifiers, the central question is whether interactive pools can match the label rate of \cref{thm:main} under feedback-induced state shift, without the coverage assumption (see also \cref{sec:three-way}).

\section*{AI Disclosure and Acknowledgment}
AI assistants (primarily ChatGPT) were used for literature search and positioning, for drafting and revising text, and for reviewing proofs. The open question for $s$-concave distributions was suggested by Nina Balcan. This work was supported in part by the National Science Foundation under grants ECCS-2216899,
ECCS-2216970.

\appendix
\crefalias{section}{appendix}
\section{Deferred proofs from Sections~\ref{sec:simulation} and~\ref{sec:lower}}\label{app:deferred}
\subsection{Margin floor and unlabeled cost}
\begin{proof}[Proof of \cref{lem:anticonc}]
For the first implication, suppose bands of widths $\theta_j\downarrow0$ and normals $w_j$ have mass at least $c>0$. Pass to a subsequence with $w_j\to w$. For each $a>0$, eventually those bands lie in $\{u:|\langle w,u\rangle|<a\}$. Continuity from above then gives $\mu(w^\perp\cap\Sph)\ge c$, a contradiction. The converse follows because every band contains its central great subsphere. Under hyperplane-nullness, condition on the first $j<n$ samples. Their span is contained in a hyperplane of zero mass, so the next sample avoids it almost surely. Conversely, positive mass on a hyperplane gives positive probability that all $n$ samples lie in it.
\end{proof}

\begin{proof}[Proof of \cref{lem:floor}]
Origin-centered bands $\{u:|\langle u,w\rangle|<\theta\}$ are intersections of two parallel affine halfspaces, so they form a class of VC dimension $O(n)$. A relative-deviation bound gives, simultaneously for all bands $\mathcal B$, $\mu(\mathcal B)\le 2\hat\mu_{M_0}(\mathcal B)+C\frac{n\log M_0+\log(8/\delta)}{M_0}$. Let $\hat\rho(\theta)$ be the largest empirical band mass and choose the largest dyadic $\theta$ with $\hat\rho(\theta)\le\delta/(32N)$. Such a width exists almost surely. Any $n$ sampled directions are linearly independent, so for sufficiently small width no band contains $n$ samples, uniformly over orientations by compactness. The displayed sample-size conditions then give the claimed population bound.
\end{proof}

\begin{proof}[Proof of \cref{thm:unlabeled}]
Realize the waiting from a dedicated i.i.d.\ stream $Y_1,Y_2,\dots\sim\mu$, independent of the margin-floor batch, the passive sample, and the learner's coins. For each query $j$ let $C_j$ be the requested $\gamma_j$-cap and $E_j=C_j\cup-C_j$ the corresponding double cap. Define the \emph{transcript} $\mathcal T$ to consist of these three objects together with, for each query $j$, the double cap $E_j$, the accepted pool point $P_j$ and its label, but not the rejected draws or the waiting lengths $W_j$. Success of point location is an event determined by $\mathcal T$, since the learner sees only accepted points, labels and its coins.

We use a standard fact about i.i.d.\ sequences. For a fixed set $E$ with $\mu(E)>0$, the first hitting time $W$ of $E$, the hit point $Y_W$, and the post-hit sequence $(Y_{W+i})_{i\ge1}$ are mutually independent, with $W\sim\mathrm{Geom}(\mu(E))$, $Y_W\sim\mu(\cdot\mid E)$, and the post-hit sequence again i.i.d.\ $\mu$. The double cap $E_j$ is a function of the transcript before query $j$. Applying the fact inductively, conditional on $(W_1,\dots,W_{j-1})$ and the transcript before query $j$, the triple $(W_j,P_j,\text{remaining stream})$ has this product form. Hence the joint law factorizes as the law of $\mathcal T$ times $\prod_j\mathrm{Geom}(\mu(E_j))(W_j)$, so conditional on $\mathcal T$ the $W_j$ are independent with $W_j\sim\mathrm{Geom}(\mu(E_j))$ and $\mu(E_j)\ge\bar\mu(C_j)\ge a_{\bar\mu}(\gamma_j)$. The same conditional law holds given $\mathcal T$ on the success event, since success is determined by $\mathcal T$. Conditioning only on success, without fixing $\mathcal T$, need not preserve independence.

Therefore $\Pr[W_j>\lceil\log(Q/\zeta)/a_{\bar\mu}(\gamma_j)\rceil\mid\mathcal T]\le(1-a_{\bar\mu}(\gamma_j))^{\log(Q/\zeta)/a_{\bar\mu}(\gamma_j)}\le\zeta/Q$, where $Q$ is $\mathcal T$-measurable. A union bound over $j\le Q$ proves the first claim. Since $a_{\bar\mu}$ is nondecreasing, the common-$\Gamma$ bound follows.
\end{proof}

\subsection{Proofs of the additional results of \texorpdfstring{\cref{sec:simulation}}{Section 3}}\label{app:baseline-proofs}
\begin{proof}[Proof of \cref{thm:two-laws}]
Compute the margin floor $\theta_0$ from an unlabeled $D$-batch (\cref{lem:floor}) and draw the passive sample $\mathcal X$ from $D$. Both steps use only hyperplane-nullness of the angular law of $D$, and \cref{cor:passive-floor} gives every point of $\mathcal X$ normalized target margin at least $\theta_0$ with probability at least $1-\delta/4$. Run \cref{prop:robust-pl} on $\mathcal X$ and answer each query $z'$ in an instance $(V_0,T)$ from fresh $\mathcal Q$-draws as in the proof of \cref{thm:main}. Wait for a point $p$ with $\angle(p,T^{-1}z')\le\gamma$ or $\angle(-p,T^{-1}z')\le\gamma$, where $\gamma=\eta^*\theta_0/\kappa(T)$, and return its label, negated in the second case. By \cref{lem:pool-oracle} and the remark after it, which use no property of the law of $p$, the answers form an $\eta^*$-subspace-relative oracle even if $\mathcal Q$ is atomic or charges the target hyperplane. Each draw lands in the required double cap with probability at least $a_{\bar\mu_{\mathcal Q}}(\gamma)>0$, where $\mu_{\mathcal Q}$ is the angular law of $\mathcal Q$, so every wait is finite almost surely. Point location labels $\mathcal X$ correctly with probability at least $1-\delta/4$, and only $\mathcal Q$-points are ever labeled. The realizable VC bound under $D$ finishes the proof as in \cref{thm:main}.
\end{proof}

\begin{proof}[Proof of \cref{cor:affine}]
Apply \cref{thm:main} in $\R^{d+1}$ to $\tilde X=(X,1)$, whose angular law we call $\mu$. The target $\sign(\langle a,x\rangle+b)$ is the homogeneous halfspace $\sign\langle(a,b),\tilde x\rangle$, and errors under $D'$ equal errors under the law of $\tilde X$. A great subsphere $\{u:\langle(a,b),u\rangle=0\}$ pulls back to the affine hyperplane $\{\langle a,x\rangle+b=0\}$, or to the empty set if $a=0$, so $\mu$ is hyperplane-null. For symmetric full support, take $(v,c)\in S^d$ and $\gamma>0$. If $c\ne0$, the $\gamma$-cap around $\sign(c)(v,c)$ meets the open upper hemisphere in a nonempty open set, which is the image of a nonempty open subset of $\R^d$ under the homeomorphism $x\mapsto(x,1)/\|(x,1)\|$. If $c=0$, the $\gamma$-cap around $(v,0)$ contains the directions of the lifts of $\{x:\angle(x,v)<\gamma/2,\ \|x\|>2/\gamma\}$, because $\angle((x,1),(v,0))\le\angle(x,v)+\arctan(1/\|x\|)$. Both sets are open and nonempty, so $D'$ charges them, and $\mu(C\cup-C)>0$ for every cap $C$.
\end{proof}

\begin{proof}[Proof of \cref{prop:baseline}]
\emph{Precision.} Origin-centered bands form a class of VC dimension $O(n)$ (proof of \cref{lem:floor}). By the two-sided relative-deviation bound, for $M_1$ as stated, with probability at least $1-\delta/2$ every band $\mathcal B$ satisfies $\mu(\mathcal B)\le2\hat\mu_{M_1}(\mathcal B)+\eps/8$ and $\hat\mu_{M_1}(\mathcal B)\le2\mu(\mathcal B)+\eps/8$. Let $\hat\rho(\theta)$ be the largest empirical band mass at width $\theta$, and let $\hat\theta$ be the largest dyadic $\theta\le1$ with $\hat\rho(\theta)\le\eps/4$, which exists almost surely as in \cref{lem:explicit-floor} once $M_1\ge4n/\eps$. Then $\rho_\mu(\hat\theta)\le5\eps/8<\eps$, so $\hat\theta\le\rho_\mu^{-1}(\eps)$. Since $\rho_\mu$ is nondecreasing, every $\theta<\rho_\mu^{-1}(\eps/16)$ has $\rho_\mu(\theta)\le\eps/16$ and hence $\hat\rho(\theta)\le\eps/4$, so $\hat\theta\ge\rho_\mu^{-1}(\eps/16)/2$.

\emph{Search.} Write $w=w^\star$ for the unit target normal. Run the directional search in the proof of \cref{thm:dir-search} in the given coordinates ($A=I$), with cap radius $\gamma=c\hat\theta/n^{3/2}$ in place of $\eps/(Cn^2L)$ and final bisection width $b=\hat\theta/(C'\sqrt n)$ in place of $\eps/(C_2nL)$, for suitable absolute constants $c\le1/8$ and $C'$. Answer each unit query $q$ with the label of the first fresh pool point $p$ with $\angle(p,q)<\gamma$, or with the negated label of the first with $\angle(-p,q)<\gamma$. An angle below $\gamma$ implies Euclidean distance below $\gamma$, so each answer is correct whenever $|\langle w,q\rangle|>\gamma$ (in the negated case $-p$ then lies strictly on the side of $q$, so $h_w(-p)=-h_w(p)$). Following that proof, the tournament returns $k$ with $|w_k|\ge\max_i|w_i|-4n\gamma\ge1/(2\sqrt n)$ because $c\le1/8$, and \cref{lem:fbs} estimates every ratio $w_i/w_k$ within $C_1n\gamma+b/2\le(C_1c+1/(2C'))\hat\theta/\sqrt n$. The reconstructed vector differs from $w/|w_k|$, whose norm is at least one, by at most $\sqrt n$ times this, so the normalized estimate $\hat w$ satisfies $\|\hat w-w\|\le\frac\pi2(C_1c+1/(2C'))\hat\theta<\hat\theta$ for $c$ small and $C'$ large. The tournament uses $O(n)$ labels and the bisections $n\lceil\log_2(4C'n/\hat\theta)\rceil$, so the label count is $O(n\log(n/\hat\theta))=O(n\log(n/\rho_\mu^{-1}(\eps/16)))$. If $\sign\langle u,\hat w\rangle\ne\sign\langle u,w\rangle$ for a unit $u$, then $|\langle u,w\rangle|\le|\langle u,w-\hat w\rangle|<\hat\theta$. Hence the error is at most $\rho_\mu(\hat\theta)<\eps$ on the event of the precision step.

\emph{Unlabeled cost.} The number $Q$ of queries is determined by $\hat\theta$, and each query waits for a draw in a double cap of angular radius $\gamma$, whose $\mu$-mass is at least $a_{\bar\mu}(\gamma)$. The argument of \cref{thm:unlabeled}, with every $\gamma_j=\gamma$ and $\zeta=\delta/2$, bounds the number of draws by $Q\lceil\log(2Q/\delta)/a_{\bar\mu}(\gamma)\rceil$ with probability at least $1-\delta/2$.
\end{proof}

\begin{corollary}[Bounded one-dimensional marginals]\label{cor:bounded}
Suppose that $D$ satisfies the assumptions of \cref{prop:baseline}, is isotropic, and that every one-dimensional marginal $\langle X,w\rangle$, $\|w\|=1$, has density at most $c_\infty$. Then $\rho_\mu^{-1}(\eps)\ge\min\{1,\eps^{3/2}/(6c_\infty\sqrt n)\}$, and \cref{prop:baseline} uses $O(n\log(nc_\infty/\eps))$ labels. This covers isotropic log-concave laws, where $c_\infty$ is an absolute constant, and isotropic $s$-concave laws with $s>-1/(n+2)$, where $c_\infty$ depends only on the slack $-1/s-(n+2)$ (\cref{thm:density} with $k=1$).
\end{corollary}
\begin{proof}[Proof of \cref{cor:bounded}]
For a unit $w$ and $R>0$, $|\langle X/\|X\|,w\rangle|<\theta$ forces $|\langle X,w\rangle|<\theta R$ or $\|X\|\ge R$, so isotropy ($\E\|X\|^2=n$) and Markov's inequality give
\[
\mu\{u:|\langle u,w\rangle|<\theta\}\le\Pr[|\langle X,w\rangle|<\theta R]+\Pr[\|X\|\ge R]\le2c_\infty\theta R+n/R^2.
\]
With $R=\sqrt{2n/\eps}$ and $\theta=\eps^{3/2}/(6c_\infty\sqrt n)$ the right side is $(\sqrt2/3+1/2)\eps<\eps$. Isotropic one-dimensional log-concave densities are bounded by an absolute constant, and for $s>-1/(n+2)$ the one-dimensional marginals are isotropic with the same positive slack (\cref{lem:marginals}), so \cref{thm:density}(iv) with $k=1$ bounds $c_\infty$. Such laws have full-dimensional support with the mean in its interior, so their angular laws are hyperplane-null with full support (proof of \cref{cor:all-s}).
\end{proof}

\begin{proof}[Proof of \cref{ex:mixture}]
Fix an invertible $A$ and let $S_i=A\Sigma_iA^\top$, the covariance of $AX$ under the $i$th flat component. Since $\Sigma_1\succeq e_2e_2^\top$, $\lambda_{\max}(S_1)\ge\|Ae_2\|^2$. Testing $S_1$ on $A^{-\top}e_1$ gives $\lambda_{\min}(S_1)\le b^2/\|A^{-\top}e_1\|^2$. If both $S_i$ had condition number at most $K$, then $\|Ae_2\|^2\|A^{-\top}e_1\|^2\le Kb^2$, and symmetrically $\|Ae_1\|^2\|A^{-\top}e_2\|^2\le Kb^2$. Multiplying the two bounds and using $\|Ae\|\,\|A^{-\top}e\|\ge\langle Ae,A^{-\top}e\rangle=1$ for $e=e_1,e_2$ gives $K\ge b^{-2}$. So some $S_i$ has eigenvalues $\lambda_1\le\dots\le\lambda_n$, with unit eigenvectors $v_1,\dots,v_n$, such that $\lambda_1/\lambda_n\le b^2$. Under that component write $AX=\sum_k\sqrt{\lambda_k}g_kv_k$ with $g_k$ i.i.d.\ $\mathcal N(0,1)$. Then
\[
\bigl|\langle AX/\|AX\|,v_1\rangle\bigr|\le\frac{\sqrt{\lambda_1}|g_1|}{\sqrt{\lambda_n}|g_n|}\le\frac{b|g_1|}{|g_n|}\le4b
\]
whenever $|g_1|\le2$ and $|g_n|\ge1/2$, which happens with probability greater than $1/2$. The component has weight $3\eps$, so the band of width $5b$ around $v_1^\perp$ has $\mu_A$-mass greater than $3\eps/2>\eps$, and $\rho_{\mu_A}^{-1}(\eps)\le5b$. The directional search of \cref{prop:baseline} in coordinates $A$ bisects to width $\hat\theta/(C'\sqrt n)\le\rho_{\mu_A}^{-1}(\eps)/(C'\sqrt n)$ in each of $n-1$ coordinates, so it uses at least of order $n\log(1/b)$ labels.
\end{proof}

\begin{corollary}[Best of both]\label{cor:combined}
Run the precision step of \cref{prop:baseline}. If $\log(n/\hat\theta)\le\log^3n\,\log(n/(\eps\delta))$, run its directional search, and otherwise run the simulation of \cref{thm:main}. With probability at least $1-2\delta$ the output has error at most $\eps$, using
\[
O\Bigl(\min\Bigl\{n\log\frac{n}{\rho_\mu^{-1}(\eps/16)},\;n\log^3n\,\log\frac{n}{\eps\delta}\Bigr\}\Bigr)
\]
labels, uniformly over the class of \cref{thm:main}. When $\log(2n/\rho_\mu^{-1}(\eps/16))\le\log^3n\,\log(n/(\eps\delta))$, the first branch is taken and the unlabeled cost obeys the explicit bound of \cref{prop:baseline}. This holds, for example, for the marginals of \cref{cor:bounded} once $n$ exceeds a constant depending on $c_\infty$.
\end{corollary}
\begin{proof}[Proof of \cref{cor:combined}]
On the event of the precision step, $\rho_\mu^{-1}(\eps/16)/2\le\hat\theta$. In the first branch the label count is $O(n\log(n/\hat\theta))$, which is $O(n\log(2n/\rho_\mu^{-1}(\eps/16)))$ and, by the branch condition, $O(n\log^3n\log(n/(\eps\delta)))$. In the second branch it is $O(n\log^3n\log(n/(\eps\delta)))$, and the branch condition gives $\log^3n\log(n/(\eps\delta))<\log(n/\hat\theta)\le\log(2n/\rho_\mu^{-1}(\eps/16))$. The failure probabilities add. If $\log(2n/\rho_\mu^{-1}(\eps/16))\le\log^3n\log(n/(\eps\delta))$, then $\log(n/\hat\theta)$ satisfies the branch condition. Under \cref{cor:bounded}, $\log(2n/\rho_\mu^{-1}(\eps/16))\le\log(768c_\infty n^{3/2})+\frac32\log(1/\eps)$, which is at most $\log^4n+\log^3n\log(1/\eps)\le\log^3n\log(n/(\eps\delta))$ once $\log^3n\ge3/2$ and $\log^4n\ge\log(768c_\infty n^{3/2})$.
\end{proof}

\subsection{Lower-bound tools and proofs}
\begin{lemma}[Star packing]\label{lem:star}
Let $D$ be a distribution on $\R^n$. Let $h_1,\dots,h_M$ be homogeneous halfspaces, let $h_0:\R^n\to\{\pm1\}$ be measurable, and let $C_1,\dots,C_M$ be pairwise disjoint measurable sets and $R$ a measurable set with $D(R)\le u$, such that
\[
\{h_j\ne h_0\}\subseteq C_j\cup R\qquad(1\le j\le M).
\]
\begin{enumerate}[label=(\alph*)]
\item \emph{Disagreement.} If $h_0$ is a homogeneous halfspace and $D(C_j\cap\{h_j\ne h_0\})\ge a$ for every $j$, then with $r=\max_jD(h_j\ne h_0)\le\max_jD(C_j)+u$,
\[
D\bigl(\DIS(B_D(h_0,r))\bigr)\ge Ma\qquad\text{and}\qquad\theta_D(r;h_0)\ge\frac{Ma}r.
\]
\item \emph{Labels.} Suppose $D(h_i\ne h_j)>2\eps$ for $i\ne j$. Every possibly randomized, computationally unbounded pool learner that knows $D$, inspects at most $U$ stream elements, requests at most $q$ labels among them, and outputs a hypothesis (possibly improper) of error at most $\eps$ with probability at least $1-\delta$ for every target $h_1,\dots,h_M$ satisfies
\[
q\ge M(1-\delta-Uu)-1.
\]
\end{enumerate}
\end{lemma}
\begin{proof}
(a) Each $h_j$ lies in $B_D(h_0,r)$, so $\DIS(B_D(h_0,r))\supseteq\bigcup_j\{h_j\ne h_0\}\supseteq\bigcup_j(C_j\cap\{h_j\ne h_0\})$, a disjoint union of mass at least $Ma$. The bound on $r$ holds because $\{h_j\ne h_0\}\subseteq C_j\cup R$.

(b) Let $E$ be the event that \emph{all} of the first $U$ stream elements avoid $R$, whether or not the learner inspects them, so $\Pr(E^c)\le Uu$. Choose the target index $J$ uniformly from $[M]$. The event $E$ and the first $U$ stream elements depend only on the unlabeled stream, so conditioning on $E$, on these points, and on the learner's coins leaves $J$ uniform. (Defining $E$ through the inspected prefix instead would not do, since its length may depend on labels and hence on $J$.) Given this conditioning, consider the \emph{baseline run}, in which every requested label is answered by $h_0$. It queries a set $S$ of at most $q$ points, all outside $R$, and each lies in at most one $C_j$, so at most $q$ indices $j$ have $S\cap C_j\ne\emptyset$. For every other $j$, $h_j=h_0$ on $S$, so by induction along the transcript the run under target $h_j$ coincides with the baseline run and returns the same output. By the separation, that output is $\eps$-accurate for at most one $j$. Hence $\Pr[\text{success}\mid E,\cdot]\le(q+1)/M$, and $1-\delta\le Uu+(q+1)/M$ gives the claim.
\end{proof}
The baseline labeling $h_0$ need not be one of the targets, and nothing is assumed about how the targets relate to one another beyond their separation, so the same coupling covers the symmetric construction, whose targets disagree with $h_0$ on antipodal pairs of caps. If, off $R$, $h_j\ne h_0$ exactly on $C_j$, then $h_i$ and $h_j$ disagree on $(C_i\cup C_j)\setminus R$, so the separation in (b) holds as soon as every $D(C_j\setminus R)>\eps$. For bounds at a common scale $\eps$, take every $D(C_j)\asymp\eps$ with $D(C_j\setminus R)>\eps$, and $u\le\min\{c\eps,\,1/(100U)\}$. Then the same instance gives $\theta_D(C\eps;h_0)=\Omega(M)$ by (a) and $q=\Omega(M)$ at confidence $2/3$ by (b). The requirement $u=O(\eps)$ is needed for the disagreement scale. The budget condition $Uu\le1/100$ alone is not enough.

Both lower-bound mechanisms of the paper are instances of this lemma. The two parts place different requirements on the baseline. Part~(b) allows any measurable $h_0$, and the label bound of \cref{thm:affine-core} uses $h_0\equiv+1$ with $R$ the complement of the core. Part~(a) needs $h_0$ to be a homogeneous halfspace, so the disagreement bound in an affine core instead uses the homogeneous realization of an affine classifier that is positive on the whole core (in \cref{prop:dis-blowup}, the halfspace $H_0$). It agrees with $+1$ on the core, so the inclusion $\{h_j\ne h_0\}\subseteq C_j\cup R$ still holds. In the symmetric endpoint construction (\cref{thm:sym-dis,thm:sym-pool}), $h_0$ is the base target and $R$ is the complement of the antipodal caps. Part~(a) gives the disagreement lower bounds and part~(b) the label lower bounds under a fixed unlabeled budget, which is why the general and symmetric rows of \cref{fig:phase} each describe both quantities. On the upper side, every disagreement bound gives a label bound for CAL with an unlabeled budget fixed in advance. Disagreement bounds do not, however, determine the label complexity of arbitrary active learners.

\begin{lemma}[Many disjoint affine caps]\label{lem:caps}
For every $d\ge2$ there exist constants $c_d,v_d>0$ such that for every $0<v\le v_d$ there are $M\ge c_dv^{-(d-1)/(d+1)}$ affine halfspaces $h_1,\dots,h_M$ whose negative regions inside $B_d$ are, up to null boundaries, pairwise disjoint caps $C_1,\dots,C_M$ of the ball, each satisfying $\nu_d(C_j)=v$.
\end{lemma}
\begin{proof}[Proof of \cref{lem:caps}]
For $u\in S^{d-1}$ and $3/4<b<1$, let $C(u,b)=\{x\in B_d:\langle u,x\rangle\ge b\}$. Its normalized volume is $\frac{\omega_{d-1}}{\omega_d}\int_b^1(1-t^2)^{(d-1)/2}dt$. For $b\le t\le1$ we have $1-t\le1-t^2\le2(1-t)$. Hence $c'_d(1-b)^{(d+1)/2}\le\nu_d(C(u,b))\le C'_d(1-b)^{(d+1)/2}$. By continuity choose $b$ so that the cap mass equals $v$. Then $1-b\asymp_dv^{2/(d+1)}$. Take a maximal Euclidean $4\sqrt{2(1-b)}$-separated subset $\{u_1,\dots,u_M\}$ of the sphere. Standard packing estimates give $M\ge c''_d(1-b)^{-(d-1)/2}\ge c_dv^{-(d-1)/(d+1)}$. The caps are disjoint. If $x\in C(u_i,b)\cap C(u_j,b)$, then $\|x-u_i\|^2=\|x\|^2+1-2\langle u_i,x\rangle\le2(1-b)$, and similarly for $u_j$, whence $\|u_i-u_j\|\le2\sqrt{2(1-b)}$, contradicting the separation. Finally define $h_j(x)=\sign(b-\langle u_j,x\rangle)$, which is negative exactly on $C_j=C(u_j,b)$ up to its null boundary.
\end{proof}

\begin{proof}[Proof of \cref{thm:affine-core}]
Use \cref{def:affine-core} with $u=1/(100\max\{U(\eps),1\})$. Let $D,G,\Pi$ be the resulting objects and set $d=n-1$. Take $v=4\eps$ and, for $\eps_n$ sufficiently small, apply \cref{lem:caps} to obtain pairwise disjoint caps $C_1,\dots,C_M\subset B_d$ of $\nu_d$-mass $v$, with $M\ge c'_n\eps^{-(n-2)/n}$. Let $h_j$ be the affine halfspace whose negative region is $C_j$, and $H_j$ its homogeneous realization on $G$. For $i\ne j$, $d_D(H_i,H_j)\ge D(G)\nu_d(C_i\triangle C_j)=2vD(G)>7\eps$, so no output hypothesis is $\eps$-accurate for two different targets.

Apply \cref{lem:star}(b) with targets $H_1,\dots,H_M$, the labeling $h_0\equiv+1$, the sets $C_j'=G\cap\Pi^{-1}(C_j)$, and $R=G^c$, so $D(R)\le u$. On $G$, $H_j(x)=h_j(\Pi(x))$ is negative exactly on $C_j'$, so $\{H_j\ne h_0\}\subseteq C_j'\cup R$, and the $C_j'$ are disjoint. The pairwise separation $d_D(H_i,H_j)>7\eps$ was shown above. With $\delta=1/3$, the lemma gives $q\ge(2/3-1/100)M-1\ge cM$.
\end{proof}

\begin{proof}[Proof of \cref{prop:finite-budget}]
Let $m=\lfloor1/(4\eps)\rfloor\ge8$. In the first three coordinates, take the points $p_i$, the halfspaces $h_0,h_1,\dots,h_m$, and the smoothed, centrally symmetric law $D_m$ from the proof of \cref{prop:need-support} (\cref{app:necessity}), whose bumps lie in small balls $B_i\ni p_i$ and their reflections $-B_i$. The sets $B_i\cup-B_i$ are pairwise disjoint, each $B_i$ and each $-B_i$ has mass $1/(2m)$, every $h_j$ has constant sign on each $B_i$, $h_i\ne h_0$ on $B_i\cup-B_i$, and $h_i=h_0$ on $B_j\cup-B_j$ for $j\ne i$. Put $u=1/(100\max\{U(\eps),1\})$ and
\[
D=\bigl((1-u)D_m+u\,\mathcal N(0,I_3)\bigr)\otimes \mathcal N(0,I_{n-3}),
\]
which has a smooth, everywhere-positive, centrally symmetric density. Apply \cref{lem:star}(b) with targets $h_1,\dots,h_m$, extended to $\R^n$ by ignoring the last $n-3$ coordinates, baseline $h_0$, sets $C_i=(B_i\cup-B_i)\times\R^{n-3}$, and $R=(\bigcup_iC_i)^c$. Then $D(R)\le u$ and $\{h_i\ne h_0\}\subseteq C_i\cup R$. For $i\ne j$, $h_i\ne h_j$ on $C_i\cup C_j$, so $D(h_i\ne h_j)\ge2(1-u)/m\ge1/m\ge4\eps>2\eps$. With $\delta=1/3$ and $U(\eps)u\le1/100$, the lemma gives $q\ge m(2/3-1/100)-1\ge m/2\ge1/(10\eps)$, using $m\ge8$ and $\eps\le1/32$.
\end{proof}

\section{Robust point location under a subspace-relative oracle}\label{app:robust}
This appendix gives the oracle-dependent steps behind \cref{prop:robust-pl}. We modify the point-location learner of \citet{HKLM20} and re-derive every step whose behavior depends on oracle answers. Numbering refers to arXiv:2004.11380v1. \Cref{tab:hklm} lists every result of \citet{HKLM20} that we use or replace, its hypotheses, whether it involves the oracle, and where those hypotheses are checked under a subspace-relative oracle. Everything not listed is proved below.

\begin{table}[p]
\centering\footnotesize
\renewcommand{\arraystretch}{1.1}
\begin{tabular}{>{\raggedright\arraybackslash}p{2.6cm}>{\raggedright\arraybackslash}p{\dimexpr(\linewidth-2.6cm-6\tabcolsep)/2\relax}>{\raggedright\arraybackslash}p{\dimexpr(\linewidth-2.6cm-6\tabcolsep)/2\relax}}
\toprule
Item of \citet{HKLM20} & Statement and hypotheses & Use here and where re-verified\\
\midrule
Def.~3.3, Lemma~3.4 (margin oracle) & A $\lambda$-representative of a sample $S$ is a point $x_S$ with $\lambda^{-1}A_S\le|m(x_S)|\le\lambda A_S$. The Gaussian combination $x_S=\sum_ig_ix_i$ is one with probability at least $1-5/\lambda$ for $\lambda\ge5$. Oracle-free. & Same construction, restated with a norm bound and proved as \cref{lem:gauss-rep}.\\
Thm.~4.4 (PartialLearn), Lemma~4.6 (IsoLearn), Lemma~4.10 (StructureSearch), Lemma~4.15 (DimReduce) & The exact-query guarantees of the four subroutines. The proof of Thm.~4.4 simulates the margin oracle with $\lambda=\mathrm{poly}(d)/\delta_0$ and a union bound over oracle calls. & \emph{Replaced.} Primed versions are proved under a subspace-relative oracle in \cref{lem:structure,lem:dimreduce,lem:isolearn}, with $\lambda$ polynomial in $d/\delta_0$ (\cref{sec:global}). From Lemma~4.15 we reuse the three-query acceptance test against a scaled reference and the positive-semidefinite domination of spurious accepted subsets.\\
Def.~4.7 (structured pairs) & $(X,\omega)$ is $(k,t,\varsigma,c)$-structured (their $s$ is our $\varsigma$) if $\Pr_\omega[|m(x)|\ge t]\ge k/d$ and $\Pr_\omega[|m(x)|\le t/\varsigma]\ge1-ck/d$. The shorthand ``$(k,t,\varsigma)$-structured'' means $c=5$. & We use $c=50$, produced by \cref{lem:structure}. \Cref{lem:dimreduce,lem:isolearn} are proved for $c=50$, so no statement about $c=5$ is invoked.\\
Claims~4.9, 4.12, 4.13 & For $1/4$-isotropic $(X,\omega)$, the median of the largest normalized margin in a sample of size $\Theta(d)$ ($10d$ there) is $\Omega(1/\sqrt d)$ (4.9). An empirical median over $r$ subsets is a $2/5$-median (4.12, Chernoff). Hence the level-zero representative has margin $\Omega(1/(\lambda\sqrt d))$ with probability $1-O(\delta_0/\log d)$ (4.13). & Not invoked. Level zero of \cref{lem:structure} reuses the isotropy argument from the proof of Claim~4.9 with samples of size $16d$, and the round-robin quantile and the Dvoretzky--Kiefer--Wolfowitz (DKW) inequality replace deterministic selection and Claims~4.12--4.13. This gives $\tilde M_0\ge1/(4\lambda\sqrt d)$.\\
Claim~4.11 & Sample-norm tail transfer for a $u$-median. Oracle-free. & Restated and proved as \cref{lem:norm-tail}.\\
Cor.~4.20 (with Lemmas~4.18, 4.19) & For every $(X,\omega)$ and $\eps>0$ there are $1\le k\le d$, a $k$-dimensional subspace $V$ with $\omega(X\cap V)\ge k/d$, and an invertible $T:V\to V$ putting the normalized restriction in $\eps$-isotropic position. Lemma~4.18 is the theorem of \citet{Bar98} in the form stated by \citet{DSW17}, and Lemma~4.19 characterizes when such a transform exists. Oracle-free. & Used by PartialLearn$'$ (\cref{sec:global}). Since $\omega(X\cap V)>0$, $V$ contains a passive-sample point, as \cref{lem:pool-oracle} requires.\\
Lemmas~5.2, 5.3, Thm.~5.1 (boosting) & WeakLearn is $0.01$-reliable and, with probability $0.99$, has coverage $0.99$ using $O(d\log^2d)$ queries (5.2). If for every distribution $\omega$ over $X$ a learner is $0.01$-reliable, has coverage $0.99$ with probability $0.99$, and uses $Q$ queries, then point location on $X$ is $\delta$-reliably solvable with depth $O(Q\log(n/\delta))$ (5.3). The oracle is used only inside learner calls. & WeakLearn$'$ satisfies the hypotheses of Lemma~5.3 with $Q=O(n\log^3n)$ (\cref{sec:global}).\\
\bottomrule
\end{tabular}
\caption{Results of \citet{HKLM20} that we use or replace.}
\label{tab:hklm}
\end{table}

\citet{HKLM20} work with ternary answers, and their Corollary~5.4 simulates ternary queries by two binary ones. We do not rely on that simulation. Every step below is analyzed directly for binary answers, and a zero-margin query lies inside the tolerance band, where any answer is admissible.

Throughout, $n$ is the ambient dimension of the point-location problem on $X\subset\R^n$ and $d\le n$ is the dimension of the current instance. The local lemmas are stated for a supplied failure parameter $\delta_0\in(0,1/2]$, and the global argument at the end of \cref{sec:global} uses $\delta_0=n^{-C}$ in every instance.

Fix a query instance $(V_0,T)$ chosen by PartialLearn$'$ and work in its coordinates. As in \citet{HKLM20}, each $x\in X\cap V_0$ is replaced by $Tx/\|Tx\|$, which does not change its label. Write $d=\dim V_0$, rescale the transformed target normal $w'$ to unit norm, and put $m(x)=\langle x,w'\rangle$. On this instance an $\eta$-subspace-relative oracle (\cref{def:sr-oracle}) returns $\sign m(q)$ whenever $|m(q)|>\eta\|q\|$ and may answer arbitrarily otherwise. The learner knows every vector it queries, but not the margins $m(\cdot)$, which appear only in the analysis. One-dimensional instances are labeled directly with a single nonzero query, whose relative margin is $1$. Assume $d\ge2$.

\subsection{Three robust primitives}
\begin{lemma}[Imprecise comparator]\label{lem:comparator}
Query $\sigma_a=\sign m(a)$, $\sigma_b=\sign m(b)$ and then $\sign m(\sigma_aa-\sigma_bb)$, declaring $a\ge b$ if the last answer is positive. Under an $\eta$-subspace-relative oracle this comparison is correct whenever $\bigl||m(a)|-|m(b)|\bigr|>\bar\eta:=2\eta(\|a\|+\|b\|)$.
\end{lemma}
\begin{proof}
If $|m(a)|>|m(b)|+\bar\eta$, then $|m(a)|>\eta\|a\|$, so $\sigma_a$ is correct. If $\sigma_b$ is correct, the final query has margin $|m(a)|-|m(b)|>\bar\eta$. If $\sigma_b$ is wrong, its margin is $|m(a)|+|m(b)|>\bar\eta$. In either case this exceeds $\eta\|\sigma_aa-\sigma_bb\|$. If $|m(b)|>|m(a)|+\bar\eta$, $\sigma_b$ is correct and $m(\sigma_aa-\sigma_bb)\le|m(a)|-|m(b)|<-\bar\eta$ regardless of $\sigma_a$. A zero final query receives an arbitrary answer.
\end{proof}

\begin{lemma}[Round-robin quantile]\label{lem:roundrobin}
Let $r\ge10$ candidates with values $v_1,\dots,v_r$ be compared pairwise by a comparator that is correct whenever values differ by more than $\bar\eta$. Let $\hat a$ have the $(\lfloor r/2\rfloor+1)$-st smallest win count. Then at least $r/4$ candidates have value at most $v_{\hat a}+\bar\eta$, and at least $r/4$ have value at least $v_{\hat a}-\bar\eta$.
\end{lemma}
\begin{proof}
Sort the win counts $c_{(1)}\le\dots\le c_{(r)}$ and put $i_0=\lfloor r/2\rfloor+1$. The $i_0$ lowest scorers play $\binom{i_0}2$ games among themselves, so $c_{(i_0)}\ge(i_0-1)/2\ge(r-1)/4$. Let $k=r-i_0+1$ be the size of the top-scoring block. Its average score is at most $(k-1)/2+(r-k)$, and the selected candidate has the smallest score in this block, so it has at least $(k-1)/2$ losses. This and its own value give at least $r/4$ candidates on each side. A win of $\hat a$ over $b$ implies $v_b\le v_{\hat a}+\bar\eta$, while a loss implies $v_b\ge v_{\hat a}-\bar\eta$.
The round-robin rule uses $O(r^2)$ comparisons per level and is the source of the extra $\log n$ factor relative to \citet{HKLM20}. \citet[Cor.~4.7]{AFHN15} sort with $O(r\log^2r)$ comparisons and $O(\log\log r)$ additive error units, and taking the middle element of their output gives two-sided rank control after dividing the tolerance by that factor. This reduces the local comparison cost but does not by itself recover the $\log^2n$ dependence of the exact-query learner.
\end{proof}

\begin{lemma}[Imprecise binary search]\label{lem:fbs}
Let $f(\alpha)=c(\xi-\alpha)$ with $c>0$ and $\xi\in[-R,R]$, and suppose $\sign f(\alpha)$ is reported correctly whenever $|\alpha-\xi|>\Delta$. Bisection of $[-R,R]$ down to width $\alpha_0/2$ uses $\lceil\log_2(4R/\alpha_0)\rceil$ queries and returns $\hat\alpha$ with $|\hat\alpha-\xi|\le\Delta+\alpha_0/4$.
\end{lemma}
\begin{proof}
The invariants $\mathrm{lo}\le\xi+\Delta$ and $\mathrm{hi}\ge\xi-\Delta$ hold initially and are preserved by every update. At termination, midpoint error is at most half the interval width plus $\Delta$.
\end{proof}

\begin{lemma}[Gaussian representatives; cf.\ {\citealp[Def.~3.3, Lemma~3.4]{HKLM20}}]\label{lem:gauss-rep}
For a finite multiset $S$ of unit vectors let $x_S=\sum_{x\in S}g_xx$ with $g_x$ i.i.d.\ $\mathcal N(0,1)$, and $A_S=(\sum_{x\in S}m(x)^2)^{1/2}$. For $\lambda\ge1$, with probability at least $1-3/\lambda$,
\[
A_S/\lambda\le|m(x_S)|\le\lambda A_S\qquad\text{and}\qquad\|x_S\|\le\lambda\sqrt{|S|}.
\]
\end{lemma}
\begin{proof}
$m(x_S)\sim \mathcal N(0,A_S^2)$, so $\Pr[|m(x_S)|<A_S/\lambda]\le\sqrt{2/\pi}/\lambda\le1/\lambda$ and, by Markov, $\Pr[|m(x_S)|>\lambda A_S]\le1/\lambda^2$. Also $\E\|x_S\|^2=|S|$, so $\Pr[\|x_S\|>\lambda\sqrt{|S|}]\le1/\lambda^2$.
\end{proof}

\subsection{Modifications and constants}
Run the learner of \citet{HKLM20} with the following changes.
\begin{enumerate}[label=M\arabic*.]
\item Replace deterministic selection in StructureSearch by the round-robin rule, with\\ $r=\lceil800\ln(8\log_2(2d)/\delta_0)\rceil$ representatives at each level.
\item Start with samples of size $16d$ and halve while the subset size is at least $16$, so levels are $i=0,\dots,L$ with $L=\lfloor\log_2d\rfloor$ and subset sizes $z_i=\lfloor16d/2^i\rfloor$.
\item In DimReduce, accept $S$ iff the imprecise comparator declares $|m(x_S)|\le|m(c\,x_{\mathrm{ref}})|$ with $c=6\lambda^2\sqrt{|S|}/\ell$. In the analysis $|m(c\,x_{\mathrm{ref}})|=m_{\mathrm{acc}}:=2\lambda\sqrt{|S|}(t/\varsigma)$, so the unknown threshold is implemented by scaling the reference vector.
\item Use gap parameter $\ell=C_0d^3\lambda^4$, with $\lambda=(d/\delta_0)^{c_1}$ and $c_1\ge3$ large enough that the failure probability $6/\lambda$ times the deterministic bound $O(d\log(d/\delta_0))$ on the number of representative calls in one IsoLearn$'$ call is at most $\delta_0/16$. (StructureSearch$'$ makes $r(L+1)=O(\log d\log(d/\delta_0))$ calls and DimReduce$'$ makes $R=O((k+1)\log(d/\delta_0))$ with $k\le d$.) Calls with representative norm exceeding $4\lambda\sqrt d$ are declared failed. An observed norm rejection reports failure. Unobserved failures of the representative bounds are charged to the failure probability.
\item Representatives $x_S$ are the Gaussian combinations of \cref{lem:gauss-rep}. The good event of a call is the event of that lemma, which fails with probability at most $3/\lambda\le6/\lambda$.
\item In IsoLearn, the binary search for each coefficient runs over $[-1/\tau,1/\tau]$, since the reference margin is unknown and only known to be at least $\tau$.
\end{enumerate}
All representative vectors have norm at most $N_0=4\lambda\sqrt d$. Put
\[
\tau=(4\lambda\sqrt d)^{-1}(2\ell)^{-(L+1)},\qquad\alpha_0=(96\lambda d^{3/2})^{-1},\qquad\eta^*(d,\delta_0)=\frac{\alpha_0\tau^2}{16\ell N_0}=(d/\delta_0)^{-O(\log d)}.
\]
All of $\lambda,\ell,N_0,L$ are nondecreasing in $d$ and in $1/\delta_0$, so $\eta^*(d,\delta_0)$ is nonincreasing in both. The local lemmas assume an $\eta$-subspace-relative oracle with $\eta\le\eta^*(d,\delta_0)$, and below we write $\eta^*$ for $\eta^*(d,\delta_0)$ when $d,\delta_0$ are clear.

\begin{lemma}[Sample-norm tail transfer; {\citealp[Claim~4.11]{HKLM20}}]\label{lem:norm-tail}
For $z$ independent unit points, set $A_S=(\sum_{x\in S}m(x)^2)^{1/2}$. Then $\Pr[A_S\ge a]\ge u$ implies $\Pr_x[|m(x)|\ge a/\sqrt z]\ge u/z$, and $\Pr[A_S\le b]\ge u$ implies $\Pr_x[|m(x)|\le b]\ge u^{1/z}\ge1-\log(1/u)/z$.
\end{lemma}
\begin{proof}
The first event requires one of the $z$ margins to have magnitude at least $a/\sqrt z$. Apply a union bound. The second requires all $z$ margins to be at most $b$. Independence gives the $z$th power. Finally $e^{-x}\ge1-x$.
\end{proof}

\subsection{StructureSearch\texorpdfstring{$'$}{'}}
Write $\mathrm{Cmp}(a,b)$ for the comparator of \cref{lem:comparator}, which returns ``$a\ge b$'' or ``$a<b$'' for the magnitudes $|m(a)|,|m(b)|$ using three queries.

\begin{algorithm2e}[h]
\caption{StructureSearch$'$}\label{alg:structure}
\KwIn{$(X,\omega)$ in $1/4$-isotropic position in $\R^d$, $\eta$-subspace-relative oracle, parameter $\delta_0$}
$r\gets\lceil800\ln(8\log_2(2d)/\delta_0)\rceil$, $L\gets\lfloor\log_2d\rfloor$\;
\For{$i=0,\dots,L$}{
 Draw $r$ independent samples $S_1,\dots,S_r$ of size $z_i=\lfloor16d/2^i\rfloor$ from $\omega$ and form $x_{S_1},\dots,x_{S_r}$\;
 \lIf{some $\|x_{S_j}\|>N_0$}{\Return failure}
 Run $\mathrm{Cmp}(x_{S_j},x_{S_{j'}})$ for all $j<j'$, and count the wins of each $j$\;
 $\hat x_i\gets$ the representative with the $(\lfloor r/2\rfloor+1)$-st smallest win count\;
 \If{$i\ge1$ and $\mathrm{Cmp}(\hat x_{i-1},\ell\,\hat x_i)$ returns $\hat x_{i-1}\ge\ell\,\hat x_i$}{
  \Return MarginGap with $i^*=i-1$, $x_{\mathrm{ref}}=\hat x_{i-1}$ and $\sigma_{\mathrm{ref}}=\sign m(\hat x_{i-1})$ as queried\;}
}
\Return LargeMargin with $x_{\mathrm{ref}}=\hat x_L$ and $\sigma_{\mathrm{ref}}=\sign m(\hat x_L)$ as queried\;
\end{algorithm2e}

\begin{lemma}\label{lem:structure}
Let $(X,\omega)$ be $1/4$-isotropic in $\R^d$ and suppose the oracle is $\eta$-subspace-relative with $\eta\le\eta^*$. With probability at least $1-\delta_0/2$, StructureSearch$'$ (\cref{alg:structure}) returns $x_{\mathrm{ref}}$ with exact sign $\sigma_{\mathrm{ref}}$ and $M:=|m(x_{\mathrm{ref}})|\ge\tau$, and either
(i) \emph{MarginGap}, meaning $(X,\omega)$ is $(k,t,\varsigma,50)$-structured with $k=2^{i^*}/160$, $t=M/(8\lambda\sqrt d)$ and $\varsigma=\ell/(24\lambda^2\sqrt d)$, or
(ii) \emph{LargeMargin}, meaning the mass of points with $|m(x)|\ge M/(12\lambda)$ is at least $1/320$.
The query cost is $O(r^2\log d)$.
\end{lemma}
\begin{proof}
At each level, DKW places the empirical CDF of the $r$ representative values $Z=|m(x_S)|$ within $1/20$ of the true CDF except with probability $2e^{-r/200}$. Together with the good events of the representative bounds, all levels are good with probability at least $1-\delta_0/2$. Comparisons have precision $\bar\eta=4\eta N_0\le\tau/\ell$. By \cref{lem:roundrobin} and DKW, the selected value $\tilde M_i$ satisfies $\Pr[Z\ge\tilde M_i-\bar\eta]\ge1/5$. Write $A=A_S$ for a sample $S$ at level $i$. Since $Z\le\lambda A$ on the good event, subtracting the representative failure gives
(Q1) $\Pr_S[A\ge(\tilde M_i-\bar\eta)/\lambda]\ge1/10$, and symmetrically (Q2) $\Pr_S[A\le\lambda(\tilde M_i+\bar\eta)]\ge1/10$.
At level zero, $1/4$-isotropy gives $\E_\omega m(x)^2\ge3/(4d)$. Since $m(x)^2\le1$, this forces $\Pr_\omega[|m(x)|\ge\sqrt{3/(8d)}]\ge3/(8d)$, as in the proof of \citet[Claim~4.9]{HKLM20}. A sample of $16d$ points contains such a point except with probability $(1-3/(8d))^{16d}\le e^{-6}$, and then $A\ge1/(2\sqrt d)$. On the good event $Z\ge A/\lambda$, so $\Pr[Z<1/(2\lambda\sqrt d)]\le e^{-6}+3/\lambda<1/5$. On the other side, \cref{lem:roundrobin} puts at least $r/4$ representatives at or below $\tilde M_0+\bar\eta$, and DKW turns this into $\Pr[Z\le\tilde M_0+\bar\eta]\ge1/5$. The selected representative is only an approximate median, but these two bounds suffice. They force $\tilde M_0+\bar\eta\ge1/(2\lambda\sqrt d)$, hence $\tilde M_0\ge1/(4\lambda\sqrt d)$. The gap test $\mathrm{Cmp}(\hat x_i,\ell\hat x_{i+1})$ has error at most $2\eta(1+\ell)N_0\le\tau/2$. If no gap is declared, $\ell\tilde M_{i+1}\ge\tilde M_i-\tau/2$, so inductively $\tilde M_{i+1}\ge\tilde M_i/(2\ell)\ge\tau$, and if a gap is declared, $\ell\tilde M_{i+1}\le\tilde M_i+\tau/2\le2\tilde M_i$. Hence $M\ge\tau>\eta N_0$, and the queried sign of $x_{\mathrm{ref}}$ is exact.

Suppose the first gap occurs at level $i^*$. \Cref{lem:norm-tail} and (Q1) give $\Pr_x[|m(x)|\ge t]\ge k/d$. The declared gap implies $\tilde M_{i^*+1}\le2M/\ell$, hence $\lambda(\tilde M_{i^*+1}+\bar\eta)\le3\lambda M/\ell=t/\varsigma$. Since the gap is declared at level $i^*+1\le L$, $2^{i^*}\le d/2$, so $z_{i^*+1}=\lfloor16d/2^{i^*+1}\rfloor\ge8d/2^{i^*}-1\ge\frac{16\ln10}{5}\cdot\frac d{2^{i^*}}$. Applying (Q2) at level $i^*+1$, \cref{lem:norm-tail} with $u=1/10$ gives $\Pr_x[|m(x)|\le t/\varsigma]\ge1-\ln10/z_{i^*+1}\ge1-\frac5{16}\cdot\frac{2^{i^*}}d=1-50\frac kd$, using $2^{i^*}=160k$. This is MarginGap. If no gap occurs through level $L$, the last level has sample size between $16$ and $31$ and (Q1) yields LargeMargin.
\end{proof}

\subsection{DimReduce\texorpdfstring{$'$}{'}}
\begin{algorithm2e}[h]
\caption{DimReduce$'$}\label{alg:dimreduce}
\KwIn{$(X,\omega)$ in $1/4$-isotropic position in $\R^d$, $\eta$-subspace-relative oracle, $x_{\mathrm{ref}}$ and $k$ from \cref{alg:structure}, parameter $\delta_0$}
$z\gets\lfloor d/(50k)\rfloor$, $R\gets\lceil C(k+1)\log(d/\delta_0)\rceil$, $c\gets6\lambda^2\sqrt z/\ell$, $\hat C\gets0$\;
\For{$j=1,\dots,R$}{
 Draw a sample $S$ of size $z$ from $\omega$ and form $x_S$\;
 \lIf{$\|x_S\|>N_0$}{\Return failure}
 \lIf{$\mathrm{Cmp}(c\,x_{\mathrm{ref}},x_S)$ returns $c\,x_{\mathrm{ref}}\ge x_S$}{$\hat C\gets\hat C+\frac1{Rz}\sum_{x\in S}xx^\top$}
}
\Return the span $V$ of the eigenvectors of $\hat C$ with eigenvalue at least $1/(128d)$\;
\end{algorithm2e}

\begin{lemma}\label{lem:dimreduce}
In the MarginGap case, assume $50k\le d/2$ and take $|S|=\lfloor d/(50k)\rfloor$. With probability at least $1-\delta_0/4$, DimReduce$'$ (\cref{alg:dimreduce}) returns a subspace $V$ with $\dim V\ge d-100k$ and $|m(v)|\le C_2\lambda^3d^{3/2}\frac M\ell\|v\|$ for $v\in V$, using $O((k+1)\log(d/\delta_0))$ queries.
\end{lemma}
\begin{proof}
As in \citet[Lemma~4.15]{HKLM20}, acceptance is tested by comparing $|m(x_S)|$ with $|m(c\,x_{\mathrm{ref}})|$ for $c=6\lambda^2\sqrt{|S|}/\ell$, so that $|m(c\,x_{\mathrm{ref}})|=m_{\mathrm{acc}}$, using the comparator of \cref{lem:comparator}. Let $X_{\mathrm{small}}=\{x:|m(x)|\le t/\varsigma\}$. If a sampled subset lies entirely in $X_{\mathrm{small}}$, then $A_S\le\sqrt{|S|}t/\varsigma$ and $|m(x_S)|\le\lambda\sqrt{|S|}t/\varsigma=m_{\mathrm{acc}}/2$. The comparator error is at most the remaining half of the threshold, so every subset contained in $X_{\mathrm{small}}$ is accepted. Conversely, acceptance gives $|m(x_S)|\le\frac32m_{\mathrm{acc}}$, and on the good event every $x\in S$ satisfies $|m(x)|\le A_S\le\lambda|m(x_S)|\le9\lambda^3\sqrt{|S|}\,M/\ell\le18\lambda^3\sqrt d\,M/\ell=:\bar m$, using $|S|\le d/(50k)\le16d/5$ since $k\ge1/160$. Furthermore $\Pr[S\subseteq X_{\mathrm{small}}]\ge(1-50k/d)^{d/(50k)}\ge1/4$, and conditioned on this event $S$ is i.i.d.\ from $\omega|_{X_{\mathrm{small}}}$.

Write $\Sigma_{\mathrm{far}}=\E[xx^\top\mathbf 1_{(X_{\mathrm{small}})^c}]$. Since points have unit norm, $\operatorname{tr}\Sigma_{\mathrm{far}}\le50k/d$, so fewer than $100k$ eigenvalues of $\Sigma_{\mathrm{far}}$ exceed $1/(2d)$. On their orthogonal complement $W$, $\E[xx^\top\mathbf 1_{X_{\mathrm{small}}}]\succeq(3/(4d))I-\Sigma_{\mathrm{far}}\succeq(1/(4d))I$, and dividing by $\omega(X_{\mathrm{small}})\le1$ preserves this. Use $R=C(k+1)\log(d/\delta_0)$ independent trial subsets of size $z=\lfloor d/(50k)\rfloor$. With high probability at least $R/8$ lie entirely in $X_{\mathrm{small}}$, contributing $\Omega(d\log(d/\delta_0))$ independent draws from $\omega|_{X_{\mathrm{small}}}$. Matrix Chernoff on $W$ makes their empirical covariance at least $I/(8d)$ there, with failure probability at most $\delta_0/16$.

Form the accepted covariance with denominator $Rz$, including zero contributions from rejected subsets. Every subset contained in $X_{\mathrm{small}}$ is accepted, and together these contribute at least $I/(64d)$ on $W$. All other accepted outer products are positive semidefinite, which is the domination argument of \citet[Lemma~4.15]{HKLM20}. Hence at least $d-100k$ eigenvalues are at least $1/(64d)$. Take the eigenvectors with eigenvalues at least $1/(128d)$. If $\hat C$ is this covariance, acceptance bounds every contributing margin by $\bar m$ and the total weight is at most one, so $\|\hat Cw'\|\le\bar m$, and for $v$ in the selected subspace $|m(v)|=|\langle\hat Cw',\hat C^{-1}v\rangle|\le128d\bar m\|v\|$. The query count is $O((k+1)\log(d/\delta_0))$.
\end{proof}

\subsection{IsoLearn\texorpdfstring{$'$}{'}, PartialLearn\texorpdfstring{$'$}{'} and boosting}
\begin{algorithm2e}[h]
\caption{IsoLearn$'$}\label{alg:isolearn}
\KwIn{$(X,\omega)$ in $1/4$-isotropic position in $\R^d$, $\eta$-subspace-relative oracle, parameter $\delta_0$}
Run \cref{alg:structure}\;
\lIf{it reports failure}{\Return the empty partial labeling}
Let $x_{\mathrm{ref}}$ and $\sigma_{\mathrm{ref}}$ be its output\;
\lIf{MarginGap with index $i^*$}{$k\gets2^{i^*}/160$}
\lElse(\tcp*[h]{LargeMargin}){$k\gets d/320$}
\eIf{MarginGap and $50k\le d/2$}{Run \cref{alg:dimreduce} with $x_{\mathrm{ref}}$ and $k$\;
\lIf{it reports failure}{\Return the empty partial labeling}
Let $V$ be its output and $w_1,\dots,w_{d'}$ an orthonormal basis of $V^\perp$\;}{$V\gets\{0\}$, and let $w_1,\dots,w_d$ be the standard basis\;}
\For{each $i$}{Bisect $[-1/\tau,1/\tau]$ to width $\alpha_0/2$, moving the lower end to $\alpha$ when the query $w_i-\alpha\,\sigma_{\mathrm{ref}}x_{\mathrm{ref}}$ is answered positive, and let $\hat\xi_i$ be the final midpoint\;}
\For{each $y\in X$}{Write $y=v+\sum_ib_iw_i$ with $v\in V$, and put $\hat E=\sum_ib_i\hat\xi_i$\;
\lIf{$|\hat E|\ge2\varepsilon_{\mathrm{rec}}$}{label $y$ by $\sign(\hat E)$}}
\end{algorithm2e}

\begin{lemma}\label{lem:isolearn}
Let $d\ge2$ be the instance dimension and $\delta_0\in(0,1/2]$. Under an $\eta$-subspace-relative oracle with $\eta\le\eta^*(d,\delta_0)$, IsoLearn$'$ is $\delta_0$-reliable (all labels it assigns are correct with probability at least $1-\delta_0$) and, with probability at least $1-\delta_0$, labels a set of $\omega$-mass at least $k/d$, using $O((k+1)\log d\log^2(d/\delta_0))$ queries. If \cref{alg:structure} or \cref{alg:dimreduce} reports failure, \cref{alg:isolearn} returns the empty partial labeling.
\end{lemma}
\begin{proof}
Run \cref{alg:isolearn}. In the MarginGap case with $50k\le d/2$, $d'=\dim V^\perp\le100k$. Otherwise the standard basis is used, and either $k>d/100$ or we are in LargeMargin, where we take $k=d/320$. In all cases the number of coordinates is at most $320(k+1)$. Let $\sigma_{\mathrm{ref}}=\sign m(x_{\mathrm{ref}})$, which is exact, and $M=|m(x_{\mathrm{ref}})|\ge\tau$ (\cref{lem:structure}). For each $w_i$,
\[
f_i(\alpha)=m(w_i-\alpha\,\sigma_{\mathrm{ref}}x_{\mathrm{ref}})=m(w_i)-\alpha M=M(\xi_i-\alpha),\qquad\xi_i=\frac{m(w_i)}M.
\]
The learner does not know $M$, but $|\xi_i|\le1/M\le1/\tau$, so it searches $[-R,R]$ with $R=1/\tau$. Every query has norm at most $1+RN_0\le2N_0/\tau$, so its answer is correct whenever $M|\xi_i-\alpha|>2\eta N_0/\tau$. \Cref{lem:fbs} applies with $\Delta\le2\eta N_0/\tau^2\le\alpha_0/(8\ell)\le\alpha_0/4$ and gives $|\hat\xi_i-\xi_i|\le\alpha_0/2$. The cost per coordinate is $\lceil\log_2(4/(\tau\alpha_0))\rceil=O(L\log\ell+\log(\lambda d))=O(\log d\log(d/\delta_0))$, since $\ell$ and $\lambda$ are polynomial in $d/\delta_0$.

Write $y=\sum_ja_jv_j+\sum_ib_iw_i$ with $(v_j)$ orthonormal in $V$, and put $E=\sum_ib_i\xi_i$. Then
\[
\frac{m(y)}M=\frac{\sum_ja_jm(v_j)}M+E.
\]
The coordinate error is $|E-\hat E|\le\sqrt d\,\alpha_0/2=(192\lambda d)^{-1}$, and by \cref{lem:dimreduce} the residual is at most $C_2\lambda^3d^{3/2}/\ell\le(96\lambda d)^{-1}$. Thus with $\varepsilon_{\mathrm{rec}}=(48\lambda d)^{-1}$, $|m(y)/M-\hat E|\le\varepsilon_{\mathrm{rec}}$. Whenever $|\hat E|\ge2\varepsilon_{\mathrm{rec}}$, $m(y)$ has the sign of $\hat E$, so every label assigned is correct. Every $y$ with $|m(y)|\ge t=M/(8\lambda\sqrt d)\ge3\varepsilon_{\mathrm{rec}}M$ is labeled (in LargeMargin, every $y$ with $|m(y)|\ge M/(12\lambda)$), giving coverage at least $k/d$. The binary searches cost $O((k+1)\log d\log(d/\delta_0))$, StructureSearch$'$ costs $O(r^2\log d)=O(\log d\log^2(d/\delta_0))$, and DimReduce$'$ costs $O((k+1)\log(d/\delta_0))$, for a total of $O((k+1)\log d\log^2(d/\delta_0))$. Wrong labels can arise only on the unobserved bad events of \cref{lem:structure,lem:dimreduce} and the representative calls, of total probability at most $\delta_0$. An observed failure produces no labels.
\end{proof}

\subsection{Global bookkeeping for PartialLearn\texorpdfstring{$'$}{'}, WeakLearn\texorpdfstring{$'$}{'} and boosting}\label{sec:global}
Fix the ambient dimension $n$, set $\delta_0=n^{-C}$ for a constant $C\ge4$, and use the single tolerance
\[
\eta^*_n:=\eta^*(n,n^{-C})=n^{-O(\log n)}\ \le\ \eta^*(d,n^{-C})\qquad\text{for every }2\le d\le n,
\]
by monotonicity. This is the tolerance $\eta^*$ of \cref{prop:robust-pl}.

\paragraph{PartialLearn$'$ and WeakLearn$'$.} Given current weights $\omega$ on the unlabeled points of $X$, PartialLearn$'$ chooses from $\omega$ alone a subspace $V_0$ of some dimension $d\le n$ with $\omega(X\cap V_0)\ge d/n$ and a transform $T$ putting the normalized restriction in $1/4$-isotropic position \citep[Lemma~4.19, Cor.~4.20]{HKLM20}. Since $\omega(X\cap V_0)>0$, $V_0$ contains a point of $X$, and since no point of $X$ lies on the target hyperplane, $P_{V_0}w^\star\ne0$. So $(V_0,T)$ is a query instance in the sense of \cref{def:sr-oracle}, the oracle is used only on such instances, and on each of them it is $\eta^*_n$-subspace-relative with $\eta^*_n\le\eta^*(d,\delta_0)$. (In the pool simulation, \cref{lem:pool-oracle} provides these answers.) If $d=1$, one query labels all of $X\cap V_0$. Otherwise PartialLearn$'$ runs IsoLearn$'$ in these coordinates with the ambient parameter $\delta_0=n^{-C}$, not $d^{-C}$. WeakLearn$'$ runs PartialLearn$'$ repeatedly on the residual weights of unlabeled points, with a deterministic total query budget, and discards any call that the budget interrupts.

\ifextendedproofs
The local lemmas above establish the error, coverage, and query bounds used in the proof of \cref{prop:robust-pl} in \cref{sec:robust-main-proof}. That proof also gives the deterministic weak-learning budget and verifies the hypotheses of the final boosting reduction.
\fi

\section{Necessity proofs and the explicit margin floor}\label{app:lower}
\subsection{Necessity of the angular assumptions}\label{app:necessity}
\begin{table}[h]
\centering\small
\begin{tabular}{>{\raggedright\arraybackslash}p{2.6cm}>{\raggedright\arraybackslash}p{\dimexpr(\linewidth-2.6cm-5.2cm-6\tabcolsep)/1\relax}>{\raggedright\arraybackslash}p{5.2cm}}
\toprule
Assumption removed & Assumption kept & Labels, pool vs.\ membership queries (unlimited unlabeled data)\\
\midrule
Symmetric full support & Hyperplane-nullness, even with a smooth symmetric density & $\Omega(1/\eps)$ vs.\ $O(\log(1/\eps))$\\
Hyperplane-nullness & Full support & $\Omega(1/\eps)$ in expectation vs.\ $O(\log(1/\eps))$\\
None & Both & No label gap. The unlabeled cost cannot be bounded uniformly (\cref{prop:finite-budget,thm:affine-core})\\
\bottomrule
\end{tabular}
\caption{Role of each angular assumption. Without symmetric full support, unlimited unlabeled data cannot provide the informative directions. When both assumptions hold, informative directions are eventually available, but they can be so rare that no unlabeled budget fixed in advance finds them.}
\label{tab:necessity}
\end{table}
Both constructions use points on the moment curve. For $m\ge2$ put
\[
t_i=\frac{i}{m+1},\qquad p_i=(t_i,t_i^2,1)\in\R^3,\qquad \Delta=\frac{1}{4(m+1)^2},\qquad v_i=(-2t_i,1,t_i^2-\Delta),
\]
so that
\begin{equation}\label{eq:moment-curve}
\langle v_i,p_j\rangle=(t_j-t_i)^2-\Delta\quad\text{is negative exactly when } i=j.
\end{equation}
Recall the convention $\sign(0)=+1$ (\cref{sec:models}). \Cref{rem:tilts} discusses the opposite convention.

\begin{proof}[Proof of \cref{prop:need-support}]
First let $D_m$ be uniform on $p_1,\dots,p_m$, and set $h_0(x)=\sign(x_3)$ and $h_i(x)=\sign\langle v_i,x\rangle$. By \eqref{eq:moment-curve}, $h_0(p_j)=+1$ for all $j$, while $h_i(p_j)=-1$ exactly when $j=i$. Each alternative target changes the label of exactly one support point.

\emph{Pool lower bound.} Suppose a randomized learner uses at most $q$ labels and succeeds for every target. Run it under $h_0$, and let $E_i$ be the event that it queries a copy of $p_i$. Then $\sum_i\Pr(E_i)\le q$. Couple runs under $h_0$ and $h_i$ using the same unlabeled observations and random seed. Unless $E_i$ occurs, the transcripts and outputs coincide. But $D_m(h_i\ne h_0)=1/m>2\eps$, so no output is $\eps$-accurate for both targets. Hence $\Pr(\text{failure under }h_i)\ge 3/4-\Pr(E_i)$, and success with probability $3/4$ under $h_i$ forces $\Pr(E_i)\ge1/2$. Summing gives $q\ge m/2$.

\emph{Membership-query upper bound.} Query $z=(0,-1,1/2)$. It is positive under $h_0$, while under every $h_i$, $\langle v_i,z\rangle=-1+\tfrac12(t_i^2-\Delta)<0$. If the answer is negative, query $z_a=(1,2a,0)$ for $a$ strictly between consecutive $t_i$'s. Then $h_i(z_a)=\sign(2a-2t_i)$, which tests whether the unknown index lies below or above the cut. Binary search identifies it with $\lceil\log_2m\rceil$ further queries.

\emph{Smoothing and symmetry.} All labels at the support points have strict margins. Replace each $p_i$ by a small ball $B_i$ on which every $h_j$ keeps its label, and place a smooth bump of mass $1/m$ in each. A pool query in $B_i$ still tests only whether the target is $h_i$, so both arguments apply unchanged. Adding reflected bumps on $-B_i$ with equal weights makes the law centrally symmetric. Since $h(-x)=-h(x)$ for every homogeneous halfspace $h$ and every $x$ off its boundary, a query in $-B_i$ is informative only about $h_i$, and $D(h_i\ne h_0)$ is unchanged.
\end{proof}

\begin{proof}[Proof of \cref{prop:need-null}]
Write points as $(y,z)\in\R^3\times\R$ and let
\[
D_m=\tfrac12\mathcal N(0,I_4)+\tfrac1{2m}\textstyle\sum_{j=1}^m\delta_{(p_j,0)}.
\]
The Gaussian component gives full angular support, but $H=\{z=0\}$ has probability $1/2$. Consider $h_0(y,z)=\sign(z)$ and $h_{i,\tau}(y,z)=\sign(z+\tau\langle v_i,y\rangle)$ for $0<\tau\le\bar\tau$. Every atom is positive under $h_0$. Under $h_{i,\tau}$ exactly atom $i$ is negative, however small $\tau>0$ is. Hence
\begin{equation}\label{eq:null-sep}
D_m(h_{i,\tau}\ne h_0)\ge\tfrac1{2m}>2\eps.
\end{equation}

\emph{Pool lower bound.} Run the learner under $h_0$. Let $E_i$ be the event that it queries atom $i$ and $q$ its label count, so $\sum_i\Pr(E_i)\le\E_{h_0}q$. Fix $i$ and couple runs under $h_0$ and $h_{i,\tau}$ using the same infinite unlabeled sequence and seed. Almost surely every Gaussian pool point has $z\ne0$, and each terminating run queries finitely many points. For every queried Gaussian point $\sign(z+\tau\langle v_i,y\rangle)=\sign(z)$ for all sufficiently small $\tau$. If $A_{i,\tau}$ is the event that some Gaussian point queried along the $h_0$ run changes label under $h_{i,\tau}$, bounded convergence gives $\Pr(A_{i,\tau})\to0$ as $\tau\downarrow0$. On $E_i^c\cap A_{i,\tau}^c$ the transcripts and outputs agree, so by \eqref{eq:null-sep}, $\Pr(\text{failure under }h_{i,\tau})\ge 3/4-\Pr(E_i)-\Pr(A_{i,\tau})$. The success guarantee bounds the left side by $1/4$. Letting $\tau\downarrow0$ yields $\Pr(E_i)\ge1/2$, and summing gives $\E_{h_0}q\ge m/2$.

\emph{Membership-query upper bound.} Membership queries may lie in $H$. Query $((0,-1,1/2),0)$. It is positive under $h_0$ and negative under every alternative. For an alternative, query $((1,2a,0),0)$, whose label is $\sign(2\tau(a-t_i))$. Binary search finds $i$ with $\lceil\log_2 m\rceil$ labels. We need not identify $\tau$. With $\bar\tau=\eps/\max_i\|v_i\|$, output $h_{i,\bar\tau}$. It agrees with the target on all atoms, and on the Gaussian component the disagreement is the angle between normals divided by $\pi$, at most $\bar\tau\|v_i\|/\pi\le\eps/\pi$. Its total error is below $\eps$.
\end{proof}

\begin{remark}\label{rem:tilts}
The second construction uses the fixed boundary convention. With $\sign(0)=-1$, reversing the relevant signs gives the analogous example. A target that tilts by an arbitrarily small amount changes the label of a positive-mass atom immediately, while leaving every point in any finite collection of off-hyperplane queries unchanged. Membership queries evaluate the decisive comparisons exactly on the hyperplane. The family must contain arbitrarily small tilts. With a known positive minimum tilt, sufficiently precise off-hyperplane pool queries could reproduce the decisive comparisons, so a finite family would not suffice for this argument.
\end{remark}

\subsection{An explicit margin floor and the running-time status}\label{sec:explicit-floor}
\begin{lemma}[Explicit empirical floor]\label{lem:explicit-floor}
Let $Y_1,\dots,Y_{M_0}\in\Sph$ be i.i.d.\ from a hyperplane-null angular law. For $S\subseteq[M_0]$ with $|S|=n$, let $Y_S$ be the $n\times n$ matrix with rows $Y_i^\top$, $i\in S$, and define $\theta_{\mathrm{alg}}=\frac{1}{2\sqrt n}\min_{|S|=n}\sigma_{\min}(Y_S)$. Then $\theta_{\mathrm{alg}}>0$ almost surely and no band of width $\theta_{\mathrm{alg}}$ contains $n$ sample points, so every such band has empirical mass at most $(n-1)/M_0$. With the sample size $M_0$ of \cref{lem:floor}, the same relative VC bound gives $\rho_\mu(\theta_{\mathrm{alg}})\le\delta/(8N)$ with probability at least $1-\delta/8$.
\end{lemma}
\begin{proof}
Hyperplane-nullness implies every fixed $n$-tuple is linearly independent almost surely. A finite union makes this simultaneous over all subsets. If a band contained the $n$ points indexed by $S$, then for its unit normal $w$, $\|Y_Sw\|_2<\sqrt n\,\theta_{\mathrm{alg}}\le\frac12\sigma_{\min}(Y_S)$, contradicting $\|Y_Sw\|_2\ge\sigma_{\min}(Y_S)$. The empirical and population bounds follow from the same relative-deviation estimate as in \cref{lem:floor}.
\end{proof}
The direct implementation enumerates $\binom{M_0}{n}$ subsets and costs $M_0^{O(n)}$ arithmetic operations. \Cref{thm:main} is a statement about labels and unlabeled examples, and we make no running-time claim for it. Apart from the floor, the queries and tests of \cref{app:robust} use only linear algebra on the current sample. The remaining step is to find the dense subspace and approximate isotropic transform of \citet[Cor.~4.20]{HKLM20}, whose existence is all that \citet{HKLM20} need, since they measure query depth. Algorithms for approximate Forster and radial-isotropic transforms exist \citep{HM13,DTK23}, but we have not verified that they cover the weighted subspace search, the precision, and the boosted distributions required here. The explicit floor is dimension-exponential, and the cap-waiting cost in \cref{thm:unlabeled} can be arbitrarily large. \Cref{prop:finite-budget,thm:affine-core} show that no distribution-independent algorithmic optimization can remove the latter dependence.

\section{Proofs for \texorpdfstring{$s$}{s}-concave marginals}\label{app:sconcave}
This appendix proves the results summarized in \cref{sec:sconcave}. \citet{BZ17} developed their tools and sample-based active learners for $-1/(2n+3)\le s\le0$. Several of the underlying geometric and moment facts are classical. This appendix applies known convex geometry to wider ranges of $s$, gives the matching sharpness examples, and constructs the symmetric endpoint example of \cref{sec:sym-endpoint}. Throughout, $\chi=-1/s-(n+2)$ is the slack of \cref{sec:sconcave}.

\subsection{Preliminaries for \texorpdfstring{$s$}{s}-concavity}
We use \cref{def:sconcave} and the classes $\Dns$ and $\Dsym$ of \cref{sec:sconcave}. Examples of the symmetric class include centered Gaussian and Student-$t$ laws with finite covariance and uniform laws on centrally symmetric convex bodies. Throughout the structural part, $D$ is isotropic, and we write $p=-1/s$, so that $\chi=p-(n+2)$.

\begin{lemma}[Marginals \citep{BL76,BZ17}]\label{lem:marginals}
If $f$ is $s$-concave on $\R^n$ and $s>-1/(n-k)$, then each $k$-dimensional marginal is $s_k$-concave with $s_k=\frac{s}{1+(n-k)s}$ and $-1/s_k=\chi+k+2$. If $f$ is isotropic, so is the marginal. Thus all low-dimensional marginals preserve the same slack $\chi$.
\end{lemma}

\begin{lemma}[Entropy lower bound]\label{lem:entropy}
Every isotropic density $f$ on $\R^k$ satisfies $\sup f\ge(2\pi e)^{-k/2}$.
\end{lemma}
\begin{proof}
The standard Gaussian maximizes differential entropy at identity covariance, while $H(f)\ge-\log\sup f$.
\end{proof}
For an isotropic one-dimensional marginal with density $g$ and slack $\chi>0$, write $G(\chi)=\bigl(\frac{\chi+1}{\chi+2}\bigr)^{\chi+2}$ and $\bar f_1(\chi)=\frac{\chi+2}{\chi}$. The one-dimensional inputs from \citet{BZ17} give $\Pr[X\ge0]\ge G(\chi)$ and $\sup g\le\bar f_1(\chi)$ throughout $\chi>0$.

\subsection{Unlimited-pool learning has no finite-\texorpdfstring{$s$}{s} threshold}
Let $\Lambda^*(\eps,\delta;\Dns)$ denote the minimax number of labels that a pool learner with unlimited unlabeled data needs to reach error $\eps$ with probability at least $1-\delta$, uniformly over $D\in\Dns$ and homogeneous targets.
\begin{corollary}[All finite $s$]\label{cor:all-s}
For every $n\ge2$ and every finite $s\le0$,
\[
c(n-1)\log(1/\eps)\le\Lambda^*(\eps,\delta;\Dns)\le Cn\log^3n\log\frac{n}{\eps\delta}
\]
for constant confidence. In particular, with unlimited unlabeled data the minimax label complexity is $\Theta(n\log(1/\eps))$ up to a $\log^3n$ factor (and logarithmic confidence terms), for every finite $s$.
\end{corollary}
\begin{proof}
A full-dimensional $s$-concave law has convex support and its mean lies in the interior. The density is positive and continuous on a neighborhood of the mean. After centering at the mean, the angular law is therefore absolutely continuous with respect to spherical measure and every open cap receives positive probability. The upper bound follows from \cref{thm:main}. For the lower bound, take the uniform distribution on a Euclidean ball, which is $s$-concave for every $s\le0$. Angular disagreement is proportional to the angle between target normals, so a spherical packing and the information bound of \citet{KMT93} give $\Omega((n-1)\log(1/\eps))$ labels.
\end{proof}

\subsection{The full Balcan--Zhang structural range}\label{sec:bz-range}
\subsubsection{Density near the center}
For an isotropic $s$-concave density on $\R^k$ with slack $\chi>0$, define $a_k=(2-2^{-(k+1)s})^{-1/s}$ and $r_\chi=\frac{G(\chi)}{3\bar f_1(\chi)}$.

\begin{lemma}[Density near the center]\label{lem:center}
For every $x$ with $\|x\|\le r_\chi$, $f(x)\ge a_kf(0)$.
\end{lemma}
\begin{proof}
Suppose $\|v\|\le r_\chi$ and $f(v)<a_kf(0)$. The superlevel set $C=\{x:f(x)\ge a_kf(0)\}$ is convex, contains a neighborhood of the origin, and does not contain $v$. By strict separation, after a rotation there is $0<a\le\|v\|\le r_\chi$ such that $C\subseteq\{x_1\le a\}$. Hence $x_1>a$ implies $f(x)<a_kf(0)$. For such $x$, $s$-concavity at the midpoint gives $f(2x)^s\ge2f(x)^s-f(0)^s$. Since $s<0$ and $f(x)<a_kf(0)$, $f(0)^s<f(x)^s/a_k^s=(2-2^{-(k+1)s})f(x)^s$, so $f(2x)<2^{-(k+1)}f(x)$. Let $g$ be the first marginal. Integrating over slices and changing variables yields $g(2y)<\frac14g(y)$ for $y>a$. Therefore $\int_a^\infty g<2\int_a^{2a}g\le2a\bar f_1$ and $\int_0^ag\le a\bar f_1$, so $\Pr[X_1\ge0]<3a\bar f_1\le3r_\chi\bar f_1=G(\chi)$, contradicting the one-dimensional Gr\"unbaum bound.
\end{proof}

\begin{corollary}[Center from above]\label{cor:center-above}
With $\omega_k$ the volume of the unit $k$-ball, $f(0)\le\bar f_0:=\frac{1}{a_k\omega_kr_\chi^k}$.
\end{corollary}

\subsubsection{The sharp centroid-to-mode inequality}
\begin{theorem}[Centroid-to-mode inequality \citep{AGLLOPT12}]\label{thm:centroid}
Let $f$ be an $s$-concave density on $\R^k$ with $-1/(k+1)<s<0$, finite first moment, and barycenter $b$. Then $f(b)\ge\bigl(\frac{1+s}{1+(k+1)s}\bigr)^{1/s}\sup f$, and the constant is sharp.
\end{theorem}
\begin{proof}
Translate so that $b=0$ and put $K=\{f>0\}$, $\phi=f^s$, convex and positive on $K$, and $I=\int_Kf^{1+s}=\E\phi(X)$. By Jensen, $\phi(0)\le I$. Fix $z\in\operatorname{int}K$ and exhaust $K$ by compact convex sets $K_j\Subset\operatorname{int}K$ containing $z$. At a.e.\ $x\in K_j$, $\phi(z)\ge\phi(x)+\langle\nabla\phi(x),z-x\rangle$. Multiply by $f(x)$ and integrate over $K_j$. With $F=f^{1+s}$ and $f\nabla\phi=\frac{s}{1+s}\nabla F$, integration by parts gives
$\int_{K_j}f\langle\nabla\phi,z-x\rangle=\frac{s}{1+s}\bigl(\int_{\partial K_j}F\langle z-x,\nu\rangle dS+k\int_{K_j}F\bigr)$, where $\nu$ is the outer unit normal. Convexity of $K_j$ makes $\langle z-x,\nu\rangle\le0$ on $\partial K_j$, and $s/(1+s)<0$, so $\phi(z)\int_{K_j}f\ge\frac{1+(k+1)s}{1+s}\int_{K_j}F$. Letting $K_j\uparrow K$ gives $\phi(z)\ge\frac{1+(k+1)s}{1+s}I$, so $I<\infty$ and $\phi(0)\le\phi(z)\frac{1+s}{1+(k+1)s}$. Raising to the power $1/s<0$ and taking the supremum over $z$ gives the claim. For sharpness, take $C=\{x:x_1\ge0,\|(x_2,\dots,x_k)\|\le x_1\}$ and $f(x)=c(1+x_1)^{1/s}\mathbf 1_C(x)$. Then $f^s$ is affine on $C$ and a beta-integral calculation gives equality.
\end{proof}
\begin{remark}[Prior statements]
\Cref{thm:centroid} is not new. It appears in \citet[Lemma~7.2]{AGLLOPT12} and is restated in \citet[Cor.~7.7]{FLM20} with a R\'enyi-entropy proof. The resulting bound on $\|f\|_\infty$ was also obtained by \citet[Prop.~2.6]{MP21}. We include the short proof because it shows the dependence on $s$ and the exact threshold.
\end{remark}
For a $k$-dimensional marginal with slack $\chi$, write $C_k(\chi)=\bigl(\frac{\chi+k+1}{\chi+1}\bigr)^{\chi+k+2}$. Then \cref{thm:centroid} says $\sup f\le C_k(\chi)f(0)$.

\subsubsection{Full-range density and tail bounds}
Let $\underline f_0=(2\pi e)^{-k/2}/C_k(\chi)$, $\bar f_{\max}=C_k(\chi)\bar f_0$, and $\ell_k=a_k\underline f_0$.
\begin{theorem}[Full-range density bounds]\label{thm:density}
For every isotropic $s$-concave density on $\R^k$ with slack $\chi>0$: (i) $f(0)\le\bar f_0$, (ii) $\sup f\le C_k(\chi)f(0)$, (iii) $f(0)\ge\underline f_0$, (iv) $\sup f\le\bar f_{\max}$, (v) $f(x)\ge\ell_k$ for $\|x\|\le r_\chi$. Moreover every line integral through the origin is bounded by the corresponding $(k-1)$-dimensional value of $\bar f_0$.
\end{theorem}
\begin{proof}
Items (i) and (v) are \cref{lem:center,cor:center-above}, (ii) is \cref{thm:centroid}, (iii) combines (ii) with \cref{lem:entropy}, (iv) combines (i) and (ii). A line integral is the value at zero of an orthogonal marginal, whose slack is still $\chi$.
\end{proof}

\begin{lemma}[Polynomial tail]\label{lem:tail}
Let $p=-1/s$, let $I_{\max}$ bound the line integrals through the origin, and put $R_0=\frac{2^pI_{\max}}{\underline f_0}$, $c_*=\frac{(1-2^{1-p})\underline f_0^{1-1/p}}{(p-1)I_{\max}}$, and $A=\max\{2^p\bar f_{\max},(2/(c_*R_0))^p\}$. Then $f(v)\le A(1+\|v\|/R_0)^{-p}$.
\end{lemma}
\begin{proof}
Fix $v$ and write $R=\|v\|$, $a=f(0)^{-1/p}$, $b=f(v)^{-1/p}$. Along the segment, $s$-concavity gives $f(tv)\ge((1-t)f(0)^s+tf(v)^s)^{1/s}$. If $b<2a$, the segment density is at least $2^{-p}\underline f_0$, so $R\le R_0$. If $b\ge2a$, evaluating the segment integral gives $I_{\max}\ge R\frac{a^{1-p}-b^{1-p}}{(p-1)(b-a)}\ge R\frac{(1-2^{1-p})\underline f_0^{1-1/p}}{(p-1)b}$, so $b\ge c_*R$ and $f(v)\le(c_*R)^{-p}$. For $R\le R_0$ use $f(v)\le\bar f_{\max}$ and $A\ge2^p\bar f_{\max}$. For $R>R_0$ use $1+R/R_0\le2R/R_0$ and $A\ge[2/(c_*R_0)]^p$.
\end{proof}

\subsubsection{Explicit transfer of the Balcan--Zhang analysis}
Write $r_0=r_\chi$. For the $k$-dimensional marginal $f_k$, let $\ell_k>0$ be its lower density bound on the ball of radius $r_0$ around the origin, and choose $A_k,R_k$ in \cref{lem:tail}, so $f_k(x)\le A_k(1+\|x\|/R_k)^{-(\chi+k+2)}$.

\begin{theorem}[Full-range bounds for the Balcan--Zhang analysis]\label{thm:toolkit}
For an isotropic $s$-concave distribution with $-1/(n+2)<s\le0$ and unit vectors $u,v$:
\begin{enumerate}[label=(\roman*)]
\item $\ell_1t\le\Pr[|\langle u,X\rangle|\le t]\le2\bar f_1(\chi)t$ for $0<t\le r_0$,
\item $\Pr[h_u(X)\ne h_v(X)]\ge c_\chi\angle(u,v)$,
\item for every $b>0$ there is $B=B(\chi,b)$ such that, whenever $\angle(u,v)<\pi/2$, $\Pr[h_u(X)\ne h_v(X),|\langle u,X\rangle|\ge B\angle(u,v)]\le b\angle(u,v)$,
\item if $\|a\|\le1$, $\|a-u\|\le r$, and $0<t\le r_0$, then $\E[\langle a,X\rangle^2\mid|\langle u,X\rangle|\le t]\le C_\chi(r^2+t^2)$,
\item for an intersection $E$ of at most three origin-centered halfspaces, $\Pr[-E]\le K_\chi\Pr[E]$.
\end{enumerate}
\end{theorem}
\begin{proof}
Item (i) follows from the one-dimensional density bounds. Write $\varphi=\angle(u,v)$. For (ii), integrate $\ell_2$ over the disagreement wedge in the disk of radius $r_0$, of area $r_0^2\varphi$. For (iii), the wedge has angular width $2\varphi$ and its points outside the band have radius at least $B$, since $|\langle u,x\rangle|\le\|x\|\sin\varphi$ on the wedge. Its probability is at most $2\varphi A_2\int_B^\infty y(1+y/R_2)^{-(\chi+4)}dy$, which tends to zero as $B\to\infty$. For (iv), decompose $a=a_uu+a_\perp$ with $|a_u|\le1$, $\|a_\perp\|\le r$. With $Y$ the coordinate along $a_\perp$ and $Z=\langle u,X\rangle$, the two-dimensional envelope gives $\E[Y^2\mathbf 1\{|Z|\le t\}]\le2tA_2\int_\R y^2(1+|y|/R_2)^{-(\chi+4)}dy<\infty$ since $\chi+4>3$. Divide by the band mass in (i) and use $\langle a,X\rangle^2\le2t^2+2r^2Y^2$. For (v), project onto the span of the normals, of dimension $k\le3$. Both projected sets are cones with the same spherical area $\omega(E)$, and their probabilities lie between $\omega(E)\ell_kr_0^k/k$ and $\omega(E)A_k\int_0^\infty y^{k-1}(1+y/R_k)^{-(\chi+k+2)}dy$.
\end{proof}
The marginalization condition causes no further restriction. For $k\le n$, $1+(n-k)s>0$ and $-1/s_k=\chi+k+2$. \Cref{tab:bz} records the substitutions into \citet{BZ17}. Their old numerical constants are not retained.

\begin{table}[h]
\centering\small
\begin{tabular}{>{\raggedright\arraybackslash}p{5.2cm}>{\raggedright\arraybackslash}p{\dimexpr\linewidth-5.2cm-4\tabcolsep\relax}}
\toprule
Input from \citet{BZ17} & Replacement (items of \cref{thm:toolkit})\\
\midrule
Theorems 9, 11 (density, bands) & Inner-ball density $\ell_k$, upper density $\bar f_{\max}$, and item (i)\\
Theorem 12 (angle vs.\ error) & Item (ii), with $c_\chi=\ell_2r_0^2$\\
Theorem 13 (outside-band error) & Item (iii), with $B$ from the radial-tail integral\\
Theorem 14 (conditional second moment) & Item (iv)\\
Lemma 28 (three-halfspace comparison) & Item (v)\\
Theorem 5 (norm tail) & Integration of the $n$-dimensional radial envelope\\
\bottomrule
\end{tabular}
\caption{Substitutions into the Balcan--Zhang analysis.}
\label{tab:bz}
\end{table}
These inputs justify the realizable margin-localization argument of \citet[Theorem~15]{BZ17}, with angle control from (ii), a constant band multiplier from (iii), and VC generalization on the conditional sample using (i). Each round reduces error by a constant factor with $O_{n,s}(n\log(1/\eps)+\log(1/\delta))$ conditional labels, so the total is polylogarithmic in $1/\eps$. Items (iv) and (v) provide the structural inputs for their noisy localization and intersection algorithms, whose noise parameters and confidence schedules we do not re-derive.

\subsection{Sharpness of the structural threshold}
\begin{proposition}[No uniform Balcan--Zhang bounds at or below $-1/(n+2)$]\label{prop:struct-sharp}
Fix $n\ge1$ and $s\le-1/(n+2)$. Over isotropic $s$-concave densities on $\R^n$, each of $\sup f$, $f(0)$, and $\sup_{\|w\|=1,t>0}\Pr[|\langle w,X\rangle|\le t]/t$ is unbounded. The examples are rotationally symmetric, so symmetry does not remove this obstruction.
\end{proposition}
This rules out the Balcan--Zhang bounds as a whole with uniform constants, through their density and band upper bounds, although some individual estimates survive. For a rotationally symmetric law with no atom at the origin, $D(h_u\ne h_v)=\angle(u,v)/\pi$, so the angle-versus-error comparison holds exactly for these examples.
\begin{proof}
It suffices to work at $s_0=-1/(n+2)$, since $s_0$-concavity implies $s$-concavity for $s\le s_0$. For $T\ge2$ let $f_T(x)=c_T(1+\|x\|)^{-(n+2)}\mathbf 1\{\|x\|\le T\}$. Since $f_T^{s_0}$ is proportional to $1+\|x\|$ on the ball and $+\infty$ outside, $f_T$ is $s_0$-concave. The constants $c_T$ converge to a positive limit since $\int_0^\infty r^{n-1}(1+r)^{-(n+2)}dr<\infty$. Rotational symmetry gives covariance $\sigma_T^2I$ with $\sigma_T^2\asymp\int_0^Tr^{n+1}(1+r)^{-(n+2)}dr\asymp\log T\to\infty$. The isotropic rescaling $\tilde f_T(y)=\sigma_T^nf_T(\sigma_Ty)$ has $\tilde f_T(0)=\sigma_T^nc_T\to\infty$. For $n\ge2$, $f_T$ is bounded below on the ball of radius $2$, so with $t=1/\sigma_T$, $\Pr_{\tilde f_T}[|Y_1|\le t]=\Pr_{f_T}[|X_1|\le1]\ge c_n$, and the normalized band mass is at least $c_n\sigma_T\to\infty$. The case $n=1$ is identical.
\end{proof}

\subsection{Angular geometry below the structural threshold}\label{sec:angular}
\subsubsection{Ball--Bobkov bodies}
We write $\varrho$ for radial functions, to distinguish them from the band mass $\rho_\mu$.

\begin{lemma}[Ball--Bobkov bodies \citep{Bal88,Bob10}]\label{lem:bobkov}
Let $f$ be an $s$-concave density on $\R^n$ with $-1/n<s\le0$ and with the origin in the interior of its support, and for $q>0$ let $\varrho_q(u)=\bigl(q\int_0^\infty r^{q-1}f(ru)\,dr\bigr)^{1/q}$ for $u\in\Sph$ and $K_q=\{ru:u\in\Sph,\ 0\le r\le\varrho_q(u)\}$. Then $K_q$ is a convex body whenever $0<q\le-1/s-1$. In particular, $K_n$ is convex for $s\ge-1/(n+1)$, and $K_{n-1}$ is convex throughout the range $-1/n<s\le0$ (the endpoint $s=-1/n$ is not covered and is not used). Moreover, the angular law of $f$ has spherical density proportional to $\varrho_n^n$, so it is exactly the cone measure of $K_n$.
\end{lemma}
\begin{proof}
An $s$-concave density with $s>-1/n$ defines an $s/(1+ns)$-concave measure \citep{Bor75}. \citet[Theorem~2.1]{Bob10} shows that $x\mapsto(\int_0^\infty r^{q-1}f(rx)dr)^{-1/q}$ is a norm (a gauge, even without symmetry, by his Remark~2.6) whenever $0<q\le n-1-(1+ns)/s$, that is, $q\le-1/s-1$. For log-concave $f$ this is Ball's theorem \citep{Bal88}. The last assertion is polar integration. The angular density at $u$ is $\int_0^\infty r^{n-1}f(ru)dr=\varrho_n(u)^n/n$.
\end{proof}

\subsubsection{Disagreement for \texorpdfstring{$s>-1/(n+1)$}{s > -1/(n+1)}}
\begin{lemma}[Convex-body disagreement transfer]\label{lem:cb-transfer}
Let $\nu$ be uniform on an isotropic convex body in $\R^n$, and suppose every halfspace whose boundary passes through a reference point $z$ has mass at least $\alpha>0$. Then halfspaces whose boundaries pass through $z$ satisfy $\theta_\nu(\eps)\le C\alpha^{-3}(1+\log(4/\alpha))^3\bigl(1+\sqrt n(1+\log(1/\eps))\bigr)$.
\end{lemma}
\begin{proof}
Every isotropic one-dimensional log-concave marginal has density at most an absolute constant $c_\infty$ and exponential tails \citep{BL13,Han14}. With the depth assumption this gives $\|z\|\le C\log(4/\alpha)$. For any two-dimensional projection with density $g$, the projected reference $z_2$ has depth at least $\alpha$, and every point within $c\alpha$ of $z_2$ has depth at least $3\alpha/4$. Choose $R=C\log(4/\alpha)$ so the projected norm tail outside the disk of radius $R$ has mass at most $\alpha/8$. With $t=\alpha/(8\pi R^2)$ the complement of $\{g\ge t\}$ has mass at most $\alpha/4$, so a point of depth $3\alpha/4$ lies in this superlevel set, and $g\ge t$ on the disk of radius $c\alpha$ around $z_2$. Integrating a disagreement wedge gives $\nu(h_u\ne h_v)\ge c'\alpha^3R^{-2}\angle(u,v)=:c_0\angle(u,v)$. A hypothesis at error at most $r$ has normal within angle $r/c_0$ of the target. On $\|X-z\|\le T$ all such disagreements lie in a target strip of width $CTr/c_0$, of mass at most $Cc_\infty Tr/c_0$. Take $T=C[\sqrt n(1+\log(1/r))+\log(4/\alpha)]$ so the norm tail is at most $r$, divide by $r$ and maximize over $r\ge\eps$.
\end{proof}

\begin{theorem}[Uniform disagreement above $-1/(n+1)$]\label{thm:disagree-gen}
Let $D$ be $s$-concave on $\R^n$ with mean $0$ and $-1/(n+1)<s\le0$, and $\chi=-1/s-(n+2)$. Then $\theta_D(\eps)=O_\chi(\sqrt n\log(1/\eps))$.
\end{theorem}
\begin{proof}
By \cref{lem:bobkov}, the angular law of $D$ is that of the uniform distribution on a convex body $K$. The Caplin--Nalebuff Gr\"unbaum inequality \citep{CN91} gives mass at least $G(\chi)$ to every halfspace through the mean. Since $K$ need not be centered at the origin, apply an affine map $A$ putting the uniform law on $K$ in isotropic position (mean zero, identity covariance), and let $z=A(0)$. Homogeneous halfspaces map to halfspaces whose boundaries pass through $z$, and pushforward preserves both the disagreement masses and the masses of halfspaces through the reference point. Now use \cref{lem:cb-transfer} with reference point $z$ and $\alpha=G(\chi)$.
\end{proof}

\begin{corollary}[Polynomial-time CAL in this range]\label{cor:cal}
For $-1/(n+1)<s\le0$, realizable CAL, run on a stream of $\tilde O_{n,s}(n/\eps)$ unlabeled examples, uses a number of labels polylogarithmic in $1/\eps$ (by the disagreement bound of \cref{thm:disagree-gen}), and can be implemented in polynomial time using linear-programming feasibility tests for version-space disagreement. The unlabeled bound is uniform over the class.
\end{corollary}

\subsubsection{The plane}
\begin{proposition}[No angular threshold in dimension two]\label{prop:plane}
For homogeneous halfspaces in $\R^2$, every distribution with an angular density has disagreement coefficient at most $4$.
\end{proposition}
\begin{proof}
Parameterize an oriented homogeneous halfspace by its boundary angle on the circle. Hypotheses within error $r$ of a target form an interval on each side of the target in angular probability coordinates, and their disagreement region lies in the union of the two corresponding arcs, of total probability at most $4r$.
\end{proof}

\subsubsection{Elliptical laws}
\begin{proposition}[No angular threshold for elliptical laws]\label{prop:elliptical}
Let $D$ be a full-dimensional elliptically symmetric distribution with finite nonsingular covariance and no atom at its center. After centering and whitening, its angular law is uniform on $\Sph$, and for homogeneous halfspaces $\theta_D(\eps)\le C\sqrt n$ for an absolute constant $C$. This holds independently of any $s$-concavity parameter.
\end{proposition}
\begin{proof}
An elliptical vector has a representation $X=x_0+ARV$ with $V$ uniform on the sphere and $R\ge0$ independent of $V$, where $\Pr(R=0)=0$ because $D$ has no atom at its center. Finite covariance is proportional to $AA^\top$, so whitening makes the law rotationally invariant and its direction uniform. The disagreement coefficient is invariant under the corresponding linear change of coordinates. Under uniform directions, two homogeneous halfspaces whose normals form angle $\alpha$ disagree with probability $\alpha/\pi$, so hypotheses within error $r<1/2$ have normals within angle $\pi r$ of $w^\star$ and their disagreement region lies in $\{u:|\langle w^\star,u\rangle|\le\sin(\pi r)\}$. The density of a spherical coordinate is $c_n(1-t^2)^{(n-3)/2}$ with $c_n=O(\sqrt n)$, so a band of half-width $t\le1/2$ has mass at most $C\sqrt n\,t$. If $\sin(\pi r)>1/2$ then $r>1/6$ and the trivial bound $1\le6r$ suffices. For $r\ge1/2$ the mass is at most $1\le2r$.
\end{proof}
Thus heavy tails alone cannot create an angular obstruction in the symmetric class. A counterexample must combine central symmetry with substantially non-elliptical angular geometry. \Cref{sec:sym-endpoint} provides one.

\subsubsection{Symmetric laws and adjacent radial moments}
Assume $n\ge2$, $f$ symmetric, $-1/n<s<0$, and normalize $f(0)=1$. For a ray $g(r)=f(ru)$, $u\in\Sph$, and $q>0$, define $\Phi(q)=\bigl(q\int_0^\infty r^{q-1}g(r)dr\bigr)^{1/q}$.

\begin{lemma}[Radial moments increase \citep{FGP14}]
If $0<q<q'$ and $0\le g\le1$, then $\Phi(q)\le\Phi(q')$.
\end{lemma}
\begin{proof}
Let $R=\Phi(q)$. Then $\int(g-\mathbf 1_{[0,R]})d(r^q)=0$, with integrand nonpositive on $[0,R]$ and nonnegative outside. Multiplying by the increasing $r^{q'-q}$ can only increase the integral.
\end{proof}

\begin{lemma}[Adjacent-moment comparison \citep{FGP14}]\label{lem:adjacent}
For $n\ge2$ and every direction $u$, $c_{n,s}^{1/(n-1)}\varrho_n(u)\le\varrho_{n-1}(u)\le\varrho_n(u)$, where $c_{n,s}>0$ depends only on $n,s$.
\end{lemma}
\begin{proof}
Apply \citet[Corollary~10]{FGP14} to $f$, which satisfies $f(0)=\|f\|_\infty=1$, with moment orders $n-1$ and $n$. Its first inclusion gives $\varrho_{n-1}\le\varrho_n$ and its second gives $\varrho_n\le B_{n,s}\varrho_{n-1}$, with the beta-function constant of \cref{thm:normalization}. Set $c_{n,s}=B_{n,s}^{-(n-1)}$.
\end{proof}
\begin{remark}[Classical radial-body comparison]
The monotonicity lemma is \citet[Theorem~7(1)]{FGP14} and the two-sided comparison is \citet[Corollary~10]{FGP14}. We apply them to angular learning and do not claim a new inclusion. Symmetry is needed only to place the reference point at the mode. The comparison with moment orders $n-1,n$ requires $n<-1/s$, which explains where the positive range stops. \Cref{sec:sym-endpoint} shows that this range is sharp.
\end{remark}

\begin{theorem}[Symmetric disagreement]\label{thm:sym-disagree}
If $D$ is symmetric and $s$-concave with $-1/n<s\le0$, then $\theta_D(\eps)=O_{n,s}(\log(1/\eps))$.
\end{theorem}
\begin{proof}
Since $f$ is even and quasi-concave, $f(0)=\|f\|_\infty$, and we normalize $f(0)=1$. By \cref{lem:bobkov} the body $K_{n-1}$ is convex, and it is symmetric because $f$ is even. By \cref{lem:adjacent}, $c_{n,s}^{1/(n-1)}\varrho_n\le\varrho_{n-1}\le\varrho_n$ pointwise. The angular law $\mu$ of $D$ has spherical density proportional to $\varrho_n^n$ (\cref{lem:bobkov}), and the cone measure $\nu$ of $K_{n-1}$ has density proportional to $\varrho_{n-1}^n$. Normalizing both gives $L^{-1}\le d\mu/d\nu\le L$ with $L=c_{n,s}^{-n/(n-1)}$. Homogeneous halfspaces, and their disagreement regions, have the same mass under $\nu$ and under the uniform law $\nu_K$ on $K_{n-1}$. By symmetry every halfspace through the origin has $\nu_K$-mass $1/2$. A linear map putting $\nu_K$ in isotropic position fixes the origin and preserves these masses, so \cref{lem:cb-transfer} with $z=0$ and $\alpha=1/2$ gives $\theta_\nu(\eps)\le C\sqrt n(1+\log(1/\eps))$. Finally $B_D(h^\star,r)\subseteq B_\nu(h^\star,Lr)$ and $D(\DIS(\mathcal H))\le L\,\nu(\DIS(\mathcal H))$ for every set $\mathcal H$ of hypotheses, so
\[
\theta_D(\eps)\le L\sup_{r\ge\eps}\frac{\nu(\DIS(B_\nu(h^\star,Lr)))}{r}=L^2\theta_\nu(L\eps)\le L^2\theta_\nu(\eps),
\]
since $L\ge1$ and $\theta_\nu$ is nonincreasing.
\end{proof}

\subsubsection{General laws and angular normalization}
For an invertible linear map $A$, let $\mu_A$ be the angular law of $A(X-x_0)$ and $\Lambda_D=\inf_A\|d\mu_A/d\sigma\|_\infty$.

\begin{theorem}[Angular normalization from the Fradelizi--Gu\'edon--Pajor comparison]\label{thm:normalization}
For $n\ge2$, every $s$-concave density with $-1/n<s<0$ and interior reference point $x_0$ with $f(x_0)>0$ satisfies
\[
\Lambda_D\le C(n)B_{n,s}^n\Bigl(\frac{\|f\|_\infty}{f(x_0)}\Bigr)^{1/(n-1)},\qquad B_{n,s}=\frac{[n\,\mathrm B(n,-1/s-n)]^{1/n}}{[(n-1)\,\mathrm B(n-1,-1/s-n+1)]^{1/(n-1)}},
\]
where $\mathrm B$ is the beta function. The log-concave case follows by the corresponding limit. If $x_0$ is the mean and $s\ge-1/(n+1)$, then $\Lambda_D\le C(n)$.
\end{theorem}
\begin{proof}
Translate $x_0$ to zero and normalize $\tilde f=f/f(x_0)$, so $\tilde f(0)=1$ and $R=\|f\|_\infty/f(x_0)$. Use the bodies of \citet{FGP14} with radial functions $\varrho_q(u)=(q\int_0^\infty t^{q-1}\tilde f(tu)dt)^{1/q}$. \citet[Corollary~10]{FGP14} with moment orders $n-1$ and $n$ and exponent $p=-1/s$ gives $R^{-1/[n(n-1)]}K_{n-1}\subseteq K_n\subseteq B_{n,s}K_{n-1}$. Bobkov's theorem makes $K_{n-1}$ convex in this range, so $\varrho_n^n/\varrho_{n-1}^n$ lies between $R^{-1/(n-1)}$ and $B_{n,s}^n$. Normalizing both angular densities gives an upper comparison factor $B_{n,s}^nR^{1/(n-1)}$. Use a linear normalization of a John ellipsoid of $K_{n-1}$. Its image satisfies $c+B_n\subseteq K'\subseteq c+nB_n$, and $0\in K'$ gives $\|c\|\le n$, $K'\subseteq2nB_n$, $\operatorname{vol}K'\ge\operatorname{vol}B_n$. The cone density relative to $\sigma$ is $\operatorname{vol}(B_n)\varrho_{K'}(u)^n/\operatorname{vol}(K')\le(2n)^n$, where $\varrho_{K'}$ is the radial function of $K'$. Only the linear part is applied, since translating the reference point would change the halfspace problem. If $s\ge-1/(n+1)$, $K_n$ itself is convex and the same normalization gives the second assertion.
\end{proof}

\begin{proposition}[$\Lambda_D$ is unbounded strictly below $-1/(n+1)$]\label{prop:lambda-unbounded}
Let $n\ge2$ and $p=-1/s\in(n,n+1)$. For every sufficiently large $\ell$ there is an $s$-concave law $D_\ell$ with mean $x_0$ such that $\Lambda_{D_\ell}\ge c_{n,p}\ell^{(n-1)(n+1-p)/n}$ and $f(x_0)/\|f\|_\infty\asymp_{n,p}\ell^{-p}$. The law may be taken isotropic.
\end{proposition}
\begin{proof}
Let $f_R(x)=c_R(1+(\|x\|-1)_+)^{-p}\mathbf 1_{S_R}(x)$ with $S_R=\{\|x\|\le R,\,x_1\le1\}$. Since $f_R^s=c_R^s\max\{1,\|x\|\}$ on the convex set $S_R$ and $+\infty$ outside is convex, $f_R$ is $s$-concave, and $c_R$ converges to a positive limit because $p>n$. By symmetry about the $e_1$ axis the mean is $-m_Re_1$ for some $m_R\ge0$. Comparing with the radial density, $m_R\le C_{n,p}R^{n+1-p}$, while restricting to a fixed cone in $\{x_1\le1\}$ gives $m_R\ge c_{n,p}R^{n+1-p}-O_{n,p}(1)$. By continuity, for all large $\ell$ there is $R>3\ell$ with $x_0=-\ell e_1$. The unit ball has probability at least $(p-n)/p$ and, seen from $x_0$, lies in the cone $C=\{u:\angle(u,e_1)\le2/\ell\}$, so $\mu(C)\ge c_{n,p}$. For $u\in E=\{u:\pi/3\le\angle(u,e_1)\le2\pi/3\}$ and $r\in[\ell,2\ell]$, $x_0+ru\in S_R$ and $f_R(x_0+ru)\ge c_{n,p}\ell^{-p}$, so $g=d\mu/d\sigma\ge c_{n,p}\ell^{n-p}$ on $E$.

Fix invertible $A$ with $|\det A|=1$ and $t=\|A\|\ge1$. With $\Phi_A(u)=Au/\|Au\|$ and Jacobian $\|Au\|^{-n}$, $g_A(\Phi_A(u))=g(u)\|Au\|^n$. A top right singular vector $v$ has $|\langle v,u\rangle|\ge1/2$ for some $u\in E$, so $\|g_A\|_\infty\ge c_{n,p}\ell^{n-p}t^n$. Also $\|Ae_1\|\ge t^{1-n}$, and when $t^n\le c_0\ell$, $\min_{u\in C}\|Au\|\ge\frac12t^{1-n}$. Since $\sigma(\Phi_A(C))\le C_n\ell^{-(n-1)}t^{n(n-1)}$ and $\mu_A(\Phi_A(C))\ge c_{n,p}$, $\|g_A\|_\infty\ge c_{n,p}\ell^{n-1}t^{-n(n-1)}$. The two bounds balance at $t=\ell^{(p-1)/n^2}$, where $t^n=o(\ell)$ because $p<n+1$, giving the claim. If $t^n>c_0\ell$ the first bound alone is stronger. Finally $f(x_0)/\|f\|_\infty=\ell^{-p}$, and an affine isotropizing map preserves $s$-concavity, this ratio, and $\Lambda_D$.
\end{proof}

\subsubsection{Directional search}
\begin{theorem}[Directional search with supplied normalization]\label{thm:dir-search}
Suppose the learner is supplied with an invertible map $A$ and a certified bound $L\ge1$ on the angular density of $A(X-x_0)$ relative to $\sigma$, and that this angular law has full support. Halfspaces whose boundaries pass through $x_0$ are learnable from an i.i.d.\ pool, in the realizable setting, with $O(n\log(nL/\eps))$ labels. The unlabeled cost is finite almost surely for each fixed marginal.
\end{theorem}
\begin{proof}
Work in transformed coordinates with unit target normal $w$. Choose $\gamma=\eps/(Cn^2L)$ and realize a unit query $q$ by waiting for a direction within Euclidean distance $\gamma$ of $q$. Its label is correct whenever $|\langle w,q\rangle|>\gamma$. Use the imprecise comparator of \cref{lem:comparator} to compare $|w_i|,|w_j|$ using $e_i,e_j$, with tolerance $4\gamma$. A sequential tournament with $n-1$ comparisons returns $k$ with $|w_k|\ge\max_i|w_i|-4n\gamma\ge1/(2\sqrt n)$, whose sign $\sigma_k$ is reliable. Set $R=2\sqrt n$ and search each ratio $\xi_i=w_i/w_k\in[-R,R]$ with the unit query $q_\alpha=\sigma_k(e_i-\alpha e_k)/\sqrt{1+\alpha^2}$, for which $\langle w,q_\alpha\rangle=|w_k|(\xi_i-\alpha)/\sqrt{1+\alpha^2}$. The answer is correct whenever $|\xi_i-\alpha|>\gamma\sqrt{1+R^2}/|w_k|$, a quantity at most $C_1n\gamma$. Imprecise binary search (\cref{lem:fbs}) to width $b=\eps/(C_2nL)$ estimates each ratio within $b$. The reconstructed vector $\hat w=\sigma_k(\hat\xi_1,\dots,\hat\xi_{k-1},1,\hat\xi_{k+1},\dots,\hat\xi_n)$ differs from $w/|w_k|=\sigma_k(\xi_1,\dots,1,\dots,\xi_n)$ by at most $\sqrt nb$, so its direction has angle $O(\sqrt nb)\le\eps/L$. A spherical lune of angle $\varphi$ has $\sigma$-mass $\varphi/\pi$, so the error is at most $L\varphi/\pi\le\eps$.
\end{proof}
\begin{remark}[Normalization is an input]
The result is implementable when $A,L$ are supplied, including for a known marginal. It does not assume that an optimal map or density bound can be recovered from unlabeled samples. Without such input, \cref{thm:main} remains available.
\end{remark}

\subsection{The symmetric endpoint}\label{sec:sym-endpoint}
We now show that the symmetric range of \cref{thm:sym-disagree} is sharp. The mean-shifting construction of \cref{sec:rare-tail} cannot be used. Central symmetry pins the mean at the symmetry center, and averaging a density with its reflection need not preserve $s$-concavity. Instead we use the fact that $(-1/n)$-concave densities have the form $f=Z^{-1}\phi^{-n}$ with $\phi$ positive and convex. This is the density representation of full-dimensional convex measures in Borell's sense \citep{Bor75,Bob10}. At this exponent, long intervals of \emph{logarithmic} radius can carry substantial angular mass, so different radial intervals can encode different angular distributions.

\begin{lemma}[$(-1/n)$-concave densities realize mixtures of angular laws]\label{lem:radial-mixture}
Let $N_1,\dots,N_m$ be norms on $\R^n$, and let $\nu_i$ have spherical density $\frac{d\nu_i}{d\sigma}(u)=N_i(u)^{-n}/Z_i$, $Z_i=\int_{\Sph}N_i^{-n}d\sigma$. Given positive weights $\pi_i$ summing to one, there are everywhere-positive, centrally symmetric, $(-1/n)$-concave densities with finite covariance whose angular laws converge in total variation to $\nu=\sum_i\pi_i\nu_i$.
\end{lemma}
\begin{proof}
Multiplying a norm by a positive constant does not change its normalized angular law, so we may rescale successively so that $N_1\le N_2\le\cdots\le N_m$ pointwise. Compute $Z_i$ after rescaling and set $a_i=\pi_i/Z_i$, $t_0=0$, $t_i=\sum_{j\le i}a_j$. For $T>0$ define
\[
\phi_T(x)=\max\Bigl\{1,\ N_1(x),\ N_2(x)-e^{Tt_1},\ \dots,\ N_m(x)-e^{Tt_{m-1}},\ e^{-3Tt_m}\|x\|^4\Bigr\}.
\]
This is positive, even, and convex as a maximum of even convex functions, so $f_T=\phi_T^{-n}/\int\phi_T^{-n}$ is symmetric and $(-1/n)$-concave ($f_T^{-1/n}\propto\phi_T$). For large $\|x\|$, $\phi_T(x)^{-n}\le e^{3nTt_m}\|x\|^{-4n}$, so $f_T$ is integrable with finite second moments ($r^{n+1}\cdot r^{-4n}$ is integrable at infinity), and its covariance is positive definite because the density is everywhere positive.

To identify the angular limit write $A_T(u)=\int_0^\infty r^{n-1}\phi_T(ru)^{-n}dr$, so the angular density of $f_T$ is proportional to $A_T$. Since $\phi_T\ge1$, the contribution from $r<1$ is at most $1/n$. On $r\ge1$ substitute $r=e^{Tt}$:
\[
\frac1T\int_1^\infty r^{n-1}\phi_T(ru)^{-n}dr=\int_0^\infty\Bigl(\frac{\phi_T(e^{Tt}u)}{e^{Tt}}\Bigr)^{-n}dt.
\]
For $t_{i-1}<t<t_i$, $\phi_T(e^{Tt}u)/e^{Tt}\to N_i(u)$. For $j\le i$ the offsets $e^{T(t_{j-1}-t)}$ vanish, for $j>i$ the terms $N_j(u)-e^{T(t_{j-1}-t)}$ become negative, the quartic term $e^{3T(t-t_m)}$ vanishes, and the largest remaining norm is $N_i$. On $[0,t_m]$ the integrand is at most $N_1(u)^{-n}$, bounded on the sphere. For $t>t_m$ the quartic term bounds the integrand by $e^{-3nT(t-t_m)}$, whose integral is $1/(3nT)$. Dominated convergence gives $A_T/T\to\sum_ia_iN_i^{-n}=\sum_i\pi_iN_i^{-n}/Z_i$ in $L^1(\sigma)$. The limit integrates to one, so normalization preserves the convergence, which is total-variation convergence of angular laws.
\end{proof}

Let $N$ be the gauge of an elongated ellipsoid with long axis $v$, for example $N(x)^2=\langle x,v\rangle^2+a^2\|P_{v^\perp}x\|^2$ with $a$ large. Its angular law concentrates near $\{v,-v\}$ as $a\to\infty$. At angle $\alpha$ from $\pm v$, $N(u)^{-n}=(\cos^2\alpha+a^2\sin^2\alpha)^{-n/2}$, so the mass outside a fixed antipodal cap of angle $\alpha_0$ is at most $(a\sin\alpha_0)^{-n}$ while the mass within angle $1/a$ is at least $c_na^{-(n-1)}$, and the ratio tends to zero. Thus $(-1/n)$-concave densities approximate any finite antipodally symmetric discrete angular law weakly. Total variation is used only for the intermediate, continuous mixture.

\begin{theorem}[No uniform symmetric disagreement bound at the endpoint]\label{thm:sym-dis}
For every $n\ge3$ and integer $m\ge4$ there are an isotropic, centrally symmetric, everywhere-positive, $(-1/n)$-concave distribution $D_m$ and a homogeneous target $h_0$ with $\theta_{D_m}(2/m;h_0)\ge3m/8$. Consequently, for every $s\le-1/n$, the class $\Dsym$ does not have logarithmic disagreement, and its worst-case disagreement coefficient at scale $\eps=2/m$ is $\Omega(1/\eps)$.
\end{theorem}
\begin{proof}
Work in the first three coordinates. Put $v_i=\frac1{\sqrt2}(\cos\frac{2\pi i}m,\sin\frac{2\pi i}m,1,0,\dots,0)$ for $i=1,\dots,m$, let $h_0(x)=\sign(x_3)$, fix $\cos(2\pi/m)<a<1$, and define $h_i(x)=\sign(-\cos\frac{2\pi i}mx_1-\sin\frac{2\pi i}mx_2+ax_3)$. Then $h_i(v_j)$ has the sign of $a-\cos\frac{2\pi(i-j)}m$, negative for $j=i$ and positive otherwise. Since $h(-x)=-h(x)$ for every homogeneous halfspace $h$ and every $x$ off its boundary, at $-v_j$ the signs reverse, as for $h_0$. So $h_i$ disagrees with $h_0$ at $\pm v_i$ and agrees at $\pm v_j$, $j\ne i$. All inequalities are strict, so there are disjoint antipodal neighborhoods $C_i$ of $\{v_i,-v_i\}$ such that, on $\bigcup_jC_j$, $h_i\ne h_0$ iff $u\in C_i$.

Given $\xi>0$, choose elongated ellipsoids along $v_1,\dots,v_m$ whose angular laws put mass at least $1-\xi/(2m)$ on the corresponding $C_i$, form their equal-weight mixture, and apply \cref{lem:radial-mixture} to within total variation $\xi/(2m)$. The resulting angular law satisfies
\begin{equation}\label{eq:sym-caps}
\mu\Bigl(\bigl(\textstyle\bigcup_iC_i\bigr)^c\Bigr)\le\xi,\qquad\frac{1-\xi}m\le\mu(C_i)\le\frac{1+\xi}m.
\end{equation}
Take $\xi\le1/(4m)$. Then $D(h_i\ne h_0)\le\mu(C_i)+\xi\le2/m$, so every $h_i$ lies in $B_D(h_0,2/m)$, whose disagreement region contains $\bigcup_iC_i$ and so has mass at least $1-\xi\ge3/4$ (\cref{lem:star}(a), where $\xi\le1/(4m)$ keeps the exceptional mass at the scale of one cap). Hence $\theta_D(2/m;h_0)\ge(3/4)/(2/m)=3m/8$. Symmetry gives mean zero and the covariance is finite and positive definite, so whitening yields an isotropic law. An invertible linear map preserves $(-1/n)$-concavity and all homogeneous-halfspace disagreement probabilities. Finally, a $(-1/n)$-concave density is $s$-concave for every $s\le-1/n$, because $f^s$ is a positive multiple of $\phi^{-ns}$, convex since $-ns\ge1$.
\end{proof}
This is a class-uniform obstruction. The distribution changes with $m$, and we do not assert that one fixed distribution has coefficient $\Omega(1/\eps)$ at every small scale.

\begin{theorem}[Fixed-budget lower bound at the symmetric endpoint]\label{thm:sym-pool}
Fix $n\ge3$, $s\le-1/n$, an unlabeled budget $U$, and $0<\eps\le1/64$. There is an isotropic, centrally symmetric, everywhere-positive $s$-concave distribution $D$ such that any learner that receives at most $U$ i.i.d.\ unlabeled examples from $D$, queries labels only among them, and achieves error at most $\eps$ with probability at least $2/3$ for every homogeneous target, must use $q\ge\frac12\lfloor\frac1{8\eps}\rfloor=\Omega(1/\eps)$ labels in the worst case. This holds even if the learner knows $D$ and outputs an improper classifier.
\end{theorem}
\begin{proof}
Use the construction of \cref{thm:sym-dis} with $m=\lfloor1/(8\eps)\rfloor\ge8$ and\\ $\xi\le\min\{1/(4m),\,1/(100\max\{U,1\})\}$, so \eqref{eq:sym-caps} holds. Apply \cref{lem:star}(b) with targets $h_1,\dots,h_m$, base labeling $h_0$, the (direction cones of the) caps $C_i$, and $R$ the complement of $\bigcup_iC_i$, of mass at most $\xi$. On $\bigcup_jC_j$, $h_i\ne h_0$ exactly on $C_i$, so $\{h_i\ne h_0\}\subseteq C_i\cup R$. For $i\ne j$, $h_i$ and $h_j$ disagree on $C_i\cup C_j$, so $D(h_i\ne h_j)\ge2(1-\xi)/m\ge3/(2m)\ge12\eps>2\eps$. With $\delta=1/3$ and $U\xi\le1/100$, the lemma gives $q\ge(2/3-1/100)m-1\ge m/2$. Whitening preserves the argument.
\end{proof}
This is consistent with \cref{thm:main}. The complement of $\bigcup_iC_i$ has positive mass and contains informative queries, but it is rare enough that a pool of $U$ examples misses it with high probability.

\begin{corollary}[Sharp symmetric threshold and a stronger general lower bound]\label{cor:sym-sharp}
For $n\ge3$, on $\Dsym$, class-uniform logarithmic disagreement and polylogarithmic labels with an unlabeled budget fixed in advance hold for $s>-1/n$ and fail for every $s\le-1/n$, where $\Omega(1/\eps)$ lower bounds hold for both. Since $\Dsym\subseteq\Dns$, the general class also obeys $\Omega(1/\eps)$ lower bounds for $s\le-1/n$, improving on $\Omega_n(\eps^{-(n-2)/n})$ in that range.
\end{corollary}

\subsection{Rare-tail mean shifting in the general class}\label{sec:rare-tail}
Fix $n\ge3$, $d=n-1$ and $s_0=-1/(n+1)=-1/(d+2)$. For $\tau\ge2$ and $R>0$ define $K_R=\{(t,z)\in\R\times\R^d:0\le t\le R,\|z\|\le1+t\}$ and $f_{\tau,R}(t,z)=Z_{\tau,R}^{-1}(1+\tau t)^{-(d+2)}\mathbf 1_{K_R}(t,z)$.

\begin{lemma}[Basic properties of the hard family]\label{lem:hard-family}
For every $\tau$: (i) $f_{\tau,R}$ is $s_0$-concave, hence $s$-concave for every $s\le s_0$, (ii) there is a finite $R=R(\tau)$ for which the mean equals $x_0=(2,0)$, (iii) if $(T,Z)\sim f_{\tau,R(\tau)}$, then for every $1\le a\le\tau$, $\Pr[T>a/\tau]\le C_n(a^{-n}+\tau^{-n})$.
\end{lemma}
\begin{proof}
(i) On $K_R$, $f^{s_0}_{\tau,R}=Z^{-s_0}_{\tau,R}(1+\tau t)$ is affine. Extended by $+\infty$ it is convex, and for $s\le s_0$, $f^s=(f^{s_0})^{s/s_0}$ is convex. (ii) The unnormalized $t$-density is $w_\tau(t)=(1+t)^d(1+\tau t)^{-(d+2)}$, integrable on $[0,\infty)$, while $\int_1^Rtw_\tau(t)dt\asymp\int_1^Rdt/t\to\infty$. The truncated mean is continuous in $R$ and diverges, so some finite $R$ gives $\E T=2$, and symmetry in $z$ gives $\E Z=0$. (iii) $\int_0^Rw_\tau\ge c_n/\tau$. For $a/\tau\le t\le1$, $w_\tau(t)\le C_n\tau^{-(d+2)}t^{-(d+2)}$ with integral at most $C_n/(\tau a^{d+1})$. For $t\ge1$, $w_\tau(t)\le C_n\tau^{-(d+2)}t^{-2}$. Divide, using $d+1=n$.
\end{proof}

Define the good core $G_{a,\tau}=\{(t,z):t\le a/\tau,\ \|z\|\le1-t/2\}$.
\begin{lemma}[The core is a uniform ball]\label{lem:core}
For $a/\tau\le1$, $\Pr[X\notin G_{a,\tau}]\le C_n(a^{-n}+\tau^{-n}+a/\tau)$. Conditional on $G_{a,\tau}$, $\Pi(T,Z):=Z/(1-T/2)$ is exactly uniform on $B_d$ and independent of $T$.
\end{lemma}
\begin{proof}
Given $T=t\le a/\tau$, $Z$ is uniform on $B_d(1+t)$, so $\Pr[\|Z\|>1-t/2\mid T=t]=1-(\frac{1-t/2}{1+t})^d\le C_dt$. Conditional on $T=t$ and the inner-ball event, $Z$ is uniform on $B_d(1-t/2)$. Rescale.
\end{proof}

For $u\in S^{d-1}$ and $c\in\R$ define $H_{u,c}(t,z)=\sign(c(1-t/2)-\langle u,z\rangle)$, a halfspace whose boundary passes through $x_0=(2,0)$. On the core it is exactly $\sign(c-\langle u,\Pi(t,z)\rangle)$.

\begin{proposition}[Disagreement blow-up in the general class]\label{prop:dis-blowup}
Fix $n\ge3$ and $s\le-1/(n+1)$. There are $c_n,\eps_n>0$ such that for every $0<\eps\le\eps_n$ some isotropic $s$-concave distribution $D$ satisfies $\theta_D(\eps)\ge c_n\eps^{-(n-2)/n}$ at a homogeneous target.
\end{proposition}
\begin{proof}
Choose $a,\tau$ with $\Pr[X\notin G_{a,\tau}]\le\eps/2$. Let $H_0=H_{e_1,2}$, identically positive on the core. For every $u$ choose $b$ with $\nu_d(C(u,b))=\eps/2$, where $C(u,b)$ is the cap of the proof of \cref{lem:caps}. Then $H_{u,b}$ differs from $H_0$ on the core exactly on the preimage of $C(u,b)$, up to a null set, so $d_D(H_0,H_{u,b})\le\eps$. As $u$ varies these caps sweep out the shell $\{b\le\|y\|\le1\}$, of $\nu_d$-mass at least $1-b\ge c_d\eps^{2/(d+1)}$ by the proof of \cref{lem:caps}. Hence $\theta_D(\eps)\ge c_d\eps^{2/(d+1)-1}=c_n\eps^{-(n-2)/n}$. (Equivalently, \cref{lem:star}(a) applied to the disjoint caps of \cref{lem:caps} gives the same order.) Whiten.
\end{proof}

\begin{corollary}[General-class finite-pool lower bound]\label{cor:gen-pool}
Fix $n\ge3$ and $s\le-1/(n+1)$. For every unlabeled budget $U:(0,\eps_n]\to\mathbb N$ and every sufficiently small $\eps$ there is an isotropic $s$-concave $D=D_{\eps,U}$ such that every possibly randomized, computationally unbounded pool learner that inspects at most $U(\eps)$ unlabeled examples and succeeds to error $\eps$ for every homogeneous target with probability at least $2/3$ must use $\Omega_n(\eps^{-(n-2)/n})$ labels. Each hard distribution has an absolutely continuous full-support angular law. For $s\le-1/n$ the bound improves to $\Omega(1/\eps)$ by \cref{thm:sym-pool}.
\end{corollary}
\begin{proof}
Choose $a$ and then $\tau$ in \cref{lem:core} so that $\Pr[X\notin G_{a,\tau}]\le1/(100\max\{U(\eps),1\})$. Conditional on the core, $\Pi(t,z)=z/(1-t/2)$ is uniform on $B_{n-1}$ and every affine halfspace on that ball is realized by a halfspace through $(2,0)$. So the family is affine-core rich and \cref{thm:affine-core} applies. Translate by the mean and whiten. Because the mean is interior and the density positive near it, the centered angular law is absolutely continuous and charges every open cap.
\end{proof}

\subsection{Comparison with prior work and status by regime}\label{app:regimes}
\Cref{tab:before-after} compares each row of \cref{fig:phase} with what was previously established, and \cref{tab:regimes} summarizes the label complexity in each regime.
\begin{table}[htbp]
\centering\small
\renewcommand{\arraystretch}{1.15}
\begin{tabular}{>{\raggedright\arraybackslash}p{2.9cm}>{\raggedright\arraybackslash}p{\dimexpr(\linewidth-2.9cm-6\tabcolsep)/2\relax}>{\raggedright\arraybackslash}p{\dimexpr(\linewidth-2.9cm-6\tabcolsep)/2\relax}}
\toprule
Row of \cref{fig:phase} & Previously established & Added here\\
\midrule
Unlimited pool & \citet{BZ17} learn from samples for $s\ge-1/(2n+3)$. Qualitative links between rich pools and membership queries \citep{GSS13,DKM24} & A label bound uniform over admissible marginals and targets, for every finite $s$. The unlabeled cost remains marginal-dependent\\
Balcan--Zhang tools & \citet{BZ17} for $s\ge-1/(2n+3)$. Several geometric and moment facts are classical & Extended to $s>-1/(n+2)$. Uniform constants fail at and below this threshold\\
Disagreement and fixed-pool labels, general class & Guarantees within the range of \citet{BZ17}. Threshold unknown & Logarithmic disagreement and polylogarithmic labels above $-1/(n+1)$. Sharp exponent $(n-2)/n$ between $-1/n$ and $-1/(n+1)$ (\cref{sec:exponent-proof}), and exponent one at and below $-1/n$\\
Disagreement and fixed-pool labels, symmetric class & Ball--Bobkov bodies and Fradelizi--Gu\'edon--Pajor comparisons available as tools & Logarithmic disagreement and polylogarithmic labels for $s>-1/n$. Exponent one at and below $-1/n$, via the radial construction\\
\bottomrule
\end{tabular}
\caption{Before-and-after comparison for \cref{fig:phase}.}
\label{tab:before-after}
\end{table}
\begin{table}[htbp]
\centering\small
\renewcommand{\arraystretch}{1.15}
\begin{tabular}{>{\raggedright\arraybackslash}p{3.7cm}>{\raggedright\arraybackslash}p{\dimexpr(\linewidth-3.7cm-6\tabcolsep)/2\relax}>{\raggedright\arraybackslash}p{\dimexpr(\linewidth-3.7cm-6\tabcolsep)/2\relax}}
\toprule
Regime & Unlimited-pool labels & Labels with an unlabeled budget fixed in advance\\
\midrule
$n=2$, every finite $s$ & Polylogarithmic & Polylogarithmic, CAL (\cref{prop:plane})\\
$n\ge3$, $s>-1/(n+1)$ & Polylogarithmic & Polylogarithmic, CAL on $\tilde O(n/\eps)$ points (\cref{cor:cal})\\
$n\ge3$, symmetric, $s>-1/n$ & Polylogarithmic & Polylogarithmic, CAL (\cref{thm:sym-disagree})\\
$n\ge3$, general, $-1/n<s\le-1/(n+1)$ & Polylogarithmic (\cref{thm:main}) & $\tilde\Theta_{n,s}(\eps^{-(n-2)/n})$, polynomial pool suffices (\cref{cor:exp-cal,cor:exp-match,cor:gen-pool})\\
$n\ge3$, $s\le-1/n$, either class & Polylogarithmic (\cref{thm:main}) & $\Theta(1/\eps)$ for fixed $n$ (\cref{thm:sym-pool}; passive upper bound $O((n+\log\frac1\delta)/\eps)$ \citep{Han16,BHMZ20})\\
Elliptical laws, any tails & Polylogarithmic & Polylogarithmic, $\theta_D=O(\sqrt n)$ (\cref{prop:elliptical})\\
\bottomrule
\end{tabular}
\caption{Status by regime, at constant confidence. Polynomial entries are optimal rates achievable with an unlabeled budget fixed in advance, and the lower bounds hold for every such budget. Unlimited unlabeled access removes these polynomial label barriers.}
\label{tab:regimes}
\end{table}

\section{Learned verifiers, modeling and proofs}\label{app:verifier}
The pool theorem also applies when labels are expert judgments of generated reasoning states. Generation provides cheap exploration, annotation provides expensive validity judgments, and deployment requires repeated verifier decisions along an adaptive trajectory. This application \emph{assumes} a realizable linear representation and exploration laws that satisfy the hypotheses of \cref{thm:main} or of its two-law form, \cref{thm:two-laws}.

\subsection{What the model assumes}\label{sec:verifier-model}
\paragraph{The angular law of generated states.} Let $Z$ be a generated state, such as a problem, a prefix, and a candidate next step, and let $\phi(Z)\in\R^n\setminus\{0\}$ be its feature representation, with correct verdict $h^\star(Z)=\sign\langle w^\star,\phi(Z)\rangle$. Since $h^\star(Z)=\sign\langle w^\star,\phi(Z)/\|\phi(Z)\|\rangle$, only the direction of $\phi(Z)$ affects the verdict. The relevant angular law is
\[
\mu=\mathcal L\bigl(\phi(Z)/\|\phi(Z)\|\bigr),\qquad Z\sim\nu,
\]
where $\nu$ is the \emph{exploration law}, the law of generated states. It describes which feature directions the generator produces, and how often. It is \emph{not} the distribution of step correctness. In this language the assumptions have the following meanings.
\begin{itemize}[leftmargin=*]
\item \emph{Full support} means that every open cap of feature directions has positive probability, so exploration eventually produces a state near any requested direction. \emph{Symmetric} full support, which suffices, asks this only of each cap or its reflection.
\item \emph{Hyperplane-nullness} means $\Pr\{\langle v,\phi(Z)\rangle=0\}=0$ for every $v\ne0$.
\item The \emph{cap mass} $\mu\{u:\angle(u,q)<\gamma\}$ determines how long the learner may wait for a state approximating query direction $q$.
\end{itemize}
These are properties of the representation and the exploration law, and they can change when verifier feedback changes which states are generated.

\paragraph{Scope of the model.} First, an affine verifier $\sign(\langle w,\phi(Z)\rangle+b)$ can be made homogeneous by appending a constant coordinate. The augmented directions lie in one hemisphere, but only the symmetrized angular law matters, and symmetric full support holds exactly when the feature law has full support in $\R^n$ (\cref{cor:affine}). It fails for bounded feature maps, whose lifted directions avoid a neighborhood of the equator (\cref{sec:necessity}). Second, for $n\ge2$ a deterministic embedding of finite token strings induces an atomic law on a countable set, and every atom lies on some hyperplane through the origin, so it cannot be hyperplane-null. This matters only for the law under which error is measured. \Cref{thm:two-laws} allows an atomic proposal law as long as the evaluation directions are hyperplane-null (\cref{sec:two-laws}). Continuous randomization of the representation could remove atoms, but would require a separate argument that it preserves labels and provides the required support. The results of this appendix are therefore a \emph{geometric model} of when selective annotation can replace synthesized queries. They make no claim about arbitrary tokenized prompts or language-model embeddings.

\paragraph{Pools versus membership queries.} An arbitrary synthesized input may be malformed, ambiguous, outside the expert's competence, or too expensive to assess. An invalid proposed step can usually still be labeled ``incorrect,'' and an unsolvable problem can be labeled if the task allows that answer. The actual obstacle is that a reliable, affordable labeling oracle may exist only on an \emph{admissible} subset of inputs. A natural formalization is a partial concept
\[
c^\star:\mathcal Z\to\{-1,+1,\bot\},\qquad A^\star=\{z:c^\star(z)\ne\bot\},
\]
where $\bot$ means the annotation task is undefined or unavailable. If the exploration law satisfies $\nu(A^\star)=1$, pool queries are always admissible, while an unrestricted membership-query learner may query outside $A^\star$. Allowing it to receive $\bot$ gives a different oracle model, connected to partial concept classes \citep{AHHM21}, which our halfspace results do not analyze. Synthesized vectors in embedding space may also correspond to no reasoning state at all, and a candidate following an invalid prefix needs a precisely specified judgment (local inference validity, valid-prefix status, or extendability to a correct solution). Membership queries remain meaningful as a stronger comparison model that allows expert judgments on specified counterfactual continuations. This does not grant access to labels of arbitrary feature vectors.

\paragraph{Three access models.} \Cref{tab:access} distinguishes selecting labels from controlling which states are available.

\begin{table}[h]
\centering\small
\renewcommand{\arraystretch}{1.15}
\begin{tabular}{>{\raggedright\arraybackslash}p{2.8cm}>{\raggedright\arraybackslash}p{\dimexpr(\linewidth-2.8cm-6\tabcolsep)/2\relax}>{\raggedright\arraybackslash}p{\dimexpr(\linewidth-2.8cm-6\tabcolsep)/2\relax}}
\toprule
Model & What it provides & Limitation under feedback-induced shift\\
\midrule
Fixed offline pool & Selective annotation of previously generated states & Cannot acquire states absent from exploration\\
Interactive pool & Fresh candidates generated at learner-induced prefixes, then selectively annotated & Requires exploration and supervision on the evolving state distribution\\
Membership queries & Labels of constructed admissible prefix--candidate pairs & Useful only if such states can be constructed and reliably labeled\\
\bottomrule
\end{tabular}
\caption{Access models for verifier supervision.}
\label{tab:access}
\end{table}

Interactive pool-based learning is a natural model for chain-of-thought verification. At a current prefix, generate several candidate continuations, inspect them cheaply, and request expert judgments on selected candidates. The verifier's decisions determine future prefixes, so the pool changes during training. We establish two kinds of guarantees. \emph{Offline replacement} (\cref{sec:offline,sec:certificate}). Coverage of the perfect-verifier path, or a known deployment set with a supplied positive margin, suffices for high-probability agreement of the whole deployment trajectory, because before the first different verdict both executions share their history. This avoids coverage after an error but does not guarantee recovery after divergence. \emph{Interaction versus pools} (\cref{sec:three-way}). Adaptive state generation can replace a large offline pool. Applying \cref{thm:main} to successive interactive pools would require additional conditional-support, sampling, confidence and deployment analysis that we do not provide. This leaves open how much expert supervision \emph{and} state exploration are needed when the verifier both selects labels and influences future examples.

\subsection{Exploration and annotation}
We identify states with their representations. The exploration law $\nu$ produces i.i.d.\ unlabeled states that can be inspected before annotation. Its angular law is hyperplane-null with symmetric full support about the verifier's origin. \Cref{sec:two-laws} separates these two roles. A noiseless expert returns $h^\star(z)$. Passive annotation labels naturally occurring states. Offline active annotation adaptively selects labels from an exploration pool. Interactive annotation additionally changes which states become available. Each labeled state counts as one annotation. If an expert call reveals a trajectory, its information and cost must be accounted for separately.

\subsection{Offline annotation under a coverage profile}\label{sec:offline}
Fix an execution making at most $K$ verifier decisions, with deployment randomness independent of training. Let $P^\star_t$ be the subprobability law of the state at decision $t$ under the perfect verifier. Its mass is the probability that decision $t$ is reached. Couple executions using the same generator randomness until their first different verdict. For tasks selected after training, all coverage bounds must hold uniformly over the selection. Suppose $P^\star_t\ll\nu$ and write
\[
g_t=\frac{dP^\star_t}{d\nu},\qquad G=\sum_{t=1}^Kg_t,\qquad \psi(r)=\sup_{\nu(E)\le r}\int_EG\,d\nu,\quad 0\le r\le1.
\]
This coverage profile measures how much the perfect-verifier state laws can concentrate on a set of small exploration mass. It uses only these laws, and no laws of states reached after a verifier error.

\begin{theorem}[Trajectory reliability under a coverage profile]\label{thm:coverage-profile}
If an offline learner outputs $\hat h$ with $\nu(\hat h\ne h^\star)\le r$ with probability at least $1-\delta$, then
\[
\Pr\{\text{deployment diverges from perfect execution}\}\le\delta+\psi(r).
\]
Furthermore, $\psi(r)\to0$ as $r\downarrow0$. For $p>1$, $\psi(r)\le\|G\|_{L^p(\nu)}r^{1-1/p}$, and if $\nu$ is nonatomic, $\psi(r)=\inf_{\lambda\ge0}\bigl\{\lambda r+\int(G-\lambda)_+\,d\nu\bigr\}$. Under the linear and angular assumptions above, suppose that $r\in(0,1/4]$ and a function $\Psi$ are supplied such that $\psi\le\Psi$ for every possible target and $\Psi(r)\le\xi$. Then $O(n\log^3n\log\frac{n}{r\delta})$ pool labels give trajectory agreement with probability at least $1-\delta-\xi$. The number of generated states is finite almost surely and depends on the marginal.
\end{theorem}
\begin{proof}
Condition on a successfully trained verifier and put $E=\{\hat h\ne h^\star\}$. Before the first different verdict, the coupled executions have identical histories, so first divergence at $t$ implies that the perfect-verifier state at decision $t$ is in $E$. Consequently $\Pr\{\text{divergence}\mid\hat h\}\le\sum_tP^\star_t(E)=\int_EG\,d\nu\le\psi(r)$. Add the training failure probability. Since $G$ is integrable with integral at most $K$, for any $B>0$ and $\nu(E)\le r$, $\int_EG\,d\nu\le Br+\int_{\{G>B\}}G\,d\nu$. Choose $B$ large and then $r$ small for continuity at zero. H\"older's inequality proves the $L^p$ bound. For any $\lambda\ge0$, $G\le\lambda+(G-\lambda)_+$ gives the variational upper bound. For $0<r<1$, choose a quantile threshold with $\nu(G>\lambda)\le r\le\nu(G\ge\lambda)$ (taking $\lambda=0$ if the positive-density set has mass below $r$). Nonatomicity allows selecting part of the level set, and if necessary a zero-density set, to form $E$ of mass $r$ attaining the bound. At $r=1$ take $E$ to be the whole space and $\lambda=0$. At $r=0$ let $\lambda\to\infty$. Finally apply \cref{thm:main} at error $r$ and failure probability $\delta$.
\end{proof}
The profile permits unbounded density ratios. For instance, a supplied uniform bound $\|G\|_p\le B$, with $B\ge1$, permits $r=\min\{1/4,(\xi/B)^{p/(p-1)}\}$ and gives label complexity $O\bigl(n\log^3n(\log\frac n\delta+\frac p{p-1}\log\frac B\xi)\bigr)$. The requirement that $\Psi$ hold for every target matters, since the unknown target may determine $P^\star_t$, and neither $\Psi$ nor its validity can be inferred from unlabeled data or linear realizability alone. A simpler sufficient condition is a bounded density ratio,
\begin{equation}\label{eq:vcov}
P^\star_t\ll\nu,\qquad\|g_t\|_\infty\le C_t,\qquad C_t\ge1.
\end{equation}

\begin{corollary}[Offline annotation for trajectory reliability]\label{cor:offline}
Under \eqref{eq:vcov}, there is an offline pool annotator using $O\bigl(n\log^3n\log\frac{n\sum_tC_t}{\xi\delta}\bigr)$ expert labels whose deployment agrees with perfect execution with probability at least $1-\delta-\xi$. No deployment expert is needed. If $C_t\le C_{\mathrm{cov}}$, replace $\sum_tC_t$ by $KC_{\mathrm{cov}}$.
\end{corollary}
\begin{proof}
Here $\psi(r)\le r\sum_tC_t$. Apply \cref{thm:coverage-profile} with $r=\min\{1/4,\xi/\sum_tC_t\}$.
\end{proof}

\begin{proposition}[Sharpness of risk-only transfer]\label{prop:risk-sharp}
For one-step deployment with law $P$, the worst deployment error among unrestricted measurable verifiers having exploration error at most $r$ is exactly $\sup_{\nu(E)\le r}P(E)$. If $P\not\ll\nu$, positive deployment error is possible even with zero exploration error.
\end{proposition}
\begin{proof}
Every disagreement set of exploration mass at most $r$ gives the upper bound. For the reverse bound, take the target constantly negative and the learned verifier positive exactly on $E$, then take the supremum. If $P\not\ll\nu$, some $E$ has $\nu(E)=0<P(E)$.
\end{proof}
\Cref{prop:risk-sharp} concerns unrestricted measurable error sets, and for a fixed linear class the profile can be conservative. \Cref{thm:coverage-profile} bounds the probability of a first divergence, but not the number of mistakes after it. Feedback-induced shift is a central difficulty in \citet{BBFLS26}. Here coverage is an assumption that controls it, and the annotation procedure does not establish it.

Only directions matter for homogeneous verifiers, so the coverage condition can be imposed on angular laws instead of state laws.

\begin{proposition}[Angular occupancy suffices]\label{prop:occupancy}
Let $P^\star_t$ be the subprobability state laws at perfect-verifier decision times $t\le K$, with nonzero features, let $\mu$ be the angular law of an evaluation law from which the learner can sample, and let $\pi_t$ be the angular pushforward of $P^\star_t$. It is enough to assume $\pi_t\ll\mu$, rather than absolute continuity of the full state laws. Put
\[
G_{\mathrm{ang}}=\sum_{t=1}^K\frac{d\pi_t}{d\mu},\qquad\psi_{\mathrm{ang}}(r)=\sup_{\mu(A)\le r}\int_AG_{\mathrm{ang}}\,d\mu.
\]
If a learned homogeneous verifier has $\mu$-error at most $r$ with probability $1-\delta$, independently of deployment randomness, then its probability of any divergence from perfect execution is at most $\delta+\psi_{\mathrm{ang}}(r)$.
\end{proposition}
\begin{proof}
On the training success event its disagreement set is the inverse image of an angular set $A$ with $\mu(A)\le r$. Couple until the first different verdict. First divergence at time $t$ requires the perfect execution's direction to belong to $A$. A union bound yields $\sum_t\pi_t(A)=\int_AG_{\mathrm{ang}}\,d\mu$. Add the training failure probability.
\end{proof}
For instance, exploration and deployment can be supported on different concentric spheres, so their full feature laws are mutually singular while their angular laws coincide, and \cref{prop:occupancy} applies directly. It refines \cref{thm:coverage-profile} rather than giving a separate learning mechanism. Uniform guarantees for tasks chosen after training still require a uniform angular coverage envelope.

\subsection{Separate evaluation and proposal laws}\label{sec:two-laws}
In this and the following subsections the feature map $\phi:\mathcal Z\to\R^n\setminus\{0\}$ is fixed and known, $n\ge2$, and the expert labels a state by $h^\star(z)=\sign\langle w^\star,\phi(z)\rangle$ with $\|w^\star\|=1$. Only actual sampled states may be submitted to the expert. No closure of $\phi(\mathcal Z)$ under interpolation, negation, or linear combinations is assumed. For a direction $u$ and $0<\gamma<\pi/2$, the double cap of \cref{sec:simulation}, read in state space, is
\[
C_\pm(u,\gamma)=\{z:\min(\angle(\phi(z),u),\angle(\phi(z),-u))<\gamma\}.
\]
A proposal law $\mathcal Q$ on states has symmetric full support (\cref{sec:angular-laws}) exactly when every such double cap has positive $\mathcal Q$-probability. This can hold for a countably atomic law, and it does not imply hyperplane-nullness.

\begin{corollary}[Separate evaluation and proposal laws]\label{cor:two-laws-verifier}
Suppose independent unlabeled sampling is available from an evaluation law $\nu$ and a proposal law $\mathcal Q$ on states. Assume the feature directions of $\nu$ are hyperplane-null, $\mathcal Q$ has symmetric full support, and the expert labels only sampled $\mathcal Q$-states. Then there is a learner using
\[
O\Bigl(n\log^3n\log\frac n{\eps\delta}\Bigr)
\]
expert labels that outputs a homogeneous verifier $\hat h$ with $\nu(\hat h\ne h^\star)\le\eps$ with probability at least $1-\delta$. Both unlabeled sample counts are finite almost surely. The proposal count need not have a distribution-uniform bound. No absolute-continuity relation between $\nu$ and $\mathcal Q$ is required.
\end{corollary}
\begin{proof}
Apply \cref{thm:two-laws} to the feature laws of $\nu$ (evaluation) and $\mathcal Q$ (queries). Each query is answered by an actual proposal state whose feature lies in the requested double cap, with its label negated when the feature lies in the reflected cap.
\end{proof}

\begin{remark}[Atomic proposal laws]
For example, $\nu$ can have uniform feature directions while $\mathcal Q$ assigns positive masses to a countable dense set of directions. The two laws are mutually singular, but \cref{cor:two-laws-verifier} applies. This removes an obstacle for annotation access only. The evaluation law must still be hyperplane-null, so atomic evaluation laws remain excluded. The corollary requires unlabeled access to $\nu$ independently of expert decisions. It does not assume that perfect-verifier states can be generated for free. If only a surrogate evaluation law is accessible, a deployment-transfer assumption is still needed.
\end{remark}

\paragraph{Verifier classes.} The results cover realizable linear probes on frozen features and linear combinations of fixed verifier scores, but not arbitrary neural verifiers or jointly learned representations. Affine rules are covered by lifting, as discussed in \cref{sec:verifier-model}. The learning reductions also apply outside verification, and the execution coupling and the proposal interface connect them to verdict-dependent generation.

\subsection{Exact verification on a known finite deployment set}\label{sec:certificate}
\begin{theorem}[Finite-state deployment certificate]\label{thm:certificate}
Let $\mathcal V\subset\R^n\setminus\{0\}$ be a known finite set of $M\ge1$ feature vectors, and suppose a supplied $\theta_{\mathcal V}>0$ satisfies
\begin{equation}\label{eq:deploy-margin}
\min_{v\in\mathcal V}\frac{|\langle w^\star,v\rangle|}{\|v\|}\ge\theta_{\mathcal V}.
\end{equation}
Suppose unlabeled states can be sampled from a proposal law $\mathcal Q$ with symmetric full support, with expert labels on requested sampled states. There is a learner using at most
\[
q_{\max}=O(n\log^3n\log(2M/\delta))
\]
expert labels which, with probability at least $1-\delta$, outputs a homogeneous verifier agreeing with $h^\star$ on every vector in $\mathcal V$.

If $\mathcal V$ contains the features of every state at which a specified deployment system can request a verdict, uniformly over tasks, random seeds, and verdict histories, then on this same event the learned verifier and the perfect verifier have identical executions for every such task and random seed. This holds without an occupancy-density bound, even if $\mathcal Q$ gives the deployment set probability zero.
\end{theorem}
\begin{proof}
Run \cref{prop:robust-pl} directly on $\mathcal V$. Every query instance contains a vector of $\mathcal V$, whose normalized target margin is at least $\theta_{\mathcal V}$ by \eqref{eq:deploy-margin}. Answer each query as in the proof of \cref{thm:two-laws}, from fresh proposal states in the double cap of radius $\min\{\pi/4,\eta^*\theta_{\mathcal V}/\kappa(T)\}$ around the requested direction, using one expert label per query. \Cref{lem:pool-oracle} and the remark after it apply with $\theta_0$ replaced by $\theta_{\mathcal V}$, since their proofs use only a point of $V_0$ with that normalized margin. The waiting times are finite almost surely. With probability $1-\delta$ all recovered labels are exact. On that event strict separation of the finite set is feasible. Return any consistent homogeneous separator. If separation is infeasible, use an arbitrary fallback.

Fix any allowed deployment task and seed, and couple the two executions with the same randomness. Initially their histories agree. If their histories agree up to a verdict, their next queried state agrees and has features in $\mathcal V$, where their verdicts agree. Induction proves equality of every finite execution prefix. This also covers infinite runs and repeated states. Because the training event gives simultaneous correctness on all of $\mathcal V$, the conclusion holds simultaneously over tasks and seeds, including choices made after training.
\end{proof}

\begin{remark}[Assumptions and costs]
The set must be an available, known superset of reachable decision states. Enumerating it may be expensive, and the theorem does not charge enumeration as expert labels. The margin bound is supplied rather than inferred from the labels. A lookup table on $\mathcal V$ would suffice for execution equivalence, but linear consistency also produces a verifier on all states. The theorem does not assert correctness outside the certified set.
\end{remark}

\begin{corollary}[Infinite deployment sets via directional covering]\label{cor:net}
Let $\emptyset\ne\mathcal U\subset S^{n-1}$ contain every possible deployment direction and suppose $|\langle w^\star,u\rangle|\ge\theta_{\mathcal U}>0$ for every $u\in\mathcal U$. Suppose a finite Euclidean $\theta_{\mathcal U}/4$-net $F\subset\mathcal U$ is available. Every $u\in\mathcal U$ is within distance $\theta_{\mathcal U}/4$ of some $v\in F$. Under the proposal access of \cref{thm:certificate}, one can certify all of $\mathcal U$ with probability $1-\delta$ using
\[
O\bigl(n\log^3n\,(\log(2|F|)+\log(1/\delta))\bigr)
\]
expert labels. In particular, choosing a net of minimum size gives a bound in terms of directional covering entropy. A net of size at most $(1+8/\theta_{\mathcal U})^n$ always exists geometrically. Making it available is a separate access requirement.
\end{corollary}
\begin{proof}
Use \cref{thm:certificate} to recover the labels $y_v$ of all $v\in F$. On the success event the convex feasibility problem
\[
\|a\|\le1,\qquad y_v\langle a,v\rangle\ge\theta_{\mathcal U}/2\quad(v\in F)
\]
has the true unit normal $w^\star$ as a feasible solution. Return the homogeneous verifier with any feasible normal $a$. For $u\in\mathcal U$, choose $v\in F$ with $\|u-v\|\le\theta_{\mathcal U}/4$. Since $|\langle w^\star,v\rangle|\ge\theta_{\mathcal U}$, the target labels at $u$ and $v$ coincide. Also
\[
y_v\langle a,u\rangle\ge y_v\langle a,v\rangle-\|a\|\|u-v\|\ge\theta_{\mathcal U}/2-\theta_{\mathcal U}/4>0.
\]
Thus the returned verifier is correct on every direction of $\mathcal U$ simultaneously. This conclusion requires the margin constraints in the fitting step. Arbitrary label consistency on the net is insufficient. If the constraints are infeasible, return a fallback, which only matters on the training failure event.

For the size bound, greedily choose points of $\mathcal U$ more than $\theta_{\mathcal U}/4$ apart until no such point remains. The disjoint balls of radius $\theta_{\mathcal U}/8$ around them lie in the ball of radius $1+\theta_{\mathcal U}/8$, so volume comparison bounds their number by $(1+8/\theta_{\mathcal U})^n$. The procedure therefore terminates and yields a net. This argument shows existence only, and computing the net requires access to $\mathcal U$.
\end{proof}

\begin{proposition}[Adaptive proposal laws and an explicit generation bound]\label{prop:adaptive-proposal}
\Cref{thm:certificate} still holds if, before query $j$, the learner chooses any proposal law $\mathcal Q_j$ based on its observed history, then holds that law fixed while drawing fresh independent states until its required double cap $C_j$ is hit. It suffices that
\[
p_j=\mathcal Q_j(C_j)>0
\]
almost surely. For any $0<\zeta<1$, with probability at least $1-\zeta$, the total number of proposal states inspected is at most
\[
\sum_{j=1}^J\Bigl\lceil\frac{\ln(q_{\max}/\zeta)}{p_j}\Bigr\rceil,\qquad J\le q_{\max}.
\]
Thus exact deployment and this generation bound hold jointly with probability at least $1-\delta-\zeta$.
\end{proposition}
\begin{proof}
At the start of query $j$, both its cap and $p_j$ are measurable with respect to the past. Conditional on that past, its waiting time $W_j$ is geometric with success probability $p_j$. For $t_j=\lceil\ln(q_{\max}/\zeta)/p_j\rceil$,
\[
\Pr\{W_j>t_j\mid\text{past}\}=(1-p_j)^{t_j}\le\zeta/q_{\max}.
\]
Taking expectations and a union bound over at most $q_{\max}$ possible queries proves the claim. No conditioning on future cap probabilities is used.
\end{proof}
Thus a generator can choose a proposal law that makes the next required cap accessible. This changes the $p_j$, and hence the number of generated states, while the number of expert labels remains controlled by point location. Symmetric full support is a convenient sufficient condition, but positivity for the caps actually requested is enough.

\subsection{An unavoidable generation cost, even for one deployment state}\label{sec:generation-cost}
\begin{theorem}[A sharp two-target margin example]\label{thm:two-target}
Let $\mathcal Q$ be uniform on $S^1$, let the only deployment feature be $v=e_1$, and fix $0<\theta\le1/2$. Consider the two possible unit target normals
\[
w_+=(\theta,\sqrt{1-\theta^2}),\qquad w_-=(-\theta,\sqrt{1-\theta^2}).
\]
Both have margin $\theta$ at $v$, with opposite labels there. A learner inspecting at most $U$ independent $\mathcal Q$-samples and requesting any subset of their labels has worst-target failure probability at least
\[
\frac12(1-p_\theta)^U,\qquad p_\theta=\frac{2\arcsin\theta}\pi.
\]
Consequently, failure at most $\delta<1/2$ on both targets requires
\[
U\ge\frac{\ln(1/(2\delta))}{-\ln(1-p_\theta)}.
\]
For $\delta\le1/4$ this is $\Omega(\theta^{-1}\log(1/\delta))$. If the two candidate normals are known, waiting for their disagreement region and querying one label achieves the same order of generation cost and uses only one expert label.
\end{theorem}
\begin{proof}
The normals make angle $2\arcsin\theta$, so their disagreement region has $\mathcal Q$-mass $p_\theta$. Couple both runs using the same sample stream and internal random seed, extending the stream to $U$ samples if the learner stops early. If none of these samples enters the disagreement region, every observed label is identical under both targets. The output is then the same, but the required deployment labels are opposite. Under an equal prior on the two targets, conditional failure is at least $1/2$ on this event. This proves the lower bound. Since $p_\theta=\Theta(\theta)$ and $p_\theta\le1/3$, the asymptotic statement follows. For the upper bound, membership in the two-target disagreement region is known from unlabeled features, and its first sampled point has a label identifying the target. Truncating after $\lceil\ln(1/\delta)/[-\ln(1-p_\theta)]\rceil$ samples fails with probability at most $\delta$. Without truncation the wait is almost surely finite with expectation $1/p_\theta$.
\end{proof}

\begin{proposition}[Zero-margin deployment points]\label{prop:zero-margin}
With proposal law uniform on $S^1$ and deployment feature $e_1$, no learner that terminates after finitely many inspected states almost surely for every target can predict the deployment label with probability at least $3/4$ uniformly over all homogeneous target normals, including zero-margin targets.
\end{proposition}
\begin{proof}
Take $w_0=e_2$, for which $h_{w_0}(e_1)=+1$ by the convention $\sign(0)=+1$, and $w_t=(-t,1)/\sqrt{1+t^2}$, $t>0$, for which $h_{w_t}(e_1)=-1$. Couple runs using one infinite sample stream and one random seed. Under $w_0$ the algorithm inspects a finite prefix almost surely, and no point of that prefix has zero second coordinate almost surely. For each realized finite prefix, all its target labels under $w_t$ agree with those under $w_0$ once $t$ is sufficiently small. By induction over the algorithm's operations, its complete terminating transcript and output are then identical. Dominated convergence implies
\[
\Pr_{w_t}\{\text{output }+1\}\longrightarrow\Pr_{w_0}\{\text{output }+1\}\ge3/4.
\]
But uniform success for $w_t$ would require the left side to be at most $1/4$, a contradiction. This argument counts all inspected states, not just labeled ones.
\end{proof}

\subsection{Separating demonstrations, active annotation, and interaction}\label{sec:three-way}
We distinguish three training interfaces. In \emph{successful-trace behavioral cloning} (BC), the learner receives independent expert demonstrations but cannot choose their states or ask for counterfactual verdicts. Our examples use one-step successful traces, so each demonstration consists of one accepted state. The learner may process these demonstrations arbitrarily. The lower bounds do not restrict the fitting algorithm. In \emph{offline active annotation}, the learner inspects an i.i.d.\ unlabeled pool from a specified target-independent law and adaptively requests verdicts only on its elements. In \emph{interactive verification}, a known generator can change its next proposal using previous expert verdicts. It has no access to the unknown target except through those verdicts. Every generated state and expert verdict is counted separately. Deployment receives no expert feedback.

The common objective below is to learn the expert's verdict function, or reproduce its verdict trajectory. This is stronger than merely producing some valid answer. In particular, a BC lower bound for this objective is not a lower bound on every possible task-success objective for an imitation policy.

\begin{theorem}[A three-way separation in two dimensions]\label{thm:interaction}
For every integer $m\ge1$, let $M=2^m$ and $\eps_M=1/(8M)$. There are a known family $\mathcal H_M$ of $M$ homogeneous linear verifiers in $\R^2$ and a fixed exploration law $\nu$, uniform on $S^1$, with the following properties for $0<\delta\le1/4$.
\begin{enumerate}[label=(\roman*)]
\item Successful demonstrations under $h$ are independent draws from $\nu(\cdot\mid h=+1)$. Any learner achieving $\nu$-error at most $\eps_M$ with probability $1-\delta$ for every target needs $\Omega(M\log(1/\delta))$ demonstrations. Conversely, $\lceil2M\log(2/\delta)\rceil$ demonstrations suffice for exact identification.
\item Any offline learner inspecting at most $U$ i.i.d.\ $\nu$-states has worst-target failure probability at least $\frac12(1-1/(2M))^U$, even if every inspected state is labeled and the output is improper or randomized. Failure at most $\delta$ therefore requires $U\ge M\log(1/(2\delta))$. A pool of $\lceil8M\log((M-1)/\delta)\rceil$ states suffices for exact identification with only $m$ expert verdicts.
\item An interactive generator identifies the target exactly using $m$ generated states and $m$ expert verdicts. Its next-state distribution depends only on earlier verdicts and its own randomness. The same generator defines an $m$-decision deployment for which distinct targets have distinct verdict sequences.
\item At confidence greater than $1/2$, at least $m$ binary expert verdicts are necessary for exact identification in either of the latter two interfaces when generation has no additional target-dependent side information.
\end{enumerate}
The lower bounds allow randomized learning and improper output. The offline lower bound also holds for exact agreement with the generator's deployment trajectory under common generator randomness. Each perfect-verifier decision-state law in that deployment has density ratio at most $8M$ relative to $\nu$.
\end{theorem}
\begin{proof}
Put $\alpha_j=(j+1/2)\pi/(2M)$ and
\[
w_j=(\cos\alpha_j,\sin\alpha_j),\qquad h_j(x)=\sign\langle w_j,x\rangle,\qquad0\le j<M.
\]
Adjacent targets have $\nu$-disagreement $1/(2M)>2\eps_M$. Therefore no output classifier can be $\eps_M$-accurate for both adjacent targets.

\emph{Demonstration lower bound.} The accepted semicircles of $h_0,h_1$ each have length $\pi$, and their intersection has length $\pi-\pi/(2M)$. Each successful-trace law is uniform on its semicircle. The two laws have a common component of mass $1-1/(2M)$, with the same conditional distribution on the intersection. For $N$ demonstrations their product laws consequently have a common component of mass $(1-1/(2M))^N$. On this component the same observations and random seed produce the same output. Under an equal prior on the two targets it fails with conditional probability at least $1/2$. Hence worst-target failure is at least $\frac12(1-1/(2M))^N$. Rearranging and using $-\log(1-x)\le2x$ for $x\le1/2$ gives $N\ge M\log(1/(2\delta))$.

\emph{Demonstration upper bound.} Keep all candidate targets positive on every demonstration. For a true interior index $j$, a positive example that is negative under $h_{j+1}$ rules out every larger index, because the portions of $h_j$'s positive semicircle removed by increasing the normal angle are nested. The analogous fact holds for $h_{j-1}$ and all smaller indices. Each nearest-neighbor witness has conditional probability $1/(2M)$ under $\nu(\cdot\mid h_j=+1)$, and at an endpoint only one neighbor is needed. Thus failure to leave a unique candidate has probability at most $2e^{-N/(2M)}$. This proves the upper bound. The demonstration interface can be realized by rejection sampling from $\nu$ with the perfect verifier, recording only the accepted one-step trace. Rejected proposals and rejection counts are not included in the observed demonstration. Acceptance probability is $1/2$ for every target, so trace length and rejection counts do not reveal a target-dependent rate.

\emph{Offline lower bound.} Couple runs under $h_0,h_1$ using the same first $U$ proposal states and internal seed, including any states not inspected after early stopping. With probability $(1-1/(2M))^U$ the entire prefix misses their disagreement region. Every possible annotation transcript then agrees under the two targets, so the same two-target argument applies, for any adaptive allocation of labels in that pool. The bound on $U$ follows from $-\log(1-x)\le2x$.

\emph{Interactive upper bound.} For a cut $k\in\{1,\dots,M-1\}$, let the generator sample a direction uniformly from the arc
\[
I_k=[k\pi/(2M)+\pi/2-\pi/(8M),\,k\pi/(2M)+\pi/2+\pi/(8M)].
\]
If the sampled angle is $k\pi/(2M)+\pi/2+\vartheta$, then $\langle w_j,x\rangle=\sin(\alpha_j-k\pi/(2M)-\vartheta)$. Since the distance from a target angle to the cut is at least $\pi/(4M)$ and $|\vartheta|\le\pi/(8M)$, its label is positive exactly when $j\ge k$. The generator maintains an interval of candidate indices, proposes its midpoint cut, and retains the half indicated by the verdict. This is a binary search of depth $m$, using precisely one generated state and one label per decision.

\emph{Offline upper bound and deployment.} Every cut arc has $\nu$-mass $1/(8M)$. A union bound shows that all $M-1$ cut arcs are represented except with probability $(M-1)e^{-U/(8M)}$. On this event the offline learner performs the same binary search by annotating a stored point in each requested arc. For deployment, run the same generator without expert help. Its $m$ verdicts decode one index, and a single output policy cannot agree with the perfect trajectories of two different indices under the same random seed. This also proves the offline trajectory lower bound. Under the perfect verifier each cut is target-determined, and its state law is uniform on $I_k$, with density $8M\mathbf 1_{I_k}$ relative to $\nu$.

\emph{Label lower bound.} Fix the internal randomness and all target-independent proposal randomness. An algorithm with at most $q$ binary verdicts has at most $2^q$ transcripts and can identify at most $2^q$ of the $M$ targets correctly. Under a uniform prior its success probability is at most $2^q/M$, also after averaging over randomness. If $q\le m-1$ this is at most $1/2$. The proposal generation itself supplies no additional information about the target, because each proposal depends only on the fixed randomness and previous verdicts.
\end{proof}

\begin{table}[h]
\centering\small
\begin{tabular}{lll}
\toprule
Training interface & Expert supervision & State access for the upper bound\\
\midrule
Successful-trace BC & $\Theta(M\log(1/\delta))$ demonstrations & Same number of accepted states\\
Offline active annotation & $\log_2M$ verdicts & $O(M\log(M/\delta))$ i.i.d.\ proposals\\
Interactive verification & $\log_2M$ verdicts & $\log_2M$ controlled proposals\\
\bottomrule
\end{tabular}
\caption{Resources of the three training interfaces in \cref{thm:interaction}, with $M=2^m$ targets.}
\label{tab:interfaces}
\end{table}

\paragraph{Interpretation.} At fixed confidence, \cref{tab:interfaces} separates two resources exponentially in $m=\log_2M$. Active annotation saves supervision relative to successful traces, and interaction saves state acquisition relative to offline annotation. The offline pool bounds differ by a logarithm, and because adjacent targets differ in population by only $1/(2M)$, the offline lower bound concerns deployment-sensitive coverage. The difference lies in the interface. A fixed pool contains only states generated in advance, whereas DAgger \citep{RGB11} and the offline datasets of \citet[Remark~2.15]{BBFLS26} can supply labels on learner-induced or adaptively generated prefixes, a stronger form of access.

\begin{proposition}[Successful traces need not identify counterfactual verdicts]\label{prop:bc-blind}
There is a two-element class of homogeneous verifiers in $\R^2$ and an expert demonstration policy producing identical successful traces under both targets, such that any learner using only those traces has worst-target error at least $1/2$ on one specified deployment verdict. Active annotation from a uniform-circle pool identifies the target with one expert verdict and finite expected generation cost. Controlled interaction does so with one state and one verdict.
\end{proposition}
\begin{proof}
Take $w_+=(1,1)/\sqrt2$ and $w_-=(-1,1)/\sqrt2$. The demonstration policy always returns the valid one-step trace with feature $e_2$, positive under both targets. Deployment asks for the verdict at $e_1$, whose labels are opposite. Identical training laws force average failure at least $1/2$ under the equal target prior, even with arbitrarily many demonstrations. The two targets disagree on half of the circle, so an active annotator can wait for that known region and identify the target from one label. The expected wait is two samples. An interactive generator can propose $e_1$ directly.
\end{proof}
This example isolates absent coverage in the demonstration policy. It does not imply that copying the demonstrated action fails to produce a valid answer. That action is valid for both targets. The separation concerns learning a verifier for counterfactual proposals. By contrast, \cref{thm:interaction} uses nondegenerate, target-dependent demonstrations covering every accepted direction.

\paragraph{Open questions.} Beyond the interactive question raised in \cref{sec:discussion}, two problems are concrete. First, can a reachable deployment set be certified when it is only implicit, generated by the deployment system rather than enumerated as in \cref{thm:certificate}? Second, can the supplied margin be replaced by adaptation to an unknown positive deployment margin? \Cref{prop:zero-margin} shows that some positive margin is necessary.

\section{Noise}\label{sec:noise}\label{app:noise}
Both the supplied-normalization directional reduction and the full pool simulation extend to Massart label noise through a majority oracle that is correct outside the tolerance band. To avoid confusion with the geometric tolerance $\eta$ of \cref{def:mt-oracle}, we write $\beta$ for the Massart flip rate.

\begin{theorem}[Massart noise for directional search]\label{thm:massart}
Under the supplied-normalization and full-support assumptions of \cref{thm:dir-search}, suppose each fresh label is independently flipped with conditional probability at most $\beta<1/2$. With probability at least $1-\delta$, directional search achieves error $\eps$ using
\[
O\Bigl(\frac{n\log(nL/\eps)\,\log(n\log(nL/\eps)/\delta)}{(1-2\beta)^2}\Bigr)
\]
labels. Write $\mu_A$ for the angular law in the supplied normalized coordinates, $\gamma=\eps/(Cn^2L)$ for the precision, $k=\lceil C(1-2\beta)^{-2}\log(Q/\delta)\rceil$, and $Q=O(n\log(nL/\eps))$. With additional failure probability at most $\delta'$, the unlabeled cost is $O\bigl(Q[k+\log(Q/\delta')]/a_{\mu_A}(\gamma)\bigr)$.
\end{theorem}
\begin{proof}
Use the same target-independent precision $\gamma$ as in \cref{thm:dir-search}. For each unit query $q$, collect $k$ fresh points in its $\gamma$-neighborhood, where $Q$ is a deterministic upper bound on the number of queries. If $|\langle w,q\rangle|>\gamma$, every selected point has the same clean label as $q$, and conditional on the past and selected features, Hoeffding's inequality makes the majority incorrect with probability at most $\delta/Q$. For $|\langle w,q\rangle|\le\gamma$ the majority may be arbitrary, which the imprecise comparator and imprecise binary search permit. A union bound over the adaptive transcript gives the oracle required by the noiseless proof. An angular cap of radius $\gamma$ lies in the Euclidean $\gamma$-neighborhood and each fresh draw hits it with conditional probability at least $a_{\mu_A}(\gamma)$. A binomial lower-tail bound and a conditional union bound give the unlabeled estimate.
\end{proof}

The same majority construction applies to every transformed query of the pool simulation.

\begin{corollary}[Pool simulation under Massart noise]\label{cor:massart-pool}
Under the assumptions of \cref{thm:main}, suppose each label is independently flipped with conditional probability at most a known $\beta<1/2$, so that the clean target is Bayes optimal. With probability at least $1-\delta$, a pool learner achieves excess error at most $\eps$ using
\[
O\Bigl(\frac{n\log^3n\,\log^2\frac{n}{\eps\delta}}{(1-2\beta)^2}\Bigr)
\]
labels, uniformly over all such marginals and targets. The unlabeled cost is that of \cref{thm:unlabeled} multiplied by the number $k$ of votes per query.
\end{corollary}
\begin{proof}
Run the proof of \cref{thm:main}. The margin floor and the passive sample use no labels, and the passive sample is labeled by point location, not by querying it. Let $Q=O(n\log^3n\log(N/\delta))$ be the deterministic query bound of \cref{prop:robust-pl}, and $k=\lceil 2(1-2\beta)^{-2}\log(4Q/\delta)\rceil$. To answer a query $z'$ in an instance $(V_0,T)$, wait for $k$ fresh pool points $p$ with $p$ or $-p$ within angle $\gamma=\eta^*\theta_0/\kappa(T)$ of $T^{-1}z'$, request their labels, negate the labels of the reflected ones, and return the majority. A negated noisy label is the correct label of $-p$ flipped with the same probability. If the query lies outside the tolerance band, then by \cref{lem:pool-oracle} and the remark after it every one of these points, after negation where applicable, has the clean label $\sign\langle z',w'\rangle$, and conditional on the past and on the selected points the observed labels are independent and each is correct with probability at least $1-\beta$. Hoeffding's inequality makes the majority wrong with probability at most $\delta/(4Q)$. Inside the band any answer is allowed. A union bound over the at most $Q$ queries shows that the majorities form an $\eta^*$-subspace-relative oracle for the clean target with probability at least $1-\delta/4$. Point location then labels the passive sample with clean labels, and a consistent halfspace has error at most $\eps$ against the clean target, hence excess error at most $\eps$. The label count is $Qk$, and $\log(4Q/\delta)=O(\log(n/(\eps\delta)))$.
\end{proof}

Population Massart noise constrains the label bias at each point, whereas adversarial noise can concentrate precisely on the rare directions used by pool simulation. In particular, directional search may repeatedly query a region of angular mass far below $\eps$, so it is not robust to adversarial noise. On classes for which the angular law is bounded above and below relative to a convex-body cone measure, radial randomization maps each point to its direction times an independent uniform-ball radius. Labels are unchanged, noise mass is preserved, and the resulting density is within a fixed factor of the uniform ball, so standard localized hinge-loss algorithms apply for sufficiently small adversarial noise. Agnostic and Tsybakov guarantees follow from disagreement-based theory wherever the disagreement coefficient admits a class-uniform bound, for example on the symmetric class with $s>-1/n$ and on elliptical laws.

An analogue of \cref{thm:main} under adversarial noise remains open (\cref{sec:discussion}). Imposing central symmetry does not by itself provide one.

\ifextendedproofs
\section{Consequences of the sharp general-class exponent}\label{app:exponent}
The projective and level-set proof of \cref{thm:exp-upper} is in \cref{sec:exponent-main}. Here we derive the label bound under an unlabeled budget fixed in advance, combine it with the lower bound, and prove the depth-dependent refinement.
\fi
\subsection{A fixed unlabeled budget and a high-probability label bound}
\begin{corollary}[CAL with an unlabeled budget fixed in advance]\label{cor:exp-cal}
Under the hypotheses of \cref{thm:exp-upper}, fix $0<\eps,\delta<1/2$ and write $K=K_{n,p}$. There are absolute constants $C_0,C_1$ and a constant $C_n$ such that, with
\[
N=\frac{C_0}{\eps}\Bigl(n\log\frac2\eps+\log\frac4\delta\Bigr),\qquad A=C_1\Bigl(n\log(2N)+\log\frac{4N}\delta\Bigr),
\]
CAL on $N$ independent unlabeled points, with a deterministic label cap
\[
Q=\bigl\lceil C_nKA^{2/n}N^{1-2/n}+\log(2/\delta)\bigr\rceil,
\]
learns every homogeneous target to error $\eps$ with probability at least $1-\delta$ (and never uses more than $N$ labels). In particular, for fixed $n,s,\delta$, a pool of $\tilde O(n/\eps)$ points suffices with $\tilde O_{n,s}(\eps^{-(n-2)/n})$ labels.
\end{corollary}
\begin{proof}
Let $\mathcal H_m$ be the version space after the first $m$ points. CAL requests a label exactly on $\DIS(\mathcal H_m)$. Otherwise every surviving hypothesis gives the correct label, since the target survives, so $\mathcal H_m$ is exactly the set of hypotheses consistent with all $m$ true labels. The realizable VC bound with a union bound over $1\le m\le N$ gives an event $E$ of probability at least $1-\delta/2$ on which $\mathcal H_m\subseteq B_D(h^\star,\min\{1,A/m\})$ for every $1\le m\le N$ and $\mathcal H_N\subseteq B_D(h^\star,\eps)$, for suitable $C_0,C_1$. Let $E_m$ be the radius condition at time $m$, let $E_0$ be the sure event, and let $I_{m+1}$ be the uncapped CAL query indicator. Conditional on the history through time $m$, on $E_m$ its mean is at most $b_m=\min\{1,K(\min\{1,A/\max\{1,m\}\})^{2/n}\}$ by \cref{thm:exp-upper}.

To avoid conditioning on a future event, use the stopped indicators $J_{m+1}=I_{m+1}\mathbf 1_{E_0\cap\dots\cap E_m}$. Their conditional means are at most the deterministic numbers $b_m$, so iterating the conditional exponential-moment bound for Bernoulli variables gives $\E\exp(\sum_{m<N}J_{m+1})\le\exp((e-1)\sum_{m<N}b_m)$. For $A\ge1$, comparison with an integral gives $\sum_{m<N}b_m\le C_nKA^{2/n}N^{1-2/n}$ (if $A>N$ this follows from the trivial bound $N$). Markov's inequality applied to the exponential bounds $\sum_mJ_{m+1}$ by $Q$, after enlarging $C_n$, except with probability $\delta/2$. On $E$ the stopped and original indicators coincide. Run CAL with the cap $Q$, and if a further label would be required, stop and output any hypothesis. With probability at least $1-\delta$ no stop occurs and the final consistent output has error at most $\eps$. The pool size and label cap are fixed before observing data.
\end{proof}

\begin{corollary}[Matching intermediate exponent]\label{cor:exp-match}
Fix $n\ge3$ and $-1/n<s\le-1/(n+1)$. For the isotropic general class $\Dns$,
\[
\sup_{D\in\Dns}\sup_{h^\star}\theta_D(\eps;h^\star)=\Theta_{n,s}\bigl(\eps^{-(n-2)/n}\bigr)\qquad\text{as }\eps\downarrow0.
\]
At success probability $2/3$, the optimal label exponent achievable with an unlabeled budget fixed in advance for the class is also $(n-2)/n$. \Cref{cor:exp-cal} achieves $\tilde O_{n,s}(\eps^{-(n-2)/n})$ labels with a polynomial pool, while every unlabeled budget fixed in advance incurs the $\Omega_n(\eps^{-(n-2)/n})$ worst-case lower bound of \cref{cor:gen-pool}.
\end{corollary}
\begin{proof}
Combine \cref{thm:exp-upper} with \cref{prop:dis-blowup}, and \cref{cor:exp-cal} with \cref{cor:gen-pool}. The lower-bound distribution may depend on $\eps$ and on the budget. We make no claim that an arbitrarily small pool suffices.
\end{proof}

\paragraph{The endpoint and dimension two.} The strict inequality $s>-1/n$ is essential, since the level-set integral has denominator $1+ns$. At $s\le-1/n$, \cref{thm:sym-dis} and the trivial bound $\theta_D(\eps)\le1/\eps$ give worst-case disagreement $\Theta(1/\eps)$, and optimal realizable passive learning, with $O((n+\log(1/\delta))/\eps)$ labels \citep{Han16,BHMZ20}, matches the $\Omega(1/\eps)$ label lower bound of \cref{thm:sym-pool} for fixed $n$. For the symmetric class with $s>-1/n$, \cref{thm:sym-disagree} is stronger than the bound here. Dimension two is covered by \cref{prop:plane}, since the proof here uses $2/n<1$. Sharp logarithmic factors and the optimal dependence on $n$ and $1+ns$ remain open.

\subsection{Depth-dependent control below the general threshold}
A uniform positive lower bound on the depth of the reference point restores logarithmic disagreement throughout the intermediate range. Define the (Tukey) depth $\mathrm{depth}_D(x_0)=\inf\{D(H):H\text{ a closed affine halfspace containing }x_0\}$.

\begin{proposition}[A positive-depth reference point]\label{prop:depth}
Let $n\ge3$ and $p>n$, and let $D$ have a density $f$ with $f^{-1/p}$ convex. If $x_0$ is an interior point of the support with $0<f(x_0)<\infty$ and $\mathrm{depth}_D(x_0)\ge\alpha>0$, then halfspaces whose boundaries pass through $x_0$ satisfy
\[
\theta_D(\eps;h^\star)\le C_{n,p,\alpha}\bigl(1+\log(1/\eps)\bigr)\qquad(0<\eps\le1)
\]
for every such target $h^\star$. No moment or mean assumption is needed, and the constant is uniform over this class.
\end{proposition}
\begin{proof}
First suppose $f$ has an attained finite maximum $f_{\max}$, and put $t=f(x_0)$ and $\lambda=1-n/p>0$. The volume bound $V(y)\le(2^p/f_{\max})(f_{\max}/y)^{n/p}$ from the proof of \cref{lem:level-int} and the layer-cake formula give
\[
D(f\le t)\le\int_0^tV(y)\,dy\le\frac{2^p}\lambda\Bigl(\frac t{f_{\max}}\Bigr)^\lambda.
\]
If $t<f_{\max}$, separate $x_0$ from the convex set $\{f>t\}$. The closed halfspace containing $x_0$ has mass at most $D(f\le t)$, so $\alpha\le(2^p/\lambda)(t/f_{\max})^\lambda$ and
\[
R:=\frac{f_{\max}}{f(x_0)}\le\Bigl(\frac{2^p}{\lambda\alpha}\Bigr)^{1/\lambda}.
\]
The case $t=f_{\max}$ is trivial. For a general density, use the capped densities $f_c$ of \cref{prop:wet-density}. Their depth at $x_0$ is at least $\alpha-\|D_c-D\|_{\mathrm{TV}}$, and their mode-to-reference ratio is $c/f(x_0)$ once $c>f(x_0)$. Letting $c$ increase gives the same bound on $R$, and in particular shows that $\operatorname{ess\,sup}f<\infty$.

Translate $x_0$ to the origin and form the radial bodies of $f/f(x_0)$ as in the proof of \cref{thm:normalization}. The same comparison gives $R^{-1/[n(n-1)]}K_{n-1}\subseteq K_n\subseteq B_{n,s}K_{n-1}$ with $s=-1/p$, and $K_{n-1}$ is convex because $p>n$ (\cref{lem:bobkov}). So the angular law of $D$ and the cone measure $\nu$ of $K_{n-1}$ have density ratio in $[L^{-1},L]$ with $L=B_{n,s}^nR^{1/(n-1)}$. Every halfspace through the origin has $D$-mass at least $\alpha$, hence $\nu$-mass at least $\alpha/L$. Apply \cref{lem:cb-transfer} to the uniform law on $K_{n-1}$, after an affine map to isotropic position that carries the origin along as the reference point. It gives $\theta_\nu(\eps)\le C_{n,p,\alpha}(1+\log(1/\eps))$. Finally $B_D(h^\star,r)\subseteq B_\nu(h^\star,Lr)$ and $D(\DIS(\mathcal H))\le L\,\nu(\DIS(\mathcal H))$ for every set $\mathcal H$ of hypotheses, which transfers the bound with an extra factor at most $L^2$ (radii $Lr>1$ are trivial).
\end{proof}
Symmetry gives depth $1/2$ at the origin, and above $-1/(n+1)$ the mean has class-uniform positive depth by the Gr\"unbaum-type inequality used in \cref{thm:disagree-gen}. In the intermediate general class the mean-shifting examples of \cref{prop:dis-blowup} lose this depth. So \cref{prop:depth} gives logarithmic disagreement at reference points of uniformly positive depth for all $s>-1/n$, while \cref{thm:exp-upper} gives the uniform polynomial bound.


\begin{thebibliography}{99}\small
\bibitem[{Adamczak et~al.(2012)}]{AGLLOPT12} R.~Adamczak, O.~Gu\'edon, R.~Lata{\l}a, A.~E.~Litvak, K.~Oleszkiewicz, A.~Pajor, and N.~Tomczak-Jaegermann. Moment estimates for convex measures. \emph{Electron.~J.~Probab.}, 17, no.~101, 2012.
\bibitem[{Ajtai et~al.(2015)}]{AFHN15} M.~Ajtai, V.~Feldman, A.~Hassidim, and J.~Nelson. Sorting and selection with imprecise comparisons. arXiv:1501.02911, 2015; preliminary version ICALP 2009.
\bibitem[{Alon et~al.(2021)}]{AHHM21} N.~Alon, S.~Hanneke, R.~Holzman, and S.~Moran. A theory of PAC learnability of partial concept classes. \emph{FOCS}, 2021.
\bibitem[{Attias et~al.(2025)}]{ABNSSW25} I.~Attias, A.~Blum, K.~Naggita, D.~Saless, D.~Sharma, and M.~Walter. PAC learning with improvements. \emph{ICML}, 2025.
\bibitem[{Awasthi et~al.(2017)}]{ABL17} P.~Awasthi, M.-F.~Balcan, and P.~M.~Long. The power of localization for efficiently learning linear separators with noise. \emph{J.~ACM}, 63(6):50, 2017.
\bibitem[{Awasthi et~al.(2013)}]{AFK13} P.~Awasthi, V.~Feldman, and V.~Kanade. Learning using local membership queries. \emph{COLT}, PMLR 30:398--431, 2013.
\bibitem[{Balcan(2020)}]{Bal20} M.-F.~Balcan. Data-driven algorithm design. In T.~Roughgarden, editor, \emph{Beyond the Worst-Case Analysis of Algorithms}. Cambridge University Press, 2020. arXiv:2011.07177.
\bibitem[{Balcan and Long(2013)}]{BL13} M.-F.~Balcan and P.~M.~Long. Active and passive learning of linear separators under log-concave distributions. \emph{COLT}, 2013.
\bibitem[{Balcan and Sharma(2021)}]{BS21} M.-F.~Balcan and D.~Sharma. Data driven semi-supervised learning. \emph{NeurIPS}, 2021.
\bibitem[{Balcan and Zhang(2017)}]{BZ17} M.-F.~Balcan and H.~Zhang. Sample and computationally efficient learning algorithms under $s$-concave distributions. \emph{NeurIPS}, 2017. arXiv:1703.07758.
\bibitem[{Balcan et~al.(2009)}]{BBL09} M.-F.~Balcan, A.~Beygelzimer, and J.~Langford. Agnostic active learning. \emph{J.~Comput.~Syst.~Sci.}, 75(1):78--89, 2009.
\bibitem[{Balcan et~al.(2026)}]{BBFLS26} M.-F.~Balcan, A.~Blum, K.~Fragkia, Z.~Li, and D.~Sharma. Verify to amplify: improving reasoning via learned chain-of-thought verification. arXiv:2603.03538, 2026.
\bibitem[{Balcan et~al.(2022)}]{BBHS22} M.-F.~Balcan, A.~Blum, S.~Hanneke, and D.~Sharma. Robustly-reliable learners under poisoning attacks. \emph{COLT}, PMLR 178:4498--4534, 2022.
\bibitem[{Balcan et~al.(2025)}]{BBLS25} M.-F.~Balcan, A.~Blum, Z.~Li, and D.~Sharma. On learning verifiers and implications to chain-of-thought reasoning. \emph{NeurIPS}, 2025.
\bibitem[{Balcan et~al.(2023)}]{BHPS23} M.-F.~Balcan, S.~Hanneke, R.~Pukdee, and D.~Sharma. Reliable learning in challenging environments. \emph{NeurIPS}, 2023.
\bibitem[{Balcan et~al.(2010)}]{BHW10} M.-F.~Balcan, S.~Hanneke, and J.~Wortman Vaughan. The true sample complexity of active learning. \emph{Machine Learning}, 80(2--3):111--139, 2010.
\bibitem[{Ball(1988)}]{Bal88} K.~Ball. Logarithmically concave functions and sections of convex sets in $\mathbb R^n$. \emph{Studia Math.}, 88(1):69--84, 1988.
\bibitem[{B\'ar\'any(2008)}]{Bar08} I.~B\'ar\'any. Random points and lattice points in convex bodies. Lecture notes, 2008. \url{https://www.renyi.hu/~barany/cikkek/lect.pdf}.
\bibitem[{B\'ar\'any and Larman(1988)}]{BarL88} I.~B\'ar\'any and D.~G.~Larman. Convex bodies, economic cap coverings, random polytopes. \emph{Mathematika}, 35(2):274--291, 1988.
\bibitem[{B\'ar\'any et~al.(2020)}]{BFGHR20} I.~B\'ar\'any, M.~Fradelizi, X.~Goaoc, A.~Hubard, and G.~Rote. Random polytopes and the wet part for arbitrary probability distributions. \emph{Annales Henri Lebesgue}, 3:701--715, 2020.
\bibitem[{Barthe(1998)}]{Bar98} F.~Barthe. On a reverse form of the Brascamp--Lieb inequality. \emph{Invent.~Math.}, 134:335--361, 1998.
\bibitem[{Baum(1990)}]{Bau90} E.~B.~Baum. A polynomial time algorithm that learns two hidden unit nets. \emph{Neural Computation}, 2(4):510--522, 1990.
\bibitem[{Beygelzimer et~al.(2009)}]{BDL09} A.~Beygelzimer, S.~Dasgupta, and J.~Langford. Importance weighted active learning. \emph{ICML}, 2009.
\bibitem[{Black et~al.(2026)}]{BLMSS26} H.~Black, K.~G.~Larsen, A.~Mazumdar, B.~Saha, and G.~So. Actively learning halfspaces without synthetic data. \emph{COLT}, PMLR 336:708--728, 2026.
\bibitem[{Blum et~al.(1995)}]{BCGS95} A.~Blum, P.~Chalasani, S.~A.~Goldman, and D.~K.~Slonim. Learning with unreliable boundary queries. \emph{COLT}, pages 98--107, 1995.
\bibitem[{Blum et~al.(2025)}]{BHPS25} A.~Blum, S.~Hanneke, C.~Pabbaraju, and D.~Saless. Proofs as explanations: short certificates for reliable predictions. \emph{COLT}, PMLR 291:401--420, 2025.
\bibitem[{Bobkov(2007)}]{Bob07} S.~G.~Bobkov. Large deviations and isoperimetry over convex probability measures with heavy tails. \emph{Electron.~J.~Probab.}, 12:1072--1100, 2007.
\bibitem[{Bobkov(2010)}]{Bob10} S.~G.~Bobkov. Convex bodies and norms associated to convex measures. \emph{Probab.~Theory Relat.~Fields}, 147:303--332, 2010.
\bibitem[{Borell(1975)}]{Bor75} C.~Borell. Convex set functions in $d$-space. \emph{Period.~Math.~Hungar.}, 6(2):111--136, 1975.
\bibitem[{Bousquet et~al.(2020)}]{BHMZ20} O.~Bousquet, S.~Hanneke, S.~Moran, and N.~Zhivotovskiy. Proper learning, Helly number, and an optimal SVM bound. \emph{COLT}, PMLR 125:582--609, 2020.
\bibitem[{Brascamp and Lieb(1976)}]{BL76} H.~J.~Brascamp and E.~H.~Lieb. On extensions of the Brunn--Minkowski and Pr\'ekopa--Leindler theorems. \emph{J.~Funct.~Anal.}, 22(4):366--389, 1976.
\bibitem[{Caplin and Nalebuff(1991)}]{CN91} A.~Caplin and B.~Nalebuff. Aggregation and social choice: a mean voter theorem. \emph{Econometrica}, 59(1):1--23, 1991.
\bibitem[{Cohn et~al.(1994)}]{CAL94} D.~Cohn, L.~Atlas, and R.~Ladner. Improving generalization with active learning. \emph{Machine Learning}, 15(2):201--221, 1994.
\bibitem[{Dasgupta(2004)}]{Das04} S.~Dasgupta. Analysis of a greedy active learning strategy. \emph{NeurIPS}, 2004.
\bibitem[{Dasgupta(2005)}]{Das05} S.~Dasgupta. Coarse sample complexity bounds for active learning. \emph{NeurIPS}, 2005.
\bibitem[{Dasgupta et~al.(2007)}]{DHM07} S.~Dasgupta, D.~Hsu, and C.~Monteleoni. A general agnostic active learning algorithm. \emph{NeurIPS}, 2007.
\bibitem[{Diakonikolas et~al.(2024)}]{DKM24} I.~Diakonikolas, D.~M.~Kane, and M.~Ma. Active learning of general halfspaces: label queries vs membership queries. \emph{NeurIPS}, 2024.
\bibitem[{Diakonikolas et~al.(2023)}]{DTK23} I.~Diakonikolas, C.~Tzamos, and D.~M.~Kane. A strongly polynomial algorithm for approximate Forster transforms and its application to halfspace learning. \emph{STOC}, 2023.
\bibitem[{Du et~al.(2025)}]{DHS25} A.~Y.~Du, E.~Huang, and D.~Sharma. Tuning algorithmic and architectural hyperparameters in graph-based semi-supervised learning with provable guarantees. \emph{UAI}, PMLR 286:1084--1111, 2025.
\bibitem[{Dvir et~al.(2017)}]{DSW17} Z.~Dvir, S.~Saraf, and A.~Wigderson. Superquadratic lower bound for 3-query locally correctable codes over the reals. \emph{Theory of Computing}, 13(11):1--36, 2017. Preliminary version in \emph{STOC}, 2014.
\bibitem[{Ezra and Sharir(2019)}]{ES19} E.~Ezra and M.~Sharir. A nearly quadratic bound for point-location in hyperplane arrangements, in the linear decision tree model. \emph{Discrete Comput.~Geom.}, 61(4):735--755, 2019.
\bibitem[{Ezra et~al.(2020)}]{EHKS20} E.~Ezra, S.~Har-Peled, H.~Kaplan, and M.~Sharir. Decomposing arrangements of hyperplanes: VC-dimension, combinatorial dimension, and point location. \emph{Discrete Comput.~Geom.}, 64(1):109--173, 2020.
\bibitem[{Forster(2002)}]{For02} J.~Forster. A linear lower bound on the unbounded error probabilistic communication complexity. \emph{J.~Comput.~Syst.~Sci.}, 65(4):612--625, 2002.
\bibitem[{Fradelizi et~al.(2014)}]{FGP14} M.~Fradelizi, O.~Gu\'edon, and A.~Pajor. Thin-shell concentration for convex measures. \emph{Studia Math.}, 223(2):123--148, 2014.
\bibitem[{Fradelizi et~al.(2020)}]{FLM20} M.~Fradelizi, J.~Li, and M.~Madiman. Concentration of information content for convex measures. \emph{Electron.~J.~Probab.}, 25, no.~20, 2020.
\bibitem[{Gonen et~al.(2013)}]{GSS13} A.~Gonen, S.~Sabato, and S.~Shalev-Shwartz. Efficient active learning of halfspaces: an aggressive approach. \emph{JMLR}, 14:2583--2615, 2013.
\bibitem[{Hanneke(2007)}]{Han07} S.~Hanneke. A bound on the label complexity of agnostic active learning. \emph{ICML}, 2007.
\bibitem[{Hanneke(2012)}]{Han12} S.~Hanneke. Activized learning: transforming passive to active with improved label complexity. \emph{JMLR}, 13:1469--1587, 2012.
\bibitem[{Hanneke(2014)}]{Han14} S.~Hanneke. Theory of disagreement-based active learning. \emph{Foundations and Trends in Machine Learning}, 7(2--3):131--309, 2014.
\bibitem[{Hanneke(2016)}]{Han16} S.~Hanneke. The optimal sample complexity of PAC learning. \emph{JMLR}, 17(38):1--15, 2016.
\bibitem[{Hanneke(2024)}]{Han24star} S.~Hanneke. The star number and eluder dimension: elementary observations about the dimensions of disagreement. \emph{COLT}, PMLR 247:2308--2359, 2024.
\bibitem[{Hanneke(2025)}]{Han25} S.~Hanneke. Agnostic active learning is always better than passive learning. \emph{NeurIPS}, 2025.
\bibitem[{Hanneke and Yang(2015)}]{HY15} S.~Hanneke and L.~Yang. Minimax analysis of active learning. \emph{JMLR}, 16:3487--3602, 2015.
\bibitem[{Hanneke and Yang(2021)}]{HY21} S.~Hanneke and L.~Yang. Toward a general theory of online selective sampling: trading off mistakes and queries. \emph{AISTATS}, PMLR 130, 2021.
\bibitem[{Hanneke et~al.(2026)}]{HKMeV26} S.~Hanneke, A.~Karbasi, A.~Mehrotra, and G.~Velegkas. Language generation with feedback: queries and mistakes. \emph{ICML}, PMLR 306:40327--40352, 2026.
\bibitem[{Hanneke et~al.(2024)}]{HKMV24} S.~Hanneke, A.~Karbasi, S.~Moran, and G.~Velegkas. Universal rates for active learning. \emph{NeurIPS}, 2024.
\bibitem[{Hanneke et~al.(2025b)}]{HMT25} S.~Hanneke, S.~Moran, and M.~Thiessen. Marginal-nonuniform PAC learnability. \emph{NeurIPS}, 2025.
\bibitem[{Hanneke et~al.(2025a)}]{HYWS25} S.~Hanneke, L.~Yang, G.~Wang, and Y.~Song. Reliable active apprenticeship learning. \emph{ALT}, PMLR 272:512--538, 2025.
\bibitem[{Hardt and Moitra(2013)}]{HM13} M.~Hardt and A.~Moitra. Algorithms and hardness for robust subspace recovery. \emph{COLT}, 2013.
\bibitem[{Hopkins et~al.(2020c)}]{HKL20} M.~Hopkins, D.~Kane, and S.~Lovett. The power of comparisons for actively learning linear classifiers. \emph{NeurIPS}, 2020.
\bibitem[{Hopkins et~al.(2020a)}]{HKLM20noise} M.~Hopkins, D.~Kane, S.~Lovett, and G.~Mahajan. Noise-tolerant, reliable active classification with comparison queries. \emph{COLT}, PMLR 125:1957--2006, 2020.
\bibitem[{Hopkins et~al.(2020b)}]{HKLM20} M.~Hopkins, D.~Kane, S.~Lovett, and G.~Mahajan. Point location and active learning: learning halfspaces almost optimally. \emph{FOCS}, 2020.
\bibitem[{Kane et~al.(2018)}]{KLM18} D.~Kane, S.~Lovett, and S.~Moran. Generalized comparison trees for point-location problems. \emph{ICALP}, 2018.
\bibitem[{Kane et~al.(2017)}]{KLMZ17} D.~Kane, S.~Lovett, S.~Moran, and J.~Zhang. Active classification with comparison queries. \emph{FOCS}, 2017.
\bibitem[{Katz-Samuels et~al.(2021)}]{KZJJ21} J.~Katz-Samuels, J.~Zhang, L.~Jain, and K.~Jamieson. Improved algorithms for agnostic pool-based active classification. \emph{ICML}, PMLR 139:5334--5344, 2021.
\bibitem[{Kearns(1998)}]{Kea98} M.~Kearns. Efficient noise-tolerant learning from statistical queries. \emph{J.~ACM}, 45(6):983--1006, 1998.
\bibitem[{Kulkarni et~al.(1993)}]{KMT93} S.~R.~Kulkarni, S.~K.~Mitter, and J.~N.~Tsitsiklis. Active learning using arbitrary binary valued queries. \emph{Machine Learning}, 11:23--35, 1993.
\bibitem[{Li and Zhang(2024)}]{LZ24} Y.~Li and C.~Zhang. Efficient active learning halfspaces with Tsybakov noise: a non-convex optimization approach. \emph{AISTATS}, PMLR 238:4744--4752, 2024.
\bibitem[{Madiman et~al.(2017)}]{MMX17} M.~Madiman, J.~Melbourne, and P.~Xu. Forward and reverse entropy power inequalities in convex geometry. In \emph{Convexity and Concentration}, IMA Vol.~161, Springer, 2017.
\bibitem[{Marsiglietti and Pandey(2021)}]{MP21} A.~Marsiglietti and P.~Pandey. On the equivalence of statistical distances for isotropic convex measures. arXiv:2112.09009, 2021.
\bibitem[{Meiser(1993)}]{Mei93} S.~Meiser. Point location in arrangements of hyperplanes. \emph{Information and Computation}, 106(2):286--303, 1993.
\bibitem[{Ross et~al.(2011)}]{RGB11} S.~Ross, G.~J.~Gordon, and J.~A.~Bagnell. A reduction of imitation learning and structured prediction to no-regret online learning. \emph{AISTATS}, PMLR 15:627--635, 2011.
\bibitem[{Sharma and Jones(2023)}]{SJ23} D.~Sharma and M.~Jones. Efficiently learning the graph for semi-supervised learning. \emph{UAI}, PMLR 216:1900--1910, 2023.
\bibitem[{Sharma and Sun(2025)}]{SS25} D.~Sharma and A.~Sun. Conservative classifiers do consistently well with improving agents: characterizing statistical and online learning. \emph{NeurIPS}, 2025.
\bibitem[{Wang and Dughmi(2026)}]{WD26} G.~Wang and S.~Dughmi. On the sample complexity of active learning with membership queries. arXiv:2609.27241, 2026.
\bibitem[{Wiener et~al.(2015)}]{WHE15} Y.~Wiener, S.~Hanneke, and R.~El-Yaniv. A compression technique for analyzing disagreement-based active learning. \emph{JMLR}, 16:713--745, 2015.
\bibitem[{Zhang and Chaudhuri(2014)}]{ZC14} C.~Zhang and K.~Chaudhuri. Beyond disagreement-based agnostic active learning. \emph{NeurIPS}, 2014.
\bibitem[{Zhang and Li(2021)}]{ZL21} C.~Zhang and Y.~Li. Improved algorithms for efficient active learning halfspaces with Massart and Tsybakov noise. \emph{COLT}, PMLR 134, 2021.
\end{thebibliography}
\end{document}